\documentclass[journal]{IEEEtran}

\usepackage{amsmath,amsfonts,amssymb}
\usepackage{graphicx}
\usepackage{cite}
\usepackage{algorithm}
\usepackage{algpseudocode}
\usepackage{pgffor}
\usepackage{multirow}
\usepackage{amssymb}   
\usepackage{pifont}  
\usepackage{colortbl}
\usepackage{xcolor}
\usepackage{placeins}
\usepackage{float}
\usepackage{xcolor}
\usepackage{comment}
\usepackage{flushend}
\usepackage{enumitem}

\def\KM{\text{ECLIPSE}}
\begin{document}

\title{$\KM$: Embedding-Centered Foundation-Model Federated Knowledge Transfer for Data-Free Edge Intelligence
}

\author{S. Kawa Atapour$^{\dag}$, S. Jamal Seyedmohammadi$^{\dag}$, Maryam Hosseini$^{\dag}$, S. Mohammad Sheikholeslami$^{\ddag}$, Jamshid Abouei$^{\dag\dag}$, Konstantinos Plataniotis$^{\ddag}$, and Arash Mohammadi$^{\dag}$
\thanks{ ${^\dag}$ S.K. Atapour, S.J. Seyedmohammadi, M. Hosseini, and A. Mohammadi are with the I-SIP Lab at Cybersecurity \& Intelligent Systems Engineering (CISE), Concordia University, Montreal, QC,
Canada. \\
${^\ddag}$ M. Sheikholeslami and K.N. Plataniotis are with Multimedia Lab at Electrical \& Computer Engineering (ECE) Department, University of Toronto, Toronto, ON, Canada.\\
${^{\dag\dag}}$ J. Abouei is with the Department of Electrical Engineering, Yazd University, Yazd, Iran.}
\thanks{This work was partially supported by Natural Sciences \& Eng. Research Council (NSERC) of Canada; NSERC Discovery Grant RGPIN-2023-05654.}}
\markboth{IEEE Transactions on Reliability, Special Section on Trustworthy and Reliable AI}%
{Atapour \MakeLowercase{\textit{et al.}}: Resource-Constrained Data-Free Federated Edge Learning}

\maketitle

\begin{abstract}
Foundation Models (FMs) are envisioned to become an integral component of intelligent federated edge networks because of their strong representation and generalization capabilities. However, their integration into resource-constrained and privacy-sensitive edge environments remains challenging. This challenge is particularly pronounced in data-free settings, where no shared, public, or server-accessible reference dataset is available to establish a common reference space for aligning distributed knowledge. To address this challenge, we propose $\KM$, which is an embedding-centered FM-based federated knowledge transfer framework developed for data-free edge intelligence. $\KM$ enables edge intelligence capitalizing on a server-hosted FM without sharing raw data, local models, or a public reference dataset. It establishes a shared embedding reference space through which edge devices encode their local task knowledge. Only the resulting embeddings are transmitted to the server for alignment and aggregation. Using the aggregated embeddings, the server constructs a task-aware surrogate dataset through an embedding-guided generative process. This surrogate dataset is then used to adapt the server-hosted FM to the downstream task. The adapted FM is subsequently distilled into a lightweight global model and distributed to the edge devices. This process allows edge devices to benefit from both distributed task knowledge and the generalization capabilities of the FM. Experimental results across diverse classification datasets show that $\KM$ outperforms conventional Federated Learning (FL) and data-free Federated Knowledge-Transfer (FKT) baselines. It also achieves competitive performance compared with data-based approaches. Furthermore, the framework remains robust under statistical and model heterogeneity across different data distributions and local architectures. Its single-round knowledge-transfer mechanism reduces communication energy, while server-side FM deployment limits the computational burden imposed on edge devices. These results demonstrate that embedding-based alignment provides an effective foundation for data-free FM adaptation in resource-constrained federated edge networks.
\end{abstract}

\section{Introduction}

The rapid advancement of edge devices in data collection and processing is fostering a new era of automation and intelligence.
However, transmitting locally collected data to centralized servers introduces significant challenges, including privacy risks, data-governance constraints, and communication overhead~\cite{collins2025federated}. These challenges limit the applicability of traditional centralized learning approaches in distributed edge systems~\cite{IoT_network}. Federated Learning (FL) addresses these challenges by enabling edge devices to collaboratively learn from distributed data while keeping raw data local~\cite{FL_survey_2024}. As a result, FL has emerged as a promising paradigm for enabling privacy-aware intelligence at the network edge. The performance of FL, however, is often constrained by the scarcity of local data and the statistical heterogeneity across edge devices. These limitations can hinder the learning of effective global models.

Recent advances in Foundation Models (FMs) have introduced promising opportunities to address these limitations in FL. Pre-trained on large-scale and diverse datasets, FMs provide strong general-purpose representations and zero-shot capabilities~\cite{CLIP2021}. Leveraging these capabilities in federated edge networks can significantly enhance the performance of edge devices, particularly when local datasets are limited or highly heterogeneous. However, for many downstream tasks whose data distributions differ from those used for FM pre-training, task-specific adaptation is required to achieve satisfactory performance~\cite{vanveldhuizen2025fmmedical}. The knowledge required for such adaptation is distributed across edge devices and cannot be directly accessed by the FM located at the server side. Therefore, enabling edge devices to effectively benefit from FMs requires mechanisms that can adapt pre-trained FMs using distributed task-specific knowledge.

Several strategies have been explored to integrate FMs into federated environments. Based on where the FM is deployed and how its knowledge is transferred~\cite{open-vocab,prompt-diversification,SIDCLIP}, such existing approaches can be broadly categorized as follows:
\begin{itemize}[nosep]
\item[(i)] \textit{Edge-FM Approaches~\cite{FedCLIP,FedPrompt,FedLoRA}:} In methods belonging to this category, FMs are integrated into FL by deploying them directly on edge devices and adapting them using local data~\cite{promptfl,HeLoRA}. Such a design enables edge devices to directly benefit from the representational capabilities of FMs and adapt them to local tasks. However, many edge devices operate under strict computational, memory, and energy constraints, whereas FMs typically require substantial resources for both fine-tuning and inference. We refer to such devices and environments as \emph{resource-constrained edge devices} and \emph{resource-constrained edge networks}, respectively. These environments lack sufficient resources to deploy and execute off-the-shelf large FMs locally. In addition, many commercial FMs are proprietary, making their parameters unavailable for deployment beyond the server. As a result, although Edge-FM methods are suitable for general FL scenarios, their reliance on local FM deployment makes them impractical for many resource-constrained federated edge networks.
%
\item[(ii)] \textit{Server-FM Approaches~\cite{KOALA,FedDCL}:} In this category, the FM remains deployed on the server, while edge devices employ lightweight local models. Such a design avoids deploying large FMs on edge devices and keeps FM parameters confined to the server. It is, therefore, more suitable for resource-constrained edge networks. However, the knowledge required for task-specific FM adaptation resides on edge devices, making it challenging to adapt a server-hosted FM to downstream tasks. Existing approaches, typically, rely on publicly available datasets or local model transfer to communicate such knowledge to the server~\cite{KOALA}. These requirements can limit practical applicability. Public datasets may be unavailable, unsuitable for the downstream task, or incompatible with data-governance requirements. Transmitting local models can also introduce additional communication overhead and potential security risks~\cite{konevcny2016federated}.
%
\item[(iii)] \textit{Compressed-FM Approaches~\cite{open-vocab,prompt-diversification,SIDCLIP}:} These methods were originally developed in centralized settings to transfer the knowledge of large FMs into lightweight models suitable for deployment on resource-constrained devices. Such methods can potentially be used in FL by deploying the transferred model on edge devices. However, the knowledge-transfer process is typically performed using a pre-trained FM without first adapting it to the downstream task using knowledge distributed across edge devices. Consequently, the resulting lightweight model may not incorporate task-specific knowledge from the federated environment and may perform poorly under heterogeneous local data distributions~\cite{li2020federated}. 
\end{itemize}
\textit{\textbf{The Reference-Space Problem:}} Regardless of how FMs are integrated into FL, adapting an FM to a downstream task requires utilizing the knowledge distributed across edge devices. This process requires a common space in which edge devices can express their local knowledge and from which a global knowledge representation can be obtained for FM adaptation. We refer to such a space as a \emph{reference space}. In conventional FL, the reference space is defined by model parameters. Edge devices encode their local knowledge as parameter updates, which are subsequently aggregated~\cite{FedAvg}. While effective, this approach incurs substantial communication overhead and may expose local models to model-inversion or related inference attacks. Federated Knowledge Transfer (FKT) replaces parameter exchange with compact knowledge representations, such as logits or feature maps~\cite{FedMD}. In this setting, a publicly available dataset, commonly, serves as the reference space. It provides a shared input domain on which knowledge from different edge devices can be aligned and aggregated. However, the availability of such a dataset cannot always be assumed because of privacy, ownership, accessibility, or domain-mismatch constraints. Data-free FKT removes the dependence on shared public data. Here, \emph{data-free} refers to the absence of a shared, public, or server-accessible reference dataset rather than the absence of private local data at the edge. Without a shared reference space, however, knowledge from different edge devices cannot be consistently aligned and aggregated. This limits the effectiveness of existing data-free FKT approaches for FM adaptation~\cite{fedd2s}. 

\setlength{\textfloatsep}{0pt}
\begin{figure*}[t!]
\centering
\includegraphics[width=\textwidth]{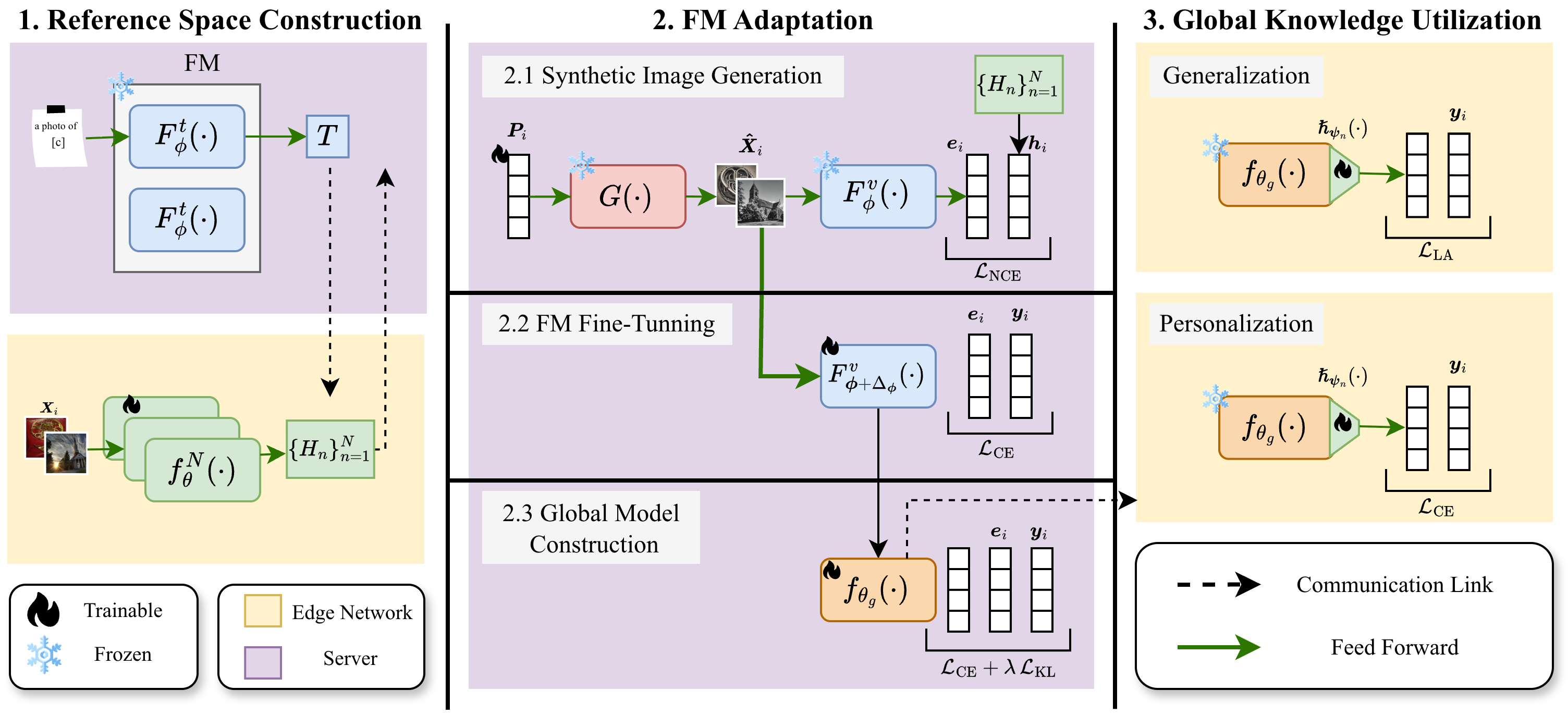}
\caption{
Overview of the proposed data-free federated knowledge distillation framework. 
\textit{(1)} \textbf{Reference Space Construction:} The server uses the VLM to generate class anchors that define a shared reference space. Each edge device trains its local model as a third encoder by aligning its embeddings to these anchors. 
\textit{(2)} \textbf{FM Adaptation:} Edge device embeddings are aggregated at the server and used as the supervision to generate synthetic samples via prompt optimization in a text-to-image generator. The generated data is then used to adapt the FM and distill its knowledge into a lightweight global model. 
\textit{(3)} \textbf{Global Knowledge Utilization:} The global model is deployed to edge devices and used either for generalization or personalization depending on the target objective.
}
\label{fig1}
\end{figure*}
\vspace{.1in}
\noindent
\textbf{\textit{Contributions:}} Motivated by the above limitations, this paper aims to address the following research question: \textit{How can the capabilities of large and proprietary FMs be leveraged for resource-constrained edge devices within a data-free FKT framework?} 

To address this question, we propose the Embedding-Centered Knowledge Transfer for Data-Free Federated Edge Intelligence $\KM$ framework. $\KM$ adopts a Server-FM architecture in which the FM remains on the server, while edge devices operate using lightweight local models. Consequently, the computational and memory demands associated with FM deployment and adaptation are shifted to the server. This design makes the proposed framework suitable for resource-constrained edge networks. 

The key challenge is not only to transfer local knowledge, but also to represent knowledge from heterogeneous local models in a common space that can be interpreted by the server-hosted FM. To address this challenge, $\KM$ replaces the conventional shared input reference space with a shared embedding reference space. As illustrated in Fig.~\ref{fig1}, the proposed framework consists of three stages: \textit{reference-space construction}, \textit{FM adaptation}, and \textit{global knowledge utilization}. First, edge devices encode their task-specific knowledge into compact embeddings within a shared embedding reference space. Only these embeddings are transmitted to the server for alignment and aggregation. The server then reconstructs a task-aware surrogate dataset from the aggregated embeddings. This surrogate dataset is used to adapt the FM and distill its knowledge into a lightweight global model. Finally, the global model is distributed to the edge devices to support either generalization or personalization.

By replacing the conventional shared input space with a shared embedding reference space, ECLIPSE enables distributed knowledge aggregation without requiring a public reference dataset or the exchange of local models. The framework also avoids transmitting raw local data and thereby reduces direct raw-data exposure. Moreover, its knowledge-transfer process requires a single edge-to-server embedding upload and a single server-to-edge global-model distribution after local training, rather than iterative server--client communication rounds. The main contributions of this paper are summarized as follows:
\begin{itemize}
\item We propose $\KM$, a data-free federated knowledge-transfer framework that replaces the conventional shared input reference space with a shared embedding reference space. The proposed design enables consistent aggregation of distributed task-specific knowledge without requiring public datasets or local model exchange. It also supports single-round knowledge transfer, substantially reducing communication overhead.
\item We develop an embedding-driven reconstruction mechanism that transforms aggregated embeddings into a task-aware surrogate dataset for adapting a server-hosted FM. By communicating compact embedding representations instead of raw data or model parameters, the proposed framework enables efficient FM adaptation while reducing communication cost and limiting information exposure at edge devices.
\item We enable the adapted FM knowledge to be transferred into a lightweight global model through server-side distillation. The resulting global model can be deployed on resource-constrained edge devices and further adapted to support either global generalization or local personalization.
\end{itemize}
We validate the proposed framework through extensive experiments on diverse image-classification tasks. Experimental results demonstrate the effectiveness of $\KM$ across diverse datasets, data distributions, and local model architectures. The proposed framework consistently outperforms conventional FL, data-free FKT, and server-FM baselines on most evaluated datasets, while remaining robust under both statistical and model heterogeneity. On CIFAR-10, $\KM$ achieves an accuracy of $64.82\%$ while consuming an average of $285.4$ J per edge device. Compared with KOALA, it improves accuracy by $13.34$ percentage points while reducing energy consumption by approximately $57.7\%$. Compared with FedDCL, it improves accuracy by $4.29$ percentage points with nearly identical energy consumption. Although Edge-FM methods attain higher accuracy, they require approximately $4.6\times$ to $6.0\times$ more edge-device energy. These results demonstrate that proposed $KM$ provides a favorable accuracy-energy tradeoff for resource-constrained federated edge networks.

The remainder of the paper is organized as follows: Section~\ref{sec:review} reviews related work on FM integration in FL, data-free FKT, and the associated research gap. Section~\ref{sec:ECLIPS} presents the proposed $\KM$ framework, including reference-space construction, FM adaptation, global knowledge utilization, and theoretical analysis. Section~IV describes the experimental setup and evaluates classification performance, robustness to statistical and model heterogeneity, the accuracy--energy tradeoff, synthetic-data quality, and the contribution of individual framework components. Finally, Section~V concludes the paper.

\section{Related Works} \label{sec:review}
This section provides a further in-depth review of existing approaches related to FM adaptation within the context of FL intelligence. First, we discuss FM integration strategies based on where the FM is deployed and how distributed knowledge is transferred. We then review data-free FKT methods, with particular emphasis on how they establish a reference space for aligning and aggregating knowledge from heterogeneous edge devices. Finally, we present the identified research gap. Through these reviews, we aim to highlight the absence of a shared reference space for data-free FM adaptation, which motivates the proposed embedding-based framework.

\vspace{.05in}
\noindent
\textit{A. Edge-FM Methods}

As stated previously, edge-FM methods adapt FM components directly on the edge side using local data. A common objective of these methods is to transfer the knowledge distributed across edge devices to the FM through parameter-efficient adaptation mechanisms. FedCLIP~\cite{FedCLIP}, FedPrompt~\cite{FedPrompt}, and FedLoRA~\cite{FedLoRA} follow this paradigm by adapting locally deployed FMs through adapters, prompts, and low-rank updates, respectively. FedHPL~\cite{FedHPL} extends this idea to heterogeneous federated environments, where edge devices perform prompt tuning and exchange class-level logits for knowledge aggregation. Similarly, FedPGA~\cite{FedPGA} utilizes CLIP on edge devices and represents local knowledge using class prototypes, which are aggregated at the server. More recently, FedMVP~\cite{FedMVP} performs federated adaptation of CLIP through multimodal prompt tuning, where prompts learned from local data are exchanged and aggregated across edge devices. Although these methods differ in how local knowledge is represented and communicated, they all require FM components to be deployed and executed on edge devices. Consequently, their applicability is limited in resource-constrained edge networks and in scenarios where FM parameters cannot be distributed beyond the server.

\vspace{.05in}
\noindent
\textit{B. Server-FM Methods}

Server-FM methods deploy the FM on a central server while edge devices employ lightweight local models. This architecture eliminates the need to deploy large FMs on edge devices while keeping FM parameters confined to the server, making it suitable for resource-constrained edge networks~\cite{KOALA, FedDCL}. The main challenge in this setting is that knowledge extracted by heterogeneous local models may reside in different representation spaces. Consequently, the server requires a mechanism that aligns the knowledge of different edge devices and relates it to the server-hosted FM before the knowledge can be consistently interpreted and utilized for FM adaptation. KOALA~\cite{KOALA} addresses this challenge by using local models as the reference space. Specifically, edge devices train lightweight local models and upload them to the server, where a co-distillation mechanism transfers knowledge between the local models and the server-hosted FM. In contrast, FedDCL~\cite{FedDCL} leverages a pretrained diffusion model at the server to generate synthetic data from class-specific prototypes communicated by edge devices. These generated samples are subsequently used for knowledge transfer and server-model adaptation. The exchanged prototypes, however, originate from heterogeneous local models and the method does not explicitly define how such prototypes are aligned into a common representation space prior to server-side utilization. Consequently, neither local-to-local alignment nor local-to-server alignment is explicitly established, and the consistency of the transferred knowledge relies on assumptions regarding the compatibility of the communicated prototypes across heterogeneous feature spaces. Although Server-FM methods avoid deploying FMs on edge devices, they either require transmitting local models or depend on an auxiliary pretrained model to establish a shared reference space, therefore, introducing additional communication overhead and system complexity.

\vspace{.05in}
\noindent
\textit{C. Compressed-FM Methods}

Compressed-FM methods aim to transfer the capabilities of large FMs into compact models suitable for deployment on resource-constrained edge devices. These methods are primarily developed in centralized settings and focus on preserving the general representations learned by pretrained FMs. Open-Vocabulary Distillation~\cite{open-vocab} transfers the zero-shot capabilities of CLIP into lightweight student models. Notably, the method operates in a data-free fashion by generating synthetic samples using a diversified style dictionary. SIDCLIP~\cite{SIDCLIP} incorporates downstream-task data to distill the few-shot capabilities of CLIP into a more efficient student model. To mitigate data scarcity, synthetic images are generated using CLIP together with a text-to-image diffusion model. Similarly, Prompt Diversification~\cite{prompt-diversification} presents a data-free customization framework that distills CLIP into a compact student model using diverse prompt-generated surrogate images. Although these methods produce lightweight models suitable for edge deployment, they primarily transfer the generic knowledge of pretrained FMs and do not incorporate the task-specific knowledge distributed across edge devices. Consequently, their ability to adapt to distributed local data within the FL settings is limited.

\vspace{.05in}
\noindent
\textit{D. Data-Free Federated Knowledge Transfer (FKT)}

The FKT approach replaces parameter exchange with the transfer of compact knowledge representations such as logits, prototypes, or feature embeddings~\cite{FedPKD, FedDCL}. A key requirement of the FKT is a reference space in which knowledge from different edge devices can be represented in a common form and aggregated. FedPKD~\cite{FedPKD} employs prototype-based knowledge distillation without relying on public data. However, the exchanged prototypes are generated independently by different edge devices, and no explicit mechanism is provided to align them into a common representation space before server-side aggregation. FedDCL~\cite{FedDCL} attempts to facilitate knowledge transfer through class-specific prototypes and diffusion-model-based synthetic data generation. The prototypes are, however, extracted from heterogeneous local models, and the method does not explicitly provide a mechanism for aligning these representations into a common feature space prior to server-side processing. As a result, although the diffusion model serves as an auxiliary mechanism for synthetic data generation, neither the alignment of heterogeneous local representations nor the alignment between local representations and the server model is explicitly addressed. These limitations highlight the need for a mechanism that can establish a shared reference space in which knowledge from different edge devices is first aligned and then effectively aggregated.

\vspace{.05in}
\noindent
\textit{E. Research Gap}

\begin{table*}[t]
\centering
\caption{Comparison of existing FL methods in terms of FM deployment, reference-space construction, and knowledge transfer mechanisms.}
\label{tab:comparison}
\vspace{2mm}
\resizebox{\textwidth}{!}{%
\begin{tabular}{c|c|c|c|c|c|c|c|c}
\hline
\textbf{Category} &
\textbf{Method} &
\textbf{Reference Space} &
\textbf{Knowledge Sent} &
\textbf{\textit{Data-Free}} &
\textbf{\textit{No Model Transfer}} &
\textbf{\textit{Low Edge Resource}} &
\textbf{\textit{Low Communication}} &
\textbf{\textit{Knowledge Alignment}} \\
\hline

\multirow{5}{*}{Edge-FM}
& FedLoRA~\cite{FedLoRA}
& Parameter space
& LoRA updates
& \checkmark
& \ding{55}
& \ding{55}
& \checkmark
& \checkmark \\

& FedHPL~\cite{FedHPL}
& Logit space
& Class logits
& \ding{55}
& \checkmark
& \ding{55}
& \checkmark
& \checkmark \\

& FedPGA~\cite{FedPGA}
& CLIP feature space
& Class prototypes
& \checkmark
& \checkmark
& \ding{55}
& \ding{55}
& \checkmark \\

& FedMVP~\cite{FedMVP}
& Prompt space
& Prompts
& \checkmark
& \checkmark
& \ding{55}
& \checkmark
& \checkmark \\

& FedPrompt~\cite{FedPrompt}
& Prompt space
& Prompts
& \checkmark
& \checkmark
& \ding{55}
& \checkmark
& \checkmark \\
\hline

\multirow{2}{*}{Server-FM}
& KOALA~\cite{KOALA}
& Local models
& Local models
& \ding{55}
& \ding{55}
& \checkmark
& \ding{55}
& \checkmark \\

& FedDCL~\cite{FedDCL}
& Diffusion-assisted prototype space
& Class prototypes
& \checkmark
& \checkmark
& \checkmark
& \checkmark
& \ding{55} \\
\hline

\multirow{3}{*}{Compressed-FM}
& Open-Vocab~\cite{open-vocab}
& Synthetic image space
& NA
& NA
& NA
& \checkmark
& NA
& NA \\

& SIDCLIP~\cite{SIDCLIP}
& Synthetic image space
& NA
& NA
& NA
& \checkmark
& NA
& NA \\

& Prompt Diversification~\cite{prompt-diversification}
& Prompt-generated image space
& NA
& NA
& NA
& \checkmark
& NA
& NA \\
\hline

\multirow{1}{*}{Data-Free FKT}
& FedPKD~\cite{FedPKD}
& \ding{55}
& Class prototypes
& \checkmark
& \checkmark
& \checkmark
& \checkmark
& \ding{55} \\
\hline

\multirow{1}{*}{ Embedding-Guided FMs }
& \text{ECLIPSE}
& Reference Space Construction
& Embeddings
& \checkmark
& \checkmark
& \checkmark
& \checkmark
& \checkmark \\
\hline

\end{tabular}
}
\end{table*}

Existing approaches to FM adaptation in FL differ primarily in where the FM is deployed, how knowledge is transferred, and how a shared reference space is established. Edge-FM methods support task-specific FM adaptation but require deploying large FMs on edge devices. Server-FM methods avoid this requirement but typically rely on local model exchange or auxiliary generative models to facilitate knowledge transfer. However, existing methods do not explicitly establish a shared representation space that aligns heterogeneous local models with one another and with the server-hosted FM before knowledge aggregation and FM adaptation. Compressed-FM methods enable lightweight deployment but do not exploit the task-specific knowledge distributed across edge devices. Although data-free FKT methods eliminate the need for public datasets, they still lack an effective shared reference. These limitations motivate the need for a framework that enables adaptation of server-hosted FMs in resource-constrained edge networks through a shared embedding reference space without requiring public datasets or local model exchange. Table~\ref{tab:comparison} summarizes the key differences among the existing approaches.

\section{The $\KM$ Framework} \label{sec:ECLIPS}
In this paper, we consider a realistic federated edge network model consisting of a central server and a set of $N$ edge devices, denoted by $\mathbb{U}=\{u_1,\ldots,u_N\}$. Each edge device $u_n \in \mathbb{U}$ possesses a local dataset $\mathbb{D}_n=\{(x_n^i,y_n^i)\}_{i=1}^{|\mathbb{D}_n|}$ drawn from a local distribution $\mathcal{D}_n$ for a $C$-class classification task, and employs a local model $f_{\boldsymbol{\theta}_n}^{\,n}(\cdot)$ parameterized by $\boldsymbol{\theta}_n$. The architecture of the local model is selected according to the computational, memory, and energy resources available at the device and may, therefore, differ across edge devices. The server hosts an FM $F_{\boldsymbol{\phi}}(\cdot)$ parameterized by $\boldsymbol{\phi}$. The objective is to adapt the server-hosted FM using the task-specific knowledge distributed across edge devices and subsequently transfer the adapted knowledge back to the edge devices, without transferring raw data or local models.

As illustrated in Fig.~\ref{fig1}, the proposed $\KM$ framework consists of three stages: (A) \textit{Reference Space Construction}, where local models are aligned to a common embedding space; (B) \textit{FM Adaptation}, where the server adapts the FM using the aggregated knowledge encoded in that space, and; (C) \textit{Global Knowledge Utilization}, where the adapted knowledge is transferred back to local models to improve either generalization or personalization. The framework operates in a single communication round, in which each edge device transmits its knowledge representation only once, therefore, significantly reducing communication overhead.

\subsection{Reference Space Construction}
The objective of this stage is to construct a shared embedding reference space that enables consistent representation and aggregation of task-specific knowledge from heterogeneous edge devices. To this end, we define the reference space as the embedding space induced by the visual encoder of the server-hosted FM, as follows
\begin{equation}
\mathcal{R}
=
\left\{
\mathbf{z}\in\mathbb{R}^{d}
\,\middle|\,
\mathbf{z}=F^{v}_{\boldsymbol{\phi}}(x),
\;
x\in\mathcal{X}
\right\},
\end{equation}
where $F^{v}_{\boldsymbol{\phi}}(\cdot)$ denotes the visual encoder of the server-hosted FM, $\mathcal{X}$ is its input domain, and $d$ is the embedding dimension. By mapping the representations learned by all edge devices into the shared embedding space $\mathcal{R}$, the server can directly compare and aggregate their task-specific knowledge.

To construct a semantically meaningful reference space, let us assume that the server-hosted FM is a Vision--Language Model (VLM) composed of a visual encoder $F^v_{\boldsymbol{\phi}}(\cdot)$ and a text encoder $F^t_{\boldsymbol{\phi}}(\cdot)$. Since these encoders are jointly trained to map semantically corresponding visual and textual inputs into a shared embedding space, $\mathcal{R}$ provides a semantically meaningful representation space~\cite{CLIP2021}. Each local model $f_{\boldsymbol{\theta}_n}^{\,n}(\cdot)$ is treated as an additional encoder that learns a mapping as 
\begin{eqnarray}
f_{\boldsymbol{\theta}_n}^{\,n}:\mathcal{X}\rightarrow\mathcal{R}.\nonumber
\end{eqnarray}
Since the server does not have access to edge devices' local data, the alignment process relies solely on class-level supervision. Specifically, let $\mathcal{Q}=\{q_1,\ldots,q_C\}$ denote the set of class names. For each class $c$, the server generates multiple textual prompts and encodes them using $F^t_{\boldsymbol{\phi}}(\cdot)$. The resulting embeddings are averaged to obtain a class anchor $\mathbf{t}_c \in \mathcal{R}$. Collectively, the anchors form the matrix $\mathbf{T}=[\mathbf{t}_1,\ldots,\mathbf{t}_C]^\top \in \mathbb{R}^{C\times d},$
which captures the class-level semantic structure of the reference space. The matrix $\mathbf{T}$ is then broadcast to all edge devices.

Using the received anchors, each edge device trains its local encoder by minimizing 
\begin{equation} \label{eq:stage1} \mathcal{L}_{\text{align}} = \underset{(x,y)\sim\mathcal{D}_n}{\mathbb{E}} \left[ -\log \frac{ \exp \left(\mathrm{sim}(f_{\boldsymbol{\theta}_n}^{n}(x),\mathbf{t}_y)/\tau \right) } { \sum_{c=1}^{C} \exp \left(\mathrm{sim}(f_{\boldsymbol{\theta}_n}^{n}(x),\mathbf{t}_c)/\tau \right) } \right], 
\end{equation}
where $x$ and $y$ denote a local input sample and its corresponding class label, respectively, $\mathrm{sim}(\cdot,\cdot)$ denotes cosine similarity, and $\tau$ is a temperature parameter. The loss in Eq.~\eqref{eq:stage1} corresponds to the InfoNCE objective~\cite{oord2018representation}. We employ InfoNCE because it encourages embeddings to align with their corresponding class anchors while remaining separated from anchors of other classes. This promotes both intra-class compactness and inter-class separability~\cite{wang2020understanding}. As a result, the local encoder learns to project samples into $\mathcal{R}$ while preserving the semantic structure induced by the FM. In particular, for a sample-label pair $(x,y)$, the optimization objective encourages 
\begin{eqnarray}
f_{\boldsymbol{\theta}_n}^{\,n}(x)\approx \mathbf{t}_y.\nonumber
\end{eqnarray}
After training, edge device $u_n$ encodes its local dataset into an embedding matrix $\mathbf{H}_n \in \mathbb{R}^{|\mathcal{D}_n|\times d},$ whose $i^{\text{th}}$ row is given by $\mathbf{h}_i=f_{\boldsymbol{\theta}_n}^{\,n}(x_i)\in\mathcal{R}$.
Since all local encoders map their data into the same reference space, the embeddings in $\mathbf{H}_n$ provide a consistent representation of the task-specific knowledge contained in $\mathcal{D}_n$. Consequently, each edge device transmits $\mathbf{H}_n$ to the server instead of sharing raw data or model parameters.

\subsection{FM Adaptation}
After the reference-space construction stage, the server receives the embedding matrices $\{\mathbf{H}_n\}_{n=1}^{N}$ from all edge devices. Collectively, these embeddings represent the distributed task knowledge in the shared reference space. The objective of the FM adaptation stage is to use this knowledge to adapt the server-hosted FM. However, the received embeddings provide supervision only in the reference space, while the corresponding input samples are unavailable at the server. Consequently, they cannot be directly used for FM adaptation. To bridge this gap, we reconstruct synthetic samples whose embeddings are consistent with the received embeddings. 

Directly optimizing input samples from embeddings, however, often produces unrealistic solutions as multiple inputs can correspond to similar embeddings~\cite{mahendran2015understanding}. Hence, we introduce a generative prior by assuming that the server has access to a pretrained text-to-image generator $G(\cdot)$~\cite{diffusion}. Given a prompt $\mathbf{p}_i$, the generator produces a synthetic sample $\hat{x}_i=G(\mathbf{p}_i)$. Each prompt is parameterized as
\begin{eqnarray}
\mathbf{p}_i=[\mathcal{T}(c_{y_i}),\boldsymbol{\xi}_i], \nonumber
\end{eqnarray}
where $\mathcal{T}(c_{y_i})$ denotes the embedding of a fixed class-specific textual prompt associated with class $y_i$, and $\boldsymbol{\xi}_i$ is a trainable latent vector. The prompt embedding preserves class semantics, while $\boldsymbol{\xi}_i$ captures intra-class variability and enables the generation of diverse samples.

Here, let
$
\mathbf{e}_i = F_{\boldsymbol{\phi}}^{v}(\hat{x}_i)
$
denote the embedding of a generated sample. The latent variables $\{\boldsymbol{\xi}_i\}$ are learned by minimizing the reconstruction loss as follows
\begin{equation}
\label{eq:recon}
\mathcal{L}_{\text{recon}}
=
\mathbb{E}_{i}
\left[
-\log
\frac{
\exp\!\left(
\mathrm{sim}(\mathbf{e}_i,\mathbf{h}_i)/\tau
\right)
}
{
\sum_{j}
\exp\!\left(
\mathrm{sim}(\mathbf{e}_i,\mathbf{h}_j)/\tau
\right)
}
\right],
\end{equation}
where $\mathbf{h}_i$ denotes the target embedding received from the edge devices. We again employ InfoNCE as it preserves the relative neighborhood structure of the reference space rather than enforcing exact embedding inversion. This is particularly desirable since multiple input samples can correspond to similar embeddings, making the reconstruction problem inherently ill-posed~\cite{wang2020understanding}.

After optimizing $\{\boldsymbol{\xi}_i\}$, the server obtains a synthetic dataset
$
\hat{\mathcal{D}}=\{(\hat{x}_i,y_i)\},
$
whose samples are consistent with the aggregated embeddings while remaining faithful to the semantic structure of the reference space. The reconstructed dataset converts the representation-level knowledge received from edge devices into input-space supervision that can be directly used for FM adaptation. Since the downstream task is image classification, only the visual encoder of the FM is adapted using LoRA~\cite{hu2022lora}. Now, let $\Delta_{\boldsymbol{\phi}}$ denote the trainable LoRA parameters associated with the visual encoder. Then, these parameters are learned by minimizing the following loss
\begin{equation}
\label{eq:fm}
\mathcal{L}_{\text{FM}}
=
\underset{(\hat{x},y)\sim\hat{\mathcal{D}}}{\mathbb{E}}
\Big[
\mathcal{L}_{\text{CE}}
\big(
F_{\boldsymbol{\phi}+\Delta_{\boldsymbol{\phi}}}^{v}(\hat{x}),y
\big)
\Big],
\end{equation}
where $\mathcal{L}_{\text{CE}}$ denotes the cross-entropy loss. To enable deployment on edge devices, the adapted FM is distilled into a global model $f_{\boldsymbol{\theta}_g}(\cdot)$. The global-model parameters $\boldsymbol{\theta}_g$ are learned by minimizing
\begin{equation}
\label{eq:kd}
\mathcal{L}_{\text{KD}}
=
\underset{(\hat{x},y)\sim\hat{\mathcal{D}}}{\mathbb{E}}
\Big[
\mathcal{L}_{\text{CE}}
\big(
f_{\boldsymbol{\theta}_g}(\hat{x}),y
\big)
+
\lambda
\,\mathrm{KL}
\big(
f_{\boldsymbol{\theta}_g}(\hat{x})
\,\|\,
F_{\boldsymbol{\phi}+\Delta_{\boldsymbol{\phi}}}^{v}(\hat{x})
\big)
\Big],
\end{equation}
where $\mathrm{KL}(\cdot\|\cdot)$ denotes the Kullback--Leibler divergence and $\lambda$ controls the distillation strength. The resulting global model combines the task-specific knowledge encoded in the edge device embeddings with the general visual representations of the FM. Unlike the FM, the global model is intentionally designed to be lightweight and compatible with the resource limitations of all edge devices, enabling practical deployment across the entire edge network.

\subsection{Global Knowledge Utilization}
The objective of this stage is to transfer the knowledge distilled into the global model back to edge devices and adapt it to the target deployment scenario. Depending on the desired objective, the global model can be utilized for either \emph{personalization} or \emph{generalization}.

\vspace{.05in}
\noindent
\textbf{\textit{Personalization:}} In this setting, training and test samples on each device are drawn from the same local distribution, i.e., $\mathbb{D}_n\sim\mathcal{D}_n$ and $\mathbb{D}_n^{\text{test}}\sim\mathcal{D}_n$. Each device uses the global model as a frozen feature extractor and learns a lightweight classifier head $\hbar_{\boldsymbol{\psi}_n}(\cdot)$. Freezing the global model preserves the globally learned representation while reducing the risk of overfitting to limited local data. The classifier parameters $\boldsymbol{\psi}_n$ are learned by minimizing
\begin{equation}
\label{eq:personal}
\mathcal{L}_{\text{pers}}
=
\mathbb{E}_{(x,y)\sim\mathbb{D}_n}
\Big[
\mathcal{L}_{\text{CE}}
\big(
\hbar_{\boldsymbol{\psi}_n}
\big(f_{\boldsymbol{\theta}_g}(x)\big),y
\big)
\Big].
\end{equation}
This allows each device to adapt the decision layer to its local distribution while preserving the shared representation learned from the aggregated knowledge of all edge devices.

\vspace{.05in}
\noindent
\textbf{\textit{Generalization:}} In this setting, evaluation is performed on a balanced test distribution $\mathcal{D}_{\text{bal}}$ constructed by uniformly sampling across classes. Unlike personalization, which seeks to optimize performance on a device-specific distribution, generalization aims to obtain a model that performs well across a balanced population-level distribution. To achieve this objective, each device initializes its local model using the global-model parameters, i.e., $\boldsymbol{\theta}_n \leftarrow \boldsymbol{\theta}_g,$ augments it with a trainable classifier head, and performs local adaptation. Since local datasets may exhibit significant class imbalance, we employ logit-adjusted cross-entropy to compensate for class-frequency bias during adaptation. The classifier parameters $\boldsymbol{\psi}_n$ and model parameters $\boldsymbol{\theta}_n$ are learned by minimizing
\begin{equation}
\label{eq:general}
\mathcal{L}_{\text{gen}}
=
\mathbb{E}_{(x,y)\sim\mathbb{D}_n}
\Big[
\mathcal{L}_{\text{LA}}
\big(
\hbar_{\boldsymbol{\psi}_n}
\big(f_{\boldsymbol{\theta}_n}(x)\big),y
\big)
\Big],
\end{equation}
where $\mathcal{L}_{\text{LA}}$ denotes the logit-adjusted cross-entropy loss. This adaptation step aligns the global model with the local task distribution observed at each device while mitigating the effect of class imbalance. Through this process, the knowledge acquired during FM adaptation is transferred to edge devices through a server-to-edge device model broadcast, enabling either personalized or generalized deployment without requiring further server-edge device communication. The overall framework is summarized in Algorithm~\ref{alg:df-FKT}.

\begin{algorithm}[t]
\caption{Embedding-Guided Data-Free FKT (ECLIPSE)}
\label{alg:df-FKT}
\begin{algorithmic}[1]

\State \textbf{Server:}
$\mathbf{T}=\{\mathbf{t}_c\}_{c=1}^{C}$,
$F_{\boldsymbol{\phi}}=\{F^v_{\boldsymbol{\phi}},F^t_{\boldsymbol{\phi}}\}$,
$G(\cdot)$

\State Broadcast $\mathbf{T}$

\ForAll{$u_n\in\mathbb{U}$ \textbf{in parallel}}

    \State
    $\boldsymbol{\theta}_n
    \leftarrow
    \arg\min_{\boldsymbol{\theta}_n}
    \mathcal{L}_{\text{align}}$

    \State
    $\mathbf{H}_n
    =
    f_{\boldsymbol{\theta}_n}^{\,n}(\mathcal{D}_n)$

    \State
    Upload $\mathbf{H}_n$

\EndFor

\State
$\mathbf{H}
=
\bigcup_{n=1}^{N}\mathbf{H}_n$

\State
Initialize $\{\boldsymbol{\xi}_i\}$

\Repeat

    \State
    $\mathbf{p}_i
    =
    [\mathcal{T}(c_{y_i}),\boldsymbol{\xi}_i]$

    \State
    $\hat{x}_i
    =
    G(\mathbf{p}_i)$

    \State
    $\mathbf{e}_i
    =
    F_{\boldsymbol{\phi}}^{v}(\hat{x}_i)$

    \State
    $\{\boldsymbol{\xi}_i\}
    \leftarrow
    \arg\min_{\{\boldsymbol{\xi}_i\}}
    \mathcal{L}_{\text{recon}}$

\Until{convergence}

\State
$\hat{\mathcal{D}}
=
\{(\hat{x}_i,y_i)\}$

\State
$\Delta_{\boldsymbol{\phi}}
\leftarrow
\arg\min_{\Delta_{\boldsymbol{\phi}}}
\mathcal{L}_{\text{FM}}$

\State
$\boldsymbol{\theta}_g
\leftarrow
\arg\min_{\boldsymbol{\theta}_g}
\mathcal{L}_{\text{KD}}$

\State
Broadcast $f_{\boldsymbol{\theta}_g}$

\ForAll{$u_n\in\mathbb{U}$}

    \If{Personalization}

        \State
        $\boldsymbol{\psi}_n
        \leftarrow
        \arg\min_{\boldsymbol{\psi}_n}
        \mathcal{L}_{\text{pers}}$

    \Else

        \State
        $\boldsymbol{\theta}_n
        \leftarrow
        \boldsymbol{\theta}_g$

        \State
        $\{\boldsymbol{\theta}_n,\boldsymbol{\psi}_n\}
        \leftarrow
        \arg\min_{\boldsymbol{\theta}_n,\boldsymbol{\psi}_n}
        \mathcal{L}_{\text{gen}}$

    \EndIf

\EndFor

\end{algorithmic}

\end{algorithm}
\vspace{-.1in}
\subsection{Theoretical Analysis}
To establish theoretical guarantees for the proposed framework, first, utilized analysis assumptions are introduced below:
\begin{itemize}
\item[\textbf{A1.}] The cross-entropy loss is Lipschitz continuous with respect to the embedding distribution.
\item[\textbf{A2.}] The visual encoder $F_{\boldsymbol{\phi}}^{v}(\cdot)$ is $L_f$-Lipschitz continuous.
\item[\textbf{A3.}] The reference-space alignment process introduces a bounded alignment error $\epsilon_a$ for all edge devices, i.e.,
\begin{equation}
W_2(\tilde P_n,P_n)\le \epsilon_a.
\end{equation}
\item[\textbf{A4.}] The reconstruction process introduces a bounded reconstruction error $\epsilon_r$, i.e.,
\begin{equation}
W_2(\hat P,\tilde P)\le \epsilon_r.
\end{equation}
\item[\textbf{A5.}] Reconstructed surrogate dataset consists of $m$ samples independently drawn from the reconstructed distribution, such that standard finite-sample concentration bounds~apply.
\end{itemize}
To analyze the effect of the reference-space construction stage, average heterogeneity in the original data space is defined~as
\begin{equation}
\Delta_{\mathrm{raw}}
=
\frac{1}{N}
\sum_{n=1}^{N}
W_2(\mathcal D_n,\mathcal D),
\end{equation}
where $N$ denotes the number of edge devices, $\mathcal D_n$ denotes the local data distribution of edge device $u_n$, and $\mathcal D$ denotes the global data distribution. The corresponding heterogeneity in the aligned embedding space is defined as
\begin{equation}
\Delta_{\mathrm{emb}}
=
\frac{1}{N}
\sum_{n=1}^{N}
W_2(\tilde P_n,\tilde P),
\end{equation}
where $\tilde P_n$ and $\tilde P$ denote the aligned local and global joint embedding-label distributions, respectively. Let $P_n$ denote the joint embedding-label distribution induced by mapping samples from $\mathcal D_n$ through the visual encoder $F_{\boldsymbol{\phi}}^{v}(\cdot)$, and let $P^\star$ denote the corresponding global joint embedding-label distribution induced by the global data distribution $\mathcal D$.

\vspace{.1in}
\noindent
\textbf{Proposition 1 - Reference-Space Alignment Controls Residual Heterogeneity:} 
Under assumptions \textbf{A2}--\textbf{A3},
\begin{equation}
\Delta_{\mathrm{emb}}
\le
L_f \Delta_{\mathrm{raw}}
+
2\epsilon_a,
\label{eq:heterogeneity_bound}
\end{equation}
where $L_f$ is the Lipschitz constant of the visual encoder and $\epsilon_a$ denotes the alignment error.

\vspace{.1in}
\noindent
\textit{Proof.}
Since the visual encoder $F_{\boldsymbol{\phi}}^{v}(\cdot)$ is $L_f$-Lipschitz,
\begin{equation}
W_2(P_n,P^\star)
\le
L_f\,W_2(\mathcal D_n,\mathcal D).
\end{equation}
Applying the triangle inequality yields
\begin{equation}
W_2(\tilde P_n,\tilde P)
\le
W_2(\tilde P_n,P_n)
+
W_2(P_n,P^\star)
+
W_2(P^\star,\tilde P).
\end{equation}
From assumption \textbf{A3},
\begin{equation}
W_2(\tilde P_n,P_n)
\le
\epsilon_a.
\end{equation}
Moreover, since
\begin{equation}
\tilde P
=
\frac{1}{N}
\sum_{n=1}^{N}
\tilde P_n,
\end{equation}
averaging the alignment errors yields
\begin{equation}
W_2(P^\star,\tilde P)
\le
\epsilon_a.
\end{equation}
Substituting the above bounds into the triangle inequality and averaging over all edge devices yields
\begin{equation}
\Delta_{\mathrm{emb}}
\le
L_f\Delta_{\mathrm{raw}}
+
2\epsilon_a.
\end{equation}
\hfill\(\square\)

Proposition~1 shows that the residual heterogeneity after aggregation is bounded by the original data heterogeneity and the alignment error. Consequently, when the alignment error is small, knowledge from different edge devices becomes more closely aligned in the shared embedding space than in the original data space. We next analyze the reconstruction stage.

\vspace{.2in}
\noindent
\textbf{Proposition 2 - Reconstruction Consistency:} Under assumptions \textbf{A3}--\textbf{A4},
\begin{equation}
W_2(\hat P,P^\star)
\le
\epsilon_r+\epsilon_a,
\label{eq:reconstruction_bound}
\end{equation}
where $\hat P$ denotes the joint embedding-label distribution induced by the reconstructed surrogate dataset.

\vspace{.1in}
\noindent
\textit{Proof.}
Applying the triangle inequality,
\begin{equation}
W_2(\hat P,P^\star)
\le
W_2(\hat P,\tilde P)
+
W_2(\tilde P,P^\star).
\end{equation}
By \textbf{A3} and \textbf{A4}, the two terms are bounded by $\epsilon_a$ and $\epsilon_r$, respectively, yielding \eqref{eq:reconstruction_bound}.
\hfill\(\square\)

Proposition~2 shows that the reconstructed surrogate dataset preserves the global semantic distribution up to the alignment and reconstruction errors. To analyze FM adaptation, we define the ideal objective as
\begin{equation}
R(\boldsymbol{\phi})
=
\mathbb E_{(\mathbf z,y)\sim P^\star}
\Big[
L_{\mathrm{CE}}
\big(
F_{\boldsymbol{\phi}}^{v}(\mathbf z),y
\big)
\Big],
\end{equation}
and the surrogate objective as
\begin{equation}
\hat R(\boldsymbol{\phi})
=
\mathbb E_{(\mathbf z,y)\sim \hat P}
\Big[
L_{\mathrm{CE}}
\big(
F_{\boldsymbol{\phi}}^{v}(\mathbf z),y
\big)
\Big].
\end{equation}
The objective $R(\boldsymbol{\phi})$ corresponds to FM adaptation using the ideal semantic distribution, whereas $\hat R(\boldsymbol{\phi})$ corresponds to FM adaptation using the reconstructed surrogate dataset.

\vspace{.1in}
\noindent
\textbf{Proposition 3 - Surrogate-Risk Consistency:} Under assumptions \textbf{A1}--\textbf{A5}, we have
\begin{equation}
|R(\boldsymbol{\phi})-\hat R(\boldsymbol{\phi})|
\le
L(\epsilon_a+\epsilon_r)
+
\mathcal O(m^{-1/2}),
\label{eq:risk_consistency}
\end{equation}
where $L$ is the Lipschitz constant of the cross-entropy loss with respect to the embedding distribution and $m$ is the number of reconstructed samples.

\vspace{.1in}
\noindent
\textit{Proof.}
By assumption \textbf{A1},
\begin{equation}
|R(\boldsymbol{\phi})-\hat R(\boldsymbol{\phi})|
\le
L\,W_2(P^\star,\hat P).
\end{equation}

\noindent Substituting \eqref{eq:reconstruction_bound} gives
\begin{equation}
|R(\boldsymbol{\phi})-\hat R(\boldsymbol{\phi})|
\le
L(\epsilon_a+\epsilon_r).
\end{equation}

\noindent Adding the finite-sample concentration term from \textbf{A5} yields \eqref{eq:risk_consistency}.
\hfill\(\square\)

Proposition~3 shows that the objective optimized during FM adaptation remains close to the ideal objective whenever the alignment and reconstruction errors are small. We finally establish an end-to-end generalization guarantee for the proposed framework.

\vspace{.1in}
\noindent
\textbf{Theorem 1 - End-to-End Generalization Bound:} Let
\begin{equation}
\boldsymbol{\phi}^{\star}
=
\arg\min_{\boldsymbol{\phi}}
R(\boldsymbol{\phi})
\end{equation}
denote the FM adapted using the ideal semantic distribution, and let
\begin{equation}
\hat{\boldsymbol{\phi}}
=
\arg\min_{\boldsymbol{\phi}}
\hat R(\boldsymbol{\phi})
\end{equation}
denote the FM obtained through the proposed framework. Under assumptions \textbf{A1}--\textbf{A5},
\begin{equation}
R(\hat{\boldsymbol{\phi}})
-
R(\boldsymbol{\phi}^{\star})
\le
C_1\epsilon_a
+
C_2\epsilon_r
+
C_3\Delta_{\mathrm{emb}}
+
\mathcal O(m^{-1/2}),
\label{eq:main_bound}
\end{equation}
where $C_1$, $C_2$, and $C_3$ are positive constants.

\vspace{.1in}
\noindent
\textit{Proof.}
Using standard empirical-risk-minimization arguments,
\begin{equation}
R(\hat{\boldsymbol{\phi}})
-
R(\boldsymbol{\phi}^{\star})
=
\bigl(
R(\hat{\boldsymbol{\phi}})
-
\hat R(\hat{\boldsymbol{\phi}})
\bigr)
+
\bigl(
\hat R(\boldsymbol{\phi}^{\star})
-
R(\boldsymbol{\phi}^{\star})
\bigr).
\end{equation}
Applying \eqref{eq:risk_consistency} bounds the discrepancy between the ideal and surrogate objectives, while \eqref{eq:heterogeneity_bound} accounts for the residual heterogeneity in the aligned embedding space. Combining the resulting bounds yields \eqref{eq:main_bound}.
\hfill\(\square\)

Theorem~1 shows that the performance gap between the FM obtained through the proposed framework and the ideal FM adapted using the complete distributed knowledge is governed by four factors: the alignment error $\epsilon_a$, the reconstruction error $\epsilon_r$, the residual embedding-space heterogeneity $\Delta_{\mathrm{emb}}$, and the size of the reconstructed surrogate dataset $m$. The theorem indicates that improving reference-space alignment, reducing reconstruction error and residual heterogeneity, and increasing the number of reconstructed samples all contribute to improved FM adaptation.

\begin{table*}[t]
\centering
\caption{Performance comparison across datasets under personalization (P) and generalization (G) settings.}
\label{tab:main_results}
\resizebox{\textwidth}{!}{%
\begin{tabular}{c|c|cc|cc|cc|cc|cc|cc}
\hline
\textbf{Category} & \textbf{Method}
& \multicolumn{2}{c|}{\textbf{CIFAR-10}}
& \multicolumn{2}{c|}{\textbf{CIFAR-100}}
& \multicolumn{2}{c|}{\textbf{Imagenette}}
& \multicolumn{2}{c|}{\textbf{PlantVillage}}
& \multicolumn{2}{c|}{\textbf{EuroSAT}}
& \multicolumn{2}{c}{\textbf{PathMNIST}} \\
\cline{3-14}
& & G & P & G & P & G & P & G & P & G & P & G & P \\
\hline

\rowcolor{gray!15}
\textbf{Bounds}
& Local
& 38.01 & 60.20 & 8.51 & 19.10 & 60.04 & 62.22 & 55.37 & 69.91 & 70.01 & 90.40 & 77.85 & 87.80 \\

\rowcolor{gray!15}
& Centralized
& 74.25 & 75.63 & 53.52 & 55.27 & 70.01 & 75.09 & 79.19 & 82.24 & 90.01 & 99.27 & 88.93 & 93.61 \\

\rowcolor{gray!15}
& Zero-shot
& 87.60 & 87.73 & 59.32 & 59.60 & 90.25 & 87.71 & 15.82 & 9.43 & 20.13 & 30.26 & 11.52 & 10.20 \\

\rowcolor{gray!15}
& LoRA Fine-tuned
& 96.32 & 96.11 & 85.29 & 85.32 & 98.35 & 98.94 & 94.87 & 92.93 & 98.32 & 97.98 & 97.62 & 95.20 \\
\hline

\textbf{Standard FL}
& FedAvg~\cite{FedAvg}
& 38.79 & 44.98 & 15.84 & 20.73 & 62.41 & 64.07 & 59.52 & 71.48 & 73.84 & 91.62 & 79.11 & 88.63 \\

& FedProx~\cite{li2020federated}
& 41.05 & 45.53 & 17.31 & 21.65 & 63.95 & 65.82 & 61.17 & 72.64 & 75.22 & 92.11 & 80.06 & 89.02 \\

& SCAFFOLD~\cite{karimireddy2020scaffold}
& 43.82 & 47.63 & 19.76 & 23.02 & 65.59 & 67.05 & 63.84 & 74.29 & 77.03 & 92.98 & 81.95 & 89.83 \\
\hline

\textbf{Data-based FKT}
& FedMD~\cite{FedMD}
& 55.96 & 50.42 & 28.47 & 30.95 & 69.34 & 70.52 & 68.19 & 77.44 & 82.47 & 94.65 & \underline{84.15} & 88.73 \\

& FedDF~\cite{lin2020ensemble}
& 58.72 & 53.18 & 31.56 & 33.98 & 70.88 & 72.43 & 70.91 & 79.03 & 84.19 & 95.54 & 83.73 & 89.51 \\
\hline

\textbf{Data-free FKT}
& FedPKD~\cite{FedPKD}
& 33.70 & 40.17 & 14.81 & 18.22 & 60.93 & 63.31 & 57.16 & 69.93 & 71.04 & 90.96 & 78.66 & 88.01 \\
\hline

\textbf{Compressed-FM}
& Z2L
& 40.75 & 60.81 & 20.47 & 23.51 & 70.29 & 70.63 & 52.71 & 64.10 & 60.01 & 90.38 & 66.71 & 60.04 \\

& Open-Vocabulary~\cite{open-vocab}
& 52.61 & 67.82 & 34.91 & 36.18 & 73.80 & 74.06 & 75.64 & 84.91 & 85.73 & 96.03 & 82.94 & 89.14 \\

& SIDCLIP~\cite{SIDCLIP}
& 54.38 & \underline{69.55}
& 37.84 & 39.77
& 75.41 & \underline{76.22}
& \underline{77.58} & \underline{86.49}
& 87.12 & 96.85
& \textbf{84.73} & \textbf{91.02} \\
\hline

\textbf{Server-FM}
& FedDCL~\cite{FedDCL}
& \underline{60.53} & 58.71
& \underline{41.77} & \underline{43.63}
& \underline{76.52} & 75.10
& 71.84 & 79.37
& \underline{84.91} & \underline{95.28}
& 56.11 & 62.44 \\

& KOALA~\cite{KOALA}
& 51.48 & 43.29
& 40.92 & 42.66
& 72.88 & 73.95
& 69.92 & 77.31
& 82.73 & 94.41
& 81.62 & \underline{89.72} \\
\hline

\textbf{Embedding-Guided FMs}
& \text{ECLIPSE}
& \textbf{64.82} & \textbf{71.46}
& \textbf{44.73} & \textbf{47.18}
& \textbf{78.61} & \textbf{79.37}
& \textbf{74.63} & \textbf{81.15}
& \textbf{86.78} & \textbf{96.74}
&  54.76 & 61.93 \\
\hline

\end{tabular}%
}

\vspace{0.2cm}
\raggedright
\footnotesize{\textit{Note:} Rows highlighted in gray denote reference bounds (Local, Centralized, Zero-shot, and LoRA Fine-tuned) and are excluded from ranking. Bold values indicate the best performance among the compared methods, and underlined values indicate the second-best performance among the compared methods.}
\end{table*}

\section{Experiments} \label{sec:Exp}
In this section, we evaluate the proposed $\KM$ framework through extensive experiments covering classification performance, robustness, accuracy-energy tradeoff, and ablation analysis. The experiments evaluate the effectiveness of the proposed framework under diverse data distributions, model heterogeneity, and resource constraints.

\subsection{Experimental Setup} 
\subsubsection{Models}
We employ lightweight convolutional neural networks (CNNs) as local models on edge devices. Unless otherwise specified, the local model is a compact VGG-style architecture consisting of three convolutional blocks followed by a fully connected feature extractor. The learned features are mapped either to a task-specific classifier for standard FL baselines or to a 512-dimensional normalized embedding aligned with the FM. For CIFAR-100~\cite{cifar100}, we use ResNet-18 as the local model due to the increased task complexity. The server hosts the Contrastive Language–Image Pre-training (CLIP) model~\cite{CLIP2021} as the underlying FM. For surrogate data reconstruction, we adopt a pretrained Denoising Diffusion Probabilistic Model (DDPM)~\cite{diffusion}, specifically the Hugging Face checkpoint \texttt{stabilityai/sd-turbo} via the \texttt{diffusers} library.

\subsubsection{Datasets}
We evaluate the proposed framework on both generic and domain-specific image classification benchmarks. CIFAR-10~\cite{cifar100}, CIFAR-100~\cite{cifar100}, and Imagenette~\cite{imagenette} are used to evaluate the framework on natural image recognition tasks where CLIP exhibits strong zero-shot performance. PlantVillage~\cite{plantvillage}, EuroSAT~\cite{helber2017eurosat}, and PathMNIST~\cite{medmnist} are used to evaluate the proposed $\KM$ framework on domain-specific tasks where the pretrained FM shows limited zero-shot performance, therefore, requires effective task-specific adaptation. Among them, PathMNIST is particularly challenging as the server-hosted FM and the text-to-image generator are less aligned with the medical domain, making accurate synthetic data reconstruction more difficult.

\subsubsection{Baselines}
We compare the proposed framework against several categories of baselines covering different aspects of FL and model adaptation as outlined below:
\begin{itemize}
\item \textbf{Standard FL} (FedAvg~\cite{FedAvg}, FedProx~\cite{li2020federated}, and SCAFFOLD~\cite{karimireddy2020scaffold}): These methods rely on parameter aggregation, serving as classical FL baselines.
\item \textbf{Data-based FKT} (FedMD~\cite{FedMD} and FedDF~\cite{lin2020ensemble}): These approaches perform knowledge transfer using a shared public dataset, providing a direct comparison to our data-free formulation.
\item \textbf{Data-free FKT} (FedPKD~\cite{FedPKD} and FedDCL~\cite{FedDCL}): These methods perform knowledge transfer without public data by exchanging compact knowledge representations and aggregating them at the server.
\item \textbf{Server-FM Methods} (KOALA~\cite{KOALA}): This method deploys the FM on the server and uses local models to transfer task-specific knowledge for FM adaptation.
\item \textbf{Edge-FM} (FedCLIP~\cite{FedCLIP}, FedPrompt~\cite{FedPrompt}, and FedLoRA~\cite{FedLoRA}): These methods deploy and adapt the FM directly on edge devices. We include them only in the accuracy--energy tradeoff section to highlight the cost of local FM adaptation.
\item \textbf{Compressed-FM} (SIDCLIP~\cite{SIDCLIP} and Open-Vocabulary~\cite{open-vocab}): These methods distill FM knowledge into lightweight models without exploiting task-specific knowledge distributed across edge devices. Zero-shot to Local (Z2L) denotes direct CLIP-to-local-model distillation using only local data, without federated knowledge aggregation or FM adaptation.
\item \textbf{Bounds:} For reference, we additionally report the following four benchmark settings:
\begin{itemize}
\item \textit{Centralized,} trains a model using the global dataset and serves as an upper bound. 
\item \textit{Local,} trains independent local models without collaboration and serves as a lower bound. 
\item \textit{Zero-shot,} denotes the zero-shot performance of the FM evaluated directly on the global test dataset. 
\item \textit{LoRA Fine-tuned,} denotes the performance obtained by adapting the FM on the global training dataset using LoRA, providing an upper bound on task-specific FM adaptation.
\end{itemize}
\end{itemize}
%

\subsubsection{Implementation Details}
The experimental configuration is adapted to the characteristics of each dataset. For CIFAR-10~\cite{cifar100}, Imagenette~\cite{imagenette}, and PlantVillage~\cite{plantvillage}, we use $N=10$ edge devices, each with 1,000 local training samples and a VGG-style CNN with three convolutional blocks. For EuroSAT~\cite{helber2017eurosat} and PathMNIST~\cite{medmnist}, we use $N=20$ edge devices with 500 training samples per device and a lightweight single-block VGG-style CNN. For CIFAR-100~\cite{cifar100}, we consider a larger-scale federated setting with $N=50$ edge devices, 1,000 training samples per device, and ResNet-18 as the local model.
To model statistical heterogeneity, data partitions are generated using a Dirichlet distribution with concentration parameter $\alpha$~\cite{hsu2019measuring}. The setting $\alpha=100$ corresponds to IID data, whereas smaller values, including $\alpha \in \{1,0.1\}$, represent increasingly heterogeneous non-IID settings. 
Unless otherwise specified, we set $\alpha = 1$ throughout the experiments.
Local training is performed for one epoch with a batch size of 64. For iterative baselines, we use 50 communication rounds. To provide a fair comparison in terms of local optimization effort, the proposed framework trains each local model for 50 local epochs within its single communication round. For energy evaluation, each device transmits with power $p_n=1\,\mathrm{W}$. The total bandwidth is set to $B=20\,\mathrm{MHz}$ and is equally allocated among devices, and the noise power is fixed at $\sigma=-94\,\mathrm{dBm}$. The CPU frequency of each edge device, as a measure of its computational capacity, is chosen by randomly sampling from the range $[1.3,2.7]\,\mathrm{GHz}$~\cite{zhou2025flwireless}.

\subsection{Performance Analysis} \label{sec:Performance Analysis}
Table~\ref{tab:main_results} reports the performance of all compared methods under both personalization and generalization settings. Compared with conventional FL approaches, the proposed framework achieves substantially higher performance, suggesting that embedding-level knowledge transfer can be more effective than parameter aggregation under heterogeneous data distributions. Among the strongest baselines, FedDCL~\cite{FedDCL} and KOALA~\cite{KOALA} achieve competitive performance due to server-side FM utilization. Nevertheless, the proposed method consistently provides additional gains, suggesting that sample-level embedding transfer preserves richer task-specific information than prototype- or model-based knowledge transfer.
The improvements are most pronounced on CIFAR-10, CIFAR-100, and Imagenette, where the pretrained FM is well aligned with the underlying data distribution. In these cases, the reconstructed surrogate data enables effective FM adaptation and knowledge distillation.

On PlantVillage and EuroSAT, the proposed method remains superior to all baselines but exhibits a larger gap from the centralized upper bound than on natural-image benchmarks. This is expected since these datasets differ from the pretraining domain of the FM and diffusion model, making the embedding-to-image reconstruction process less accurate. Consequently, the generated surrogate data cannot fully capture the information contained in the original training samples, limiting the effectiveness of FM adaptation. PathMNIST presents an even more challenging scenario. Although the proposed framework remains competitive, its performance is lower than several compressed-FM and data-based baselines. This is mainly because the medical domain is poorly represented by both the pretrained FM and the text-to-image generator, resulting in synthetic samples that do not adequately capture the underlying data distribution. As a result, the reconstructed surrogate dataset provides weaker supervision for FM adaptation, highlighting a limitation of the proposed framework in highly specialized domains. Overall, the results indicate that the proposed $\KM$ framework effectively transfers distributed knowledge to a server-hosted FM in a data-free setting, while remaining robust across both natural-image and domain-specific tasks.

\noindent \textbf{\textit{Robustness to Statistical Heterogeneity:}} We evaluate robustness under non-IID data by partitioning data with a Dirichlet distribution ($\alpha \in \{0.01, 1, 100\}$), where smaller $\alpha$ implies stronger skew. Performance is evaluated on CIFAR-10 under both personalization and generalization settings. Table~\ref{tab:heterogeneity} illustrates that the proposed $\KM$ framework maintains stable performance across all heterogeneity levels, from highly non-IID ($\alpha=0.01$) to almost IID ($\alpha=100$). In contrast, conventional FL methods (such as FedAvg and FedProx) suffer significant performance degradation under strong heterogeneity, while data-based FKT methods exhibit moderate sensitivity~\cite{li2020federated}. Such degradation is mainly caused by edge device drift and inconsistent local updates induced by non-IID data distributions~\cite{karimireddy2020scaffold}. By comparison, $\KM$ demonstrates strong robustness to statistical heterogeneity and consistently outperforms all baselines, particularly in highly skewed settings. This robustness can be attributed to the use of a shared embedding-based reference space, which aligns heterogeneous local knowledge before aggregation, therefore, reducing inconsistency across edge devices. As a result, the method effectively mitigates the adverse effects of data heterogeneity and enables stable knowledge transfer even in highly non-IID scenarios.

\begin{table}[t]
\centering
\caption{Performance on CIFAR-10 under different data heterogeneity levels ($\alpha$-Dirichlet).}
\label{tab:heterogeneity}
\resizebox{\columnwidth}{!}{%
\begin{tabular}{c|cc|cc|cc}
\hline
\textbf{Method}
& \multicolumn{2}{c|}{$\alpha=0.01$}
& \multicolumn{2}{c|}{$\alpha=1$}
& \multicolumn{2}{c}{$\alpha=100$} \\
\cline{2-7}
& G & P & G & P & G & P \\
\hline
FedAvg     & 22.31 & 28.74 & 38.79 & 44.98 & 51.47 & 57.29 \\
FedProx    & 24.68 & 30.15 & 41.05 & 45.53 & 53.26 & 58.81 \\
SCAFFOLD   & 28.73 & 33.48 & 43.82 & 47.63 & 55.84 & 60.35 \\
\hline
FedMD      & 46.91 & 42.83 & 55.96 & 50.42 & 60.33 & 55.71 \\
FedDF      & 49.55 & 45.94 & 58.72 & 53.18 & 62.71 & 57.36 \\
\hline
Z2L         & 38.96 & 58.42 & 40.75 & 60.81 & 42.09 & 62.14 \\
Open-Vocab  & 49.43 & 64.75 & 52.61 & 67.82 & 54.97 & 69.38 \\
SIDCLIP     & 51.02 & 66.41 & 54.38 & 69.55 & 56.72 & 71.04 \\
FedDCL      & \underline{57.34} & \underline{56.09} & \underline{60.53} & \underline{58.71} & \underline{62.41} & \underline{60.62} \\
KOALA       & 48.62 & 41.90 & 51.48 & 43.29 & 53.85 & 45.17 \\
\hline
\text{ECLIPSE}
& \textbf{62.18} & \textbf{68.73}
& \textbf{64.82} & \textbf{71.46}
& \textbf{67.54} & \textbf{73.95} \\
\hline
\end{tabular}%
}
\end{table}

 \textbf{\textit{Robustness Under Model Heterogeneity:}} We evaluate the framework under heterogeneous local architectures by considering CNN models with varying depth. Specifically, edge devices are equipped with lightweight VGG-style networks composed of $2$, $3$, or $4$ convolutional blocks. Each device is randomly assigned one of these architectures to reflect variability in computational resources. Table~\ref{tab:heterogeneous_models} shows that all baselines experience performance degradation under heterogeneous settings due to inconsistencies across local model architectures. While KOALA is relatively more robust, it still suffers a noticeable performance drop. In contrast, the $\KM$ maintains stable performance under heterogeneous conditions. This robustness arises as local models are first aligned in a shared embedding reference space, which removes architectural dependencies before aggregation. Furthermore, during the deployment stage, all devices use the same distilled global model, eliminating discrepancies across architectures. As a result, $\KM$ effectively mitigates the impact of model heterogeneity and achieves more consistent performance compared to all baselines.

\begin{table}[t]
\centering
\caption{Performance with model heterogeneous and non-heterogeneous settings.}
\label{tab:heterogeneous_models}

\renewcommand{\arraystretch}{1.2} 
\setlength{\tabcolsep}{8pt} 

\begin{tabular}{c|cc|cc}
\hline
\textbf{Method} 
& \multicolumn{2}{c|}{non-Het} 
& \multicolumn{2}{c}{Het} \\
\cline{2-5}
& G & P & G & P \\
\hline

FedMD      & 59.21 & 53.84 & 55.96 & 50.42 \\
FedDF      & 62.08 & 56.31 & 58.72 & 53.18 \\
\hline
Z2L        & 42.11 & 62.44 & 40.75 & 60.81 \\
Open-Vocab & 54.86 & 69.95 & 52.61 & 67.82 \\
SIDCLIP    & 56.91 & 71.82 & 54.38 & 69.55 \\
FedDCL     & \underline{63.12} & \underline{61.41} & \underline{60.53} & \underline{58.71} \\
KOALA      & 53.74 & 45.97 & 51.48 & 43.29 \\
\hline

\text{ECLIPSE} & \textbf{61.77} & \textbf{60.42} & \textbf{58.84} & \textbf{59.21} \\
\hline
\end{tabular}
\end{table}

\begin{table}[t]
\centering
\caption{Characteristics used for the energy analysis.}
\label{tab:energy_method_characteristics}
\resizebox{\columnwidth}{!}{
\begin{tabular}{l|c|c|c|c}
\hline
\textbf{Method} & \textbf{Rounds} & \textbf{Upload} & \textbf{FLOPs/Epoch} & \textbf{Epochs/Round} \\
\hline
\text{ECLIPSE} & 1  & 3.09 MB  & 157.1G & 50 \\
FedAvg~\cite{FedAvg}    & 50 & 9.35 MB  & 157.1G & 1 \\
FedCLIP~\cite{FedCLIP}    & 50 & 9.02 MB  & 722.1G & 1 \\
FedLoRA~\cite{FedLoRA}    & 50 & 10.1 MB  & 722.1G & 1 \\
FedPrompt~\cite{FedPrompt}  & 50 & 0.064 MB & 722.1G & 1 \\
FedDCL~\cite{FedDCL}     & 50 & 0.02 MB  & 157.1G & 1 \\
\hline
\end{tabular}
}
\end{table}


\subsection{Accuracy--Energy Tradeoff in Federated Edge Networks}
We evaluate the accuracy--energy tradeoff of the proposed framework in resource-constrained federated edge networks. The comparison includes representative methods from three categories, i.e., conventional FL (FedAvg), Server-FM methods (FedDCL and KOALA), and Edge-FM methods (FedPrompt, FedCLIP, and FedLoRA). Following the edge-device energy model in \cite{belnet2018cortexa76}, the total energy consumption is computed as
\begin{equation}
E^{\mathrm{total}}
=
E^{\mathrm{comm}}
+
E^{\mathrm{comp}},
\end{equation}
where \(E^{\mathrm{comm}}\) and \(E^{\mathrm{comp}}\) denote communication and computation energy, respectively. In this regard, the uplink transmission rate of edge device \(u_n\) is given by
\begin{equation} 
r_n^{\mathrm{ul}} = b_n \log_2 \big( 1+ \frac{h_n p_n}{\sigma} \big), 
\end{equation}
where $b_n=B/N$, $h_n$, $p_n$, and $\sigma$ denote the bandwidth allocated to edge device $u_n$, channel gain, transmit power, and noise power, respectively. Subsequently, the corresponding communication energy is
\begin{equation}
E^{\mathrm{comm}}
=
\sum_{r=1}^{R}
\sum_{u_n\in\mathbb U}
p_n
\frac{S^{\mathrm{ul}}}
{r_n^{\mathrm{ul}}},
\end{equation}
where \(R\) and \(S^{\mathrm{ul}}\) denote the number of communication rounds and the number of uploaded bits per round. Moreover, the computation energy is calculated as
\begin{equation}
E^{\mathrm{comp}}
=
\sum_{r=1}^{R}
\sum_{u_n\in\mathbb U}
\epsilon_n f_n^2 C_n,
\end{equation}
where \(\epsilon_n = 10^{-28}\) J/cycle is the effective switched-capacitance coefficient, and \(f_n\) is the CPU frequency. Moreover, \(C_n=\frac{\chi_n}{\eta}\) is the number of required CPU cycles, while \(\chi_n\) denotes the required FLOPs, and \(\eta=16\) states the number of FLOPs executed per cycle~\cite{belnet2018cortexa76}. The characteristics used in the energy analysis are summarized in Table~\ref{tab:energy_method_characteristics}. Although $\KM$ operates in a single communication round, the computation energy accounts for all local computations performed during both the reference-space construction and local optimization stages.

\begin{figure}[H]
    \centering
    \includegraphics[width=0.9\columnwidth]{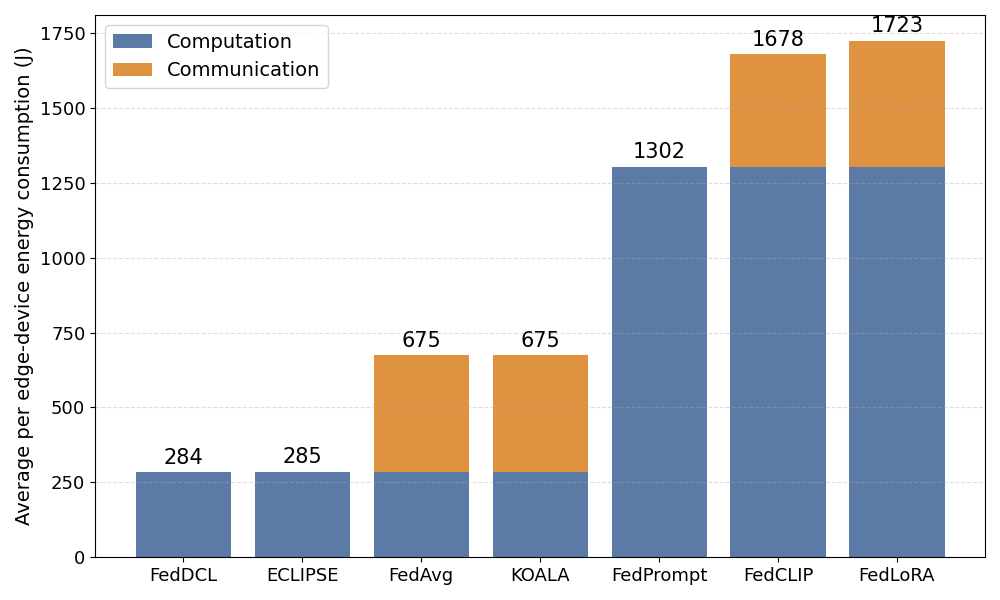}
    \caption{Average per edge-device energy consumption.}
    \label{fig:energy_comparison}
\end{figure}

Fig.~\ref{fig:energy_comparison} compares the average per edge-device energy consumption of the considered methods, together with its communication and computation components. Among the low-energy methods, FedDCL and the proposed framework require only \(283.6\) J and \(285.4\) J, respectively. In contrast, FedAvg and KOALA consume \(675.0\) J due to repeated communication rounds. Edge-FM methods incur substantially higher energy consumption, ranging from \(1302.5\) J to \(1723.4\) J, primarily because the FM must be deployed and adapted directly on edge devices.

\begin{figure}[H]
    \centering
    \includegraphics[width=0.9\columnwidth]{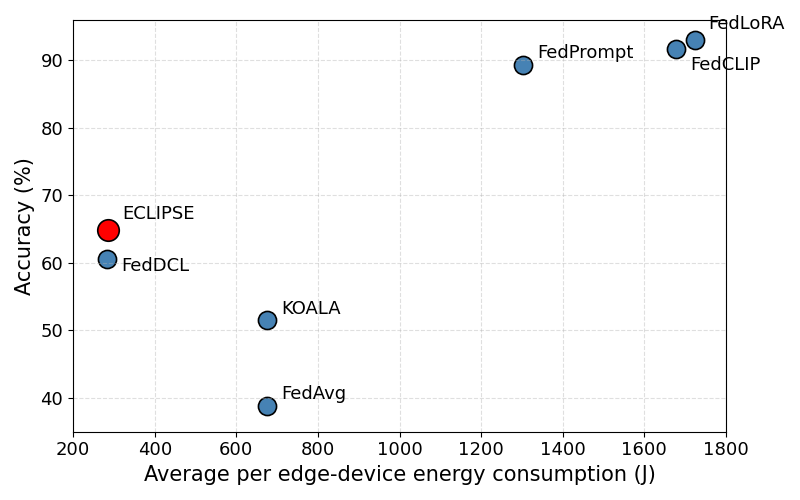}
    \caption{Accuracy versus average per edge-device energy consumption.}
    \label{fig:acc_energy_tradeoff}
\end{figure}

Fig.~\ref{fig:acc_energy_tradeoff} jointly evaluates classification accuracy and energy consumption on CIFAR-10 under the generalization setting. FedAvg achieves \(38.79\%\) accuracy while consuming \(675.0\) J. FedDCL reduces energy consumption to \(283.6\) J and achieves \(60.53\%\) accuracy. KOALA attains \(51.48\%\) accuracy at the same energy cost as FedAvg due to its iterative communication procedure. The proposed framework achieves the highest accuracy among all low-energy methods, reaching \(64.82\%\) while consuming only \(285.4\) J. Compared with KOALA, it improves accuracy by \(13.34\%\) while reducing energy consumption by approximately \(57.7\%\). Compared with FedDCL, it provides a \(4.29\%\) accuracy improvement with virtually identical energy consumption. Although Edge-FM methods achieve substantially higher accuracies (\(89.27\%\)--\(93.01\%\)), they require \(4.6\times\) to \(6.0\times\) more energy than the proposed framework. These results highlight the tradeoff between performance and edge-device resource consumption. By performing FM adaptation entirely at the server and transferring the adapted knowledge to lightweight edge models through a single-round knowledge transfer process, $\KM$ achieves a favorable accuracy--energy tradeoff among methods suitable for resource-constrained edge networks.

\begin{figure*}[t]
\centering

\textbf{CIFAR-10} \\
\begin{tabular}{ccccc}
\includegraphics[width=0.18\textwidth]{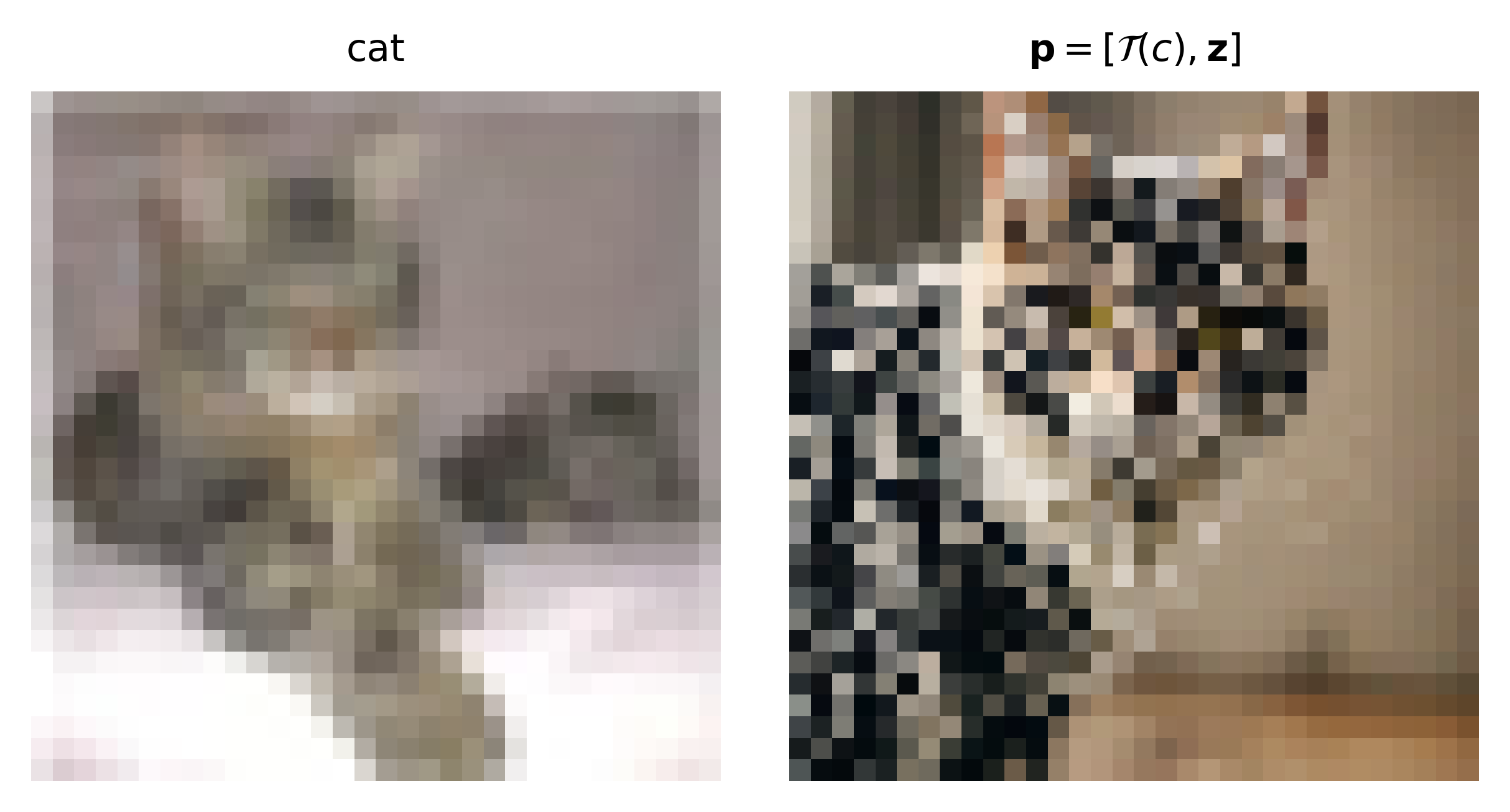} &
\includegraphics[width=0.18\textwidth]{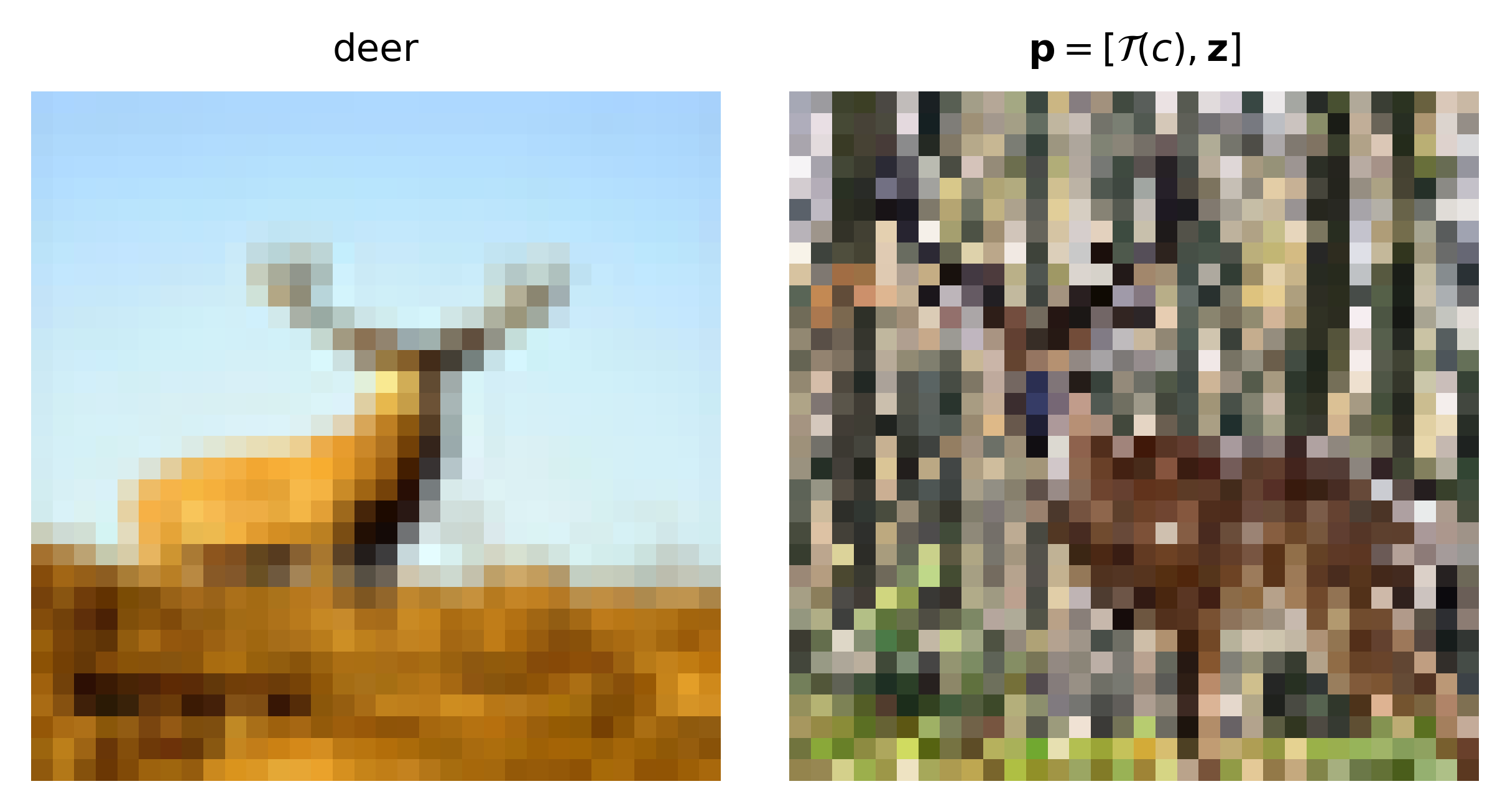} &
\includegraphics[width=0.18\textwidth]{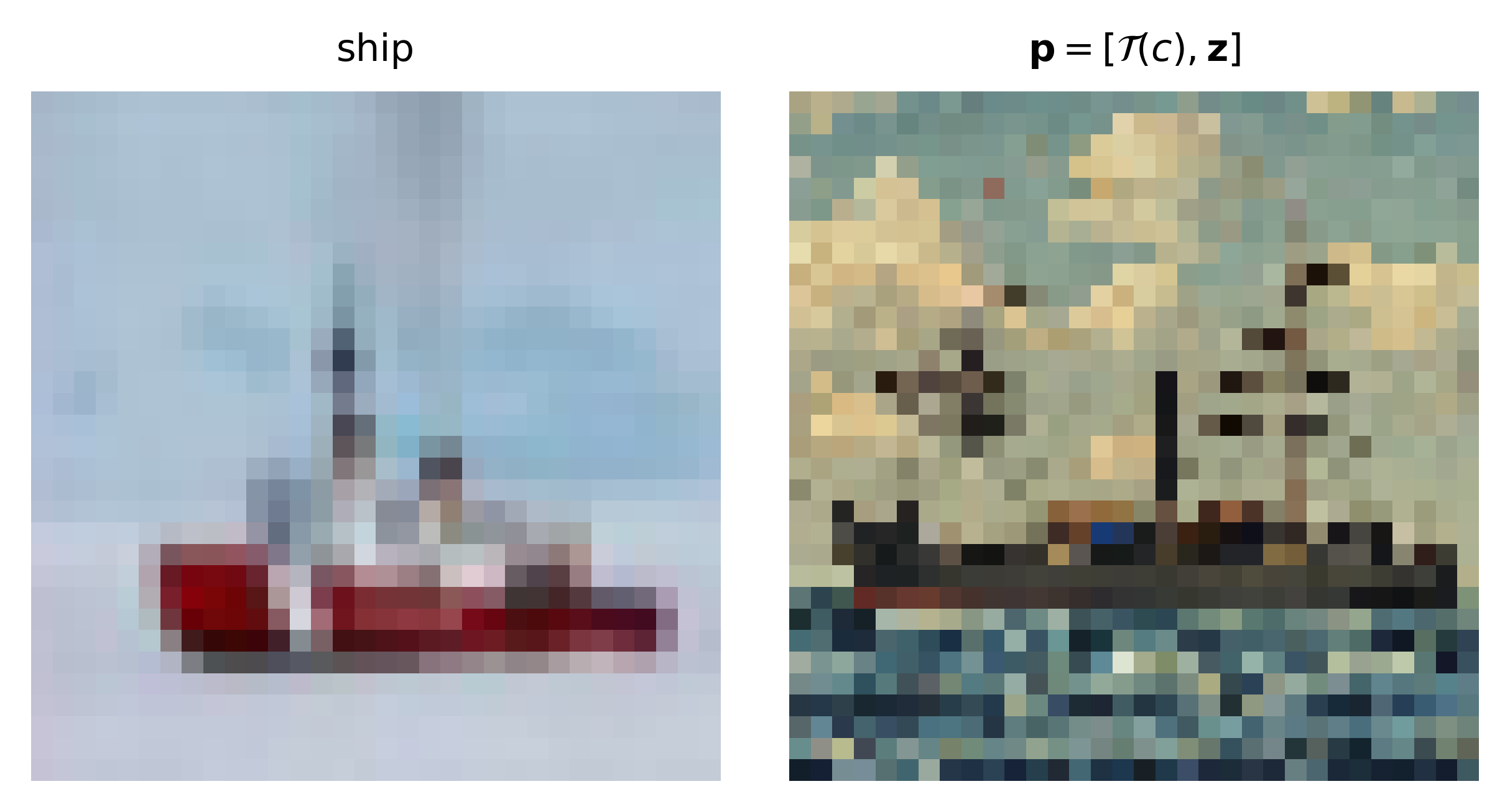} &
\includegraphics[width=0.18\textwidth]{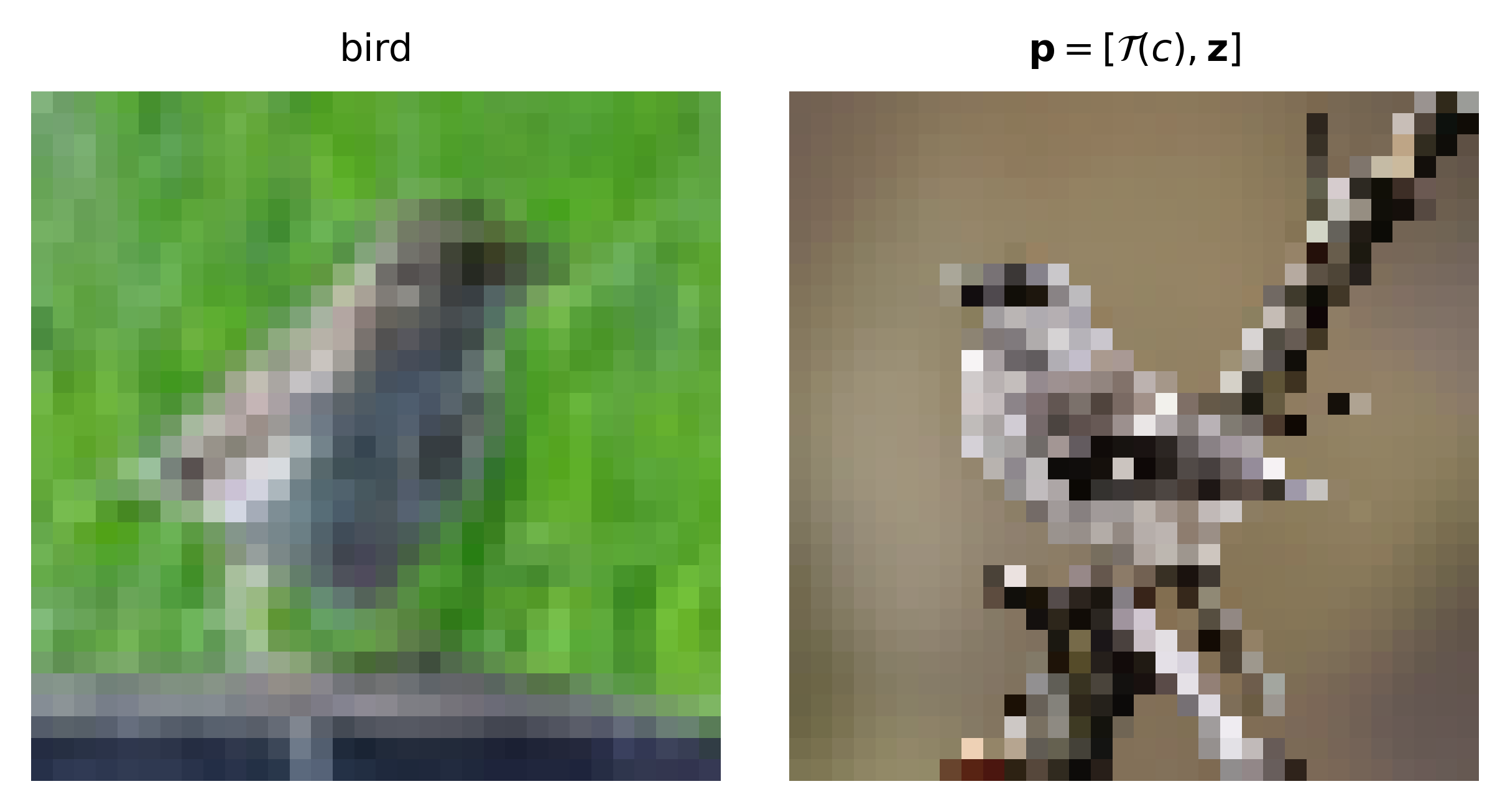} &
\includegraphics[width=0.18\textwidth]{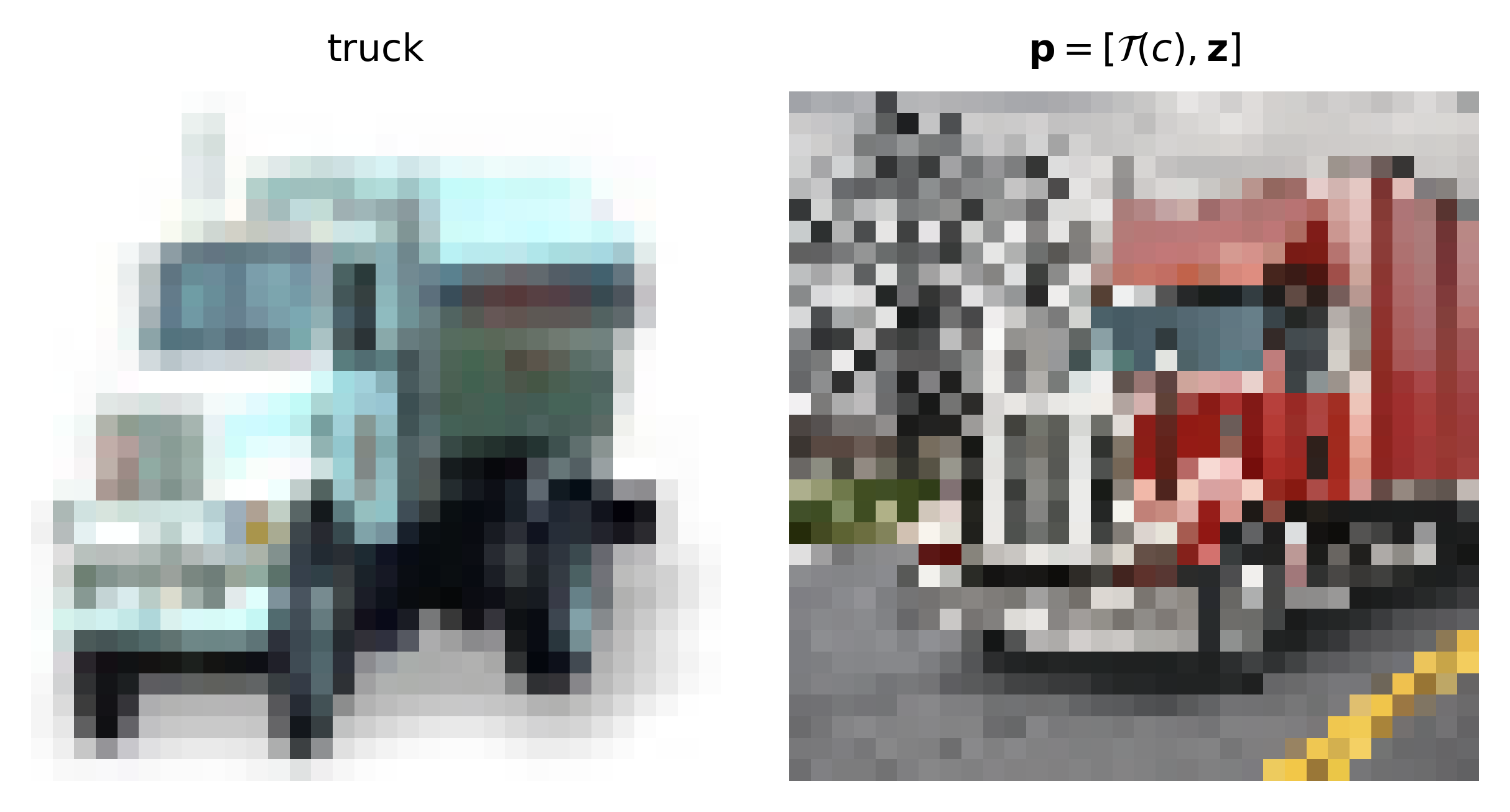} \\
\includegraphics[width=0.18\textwidth]{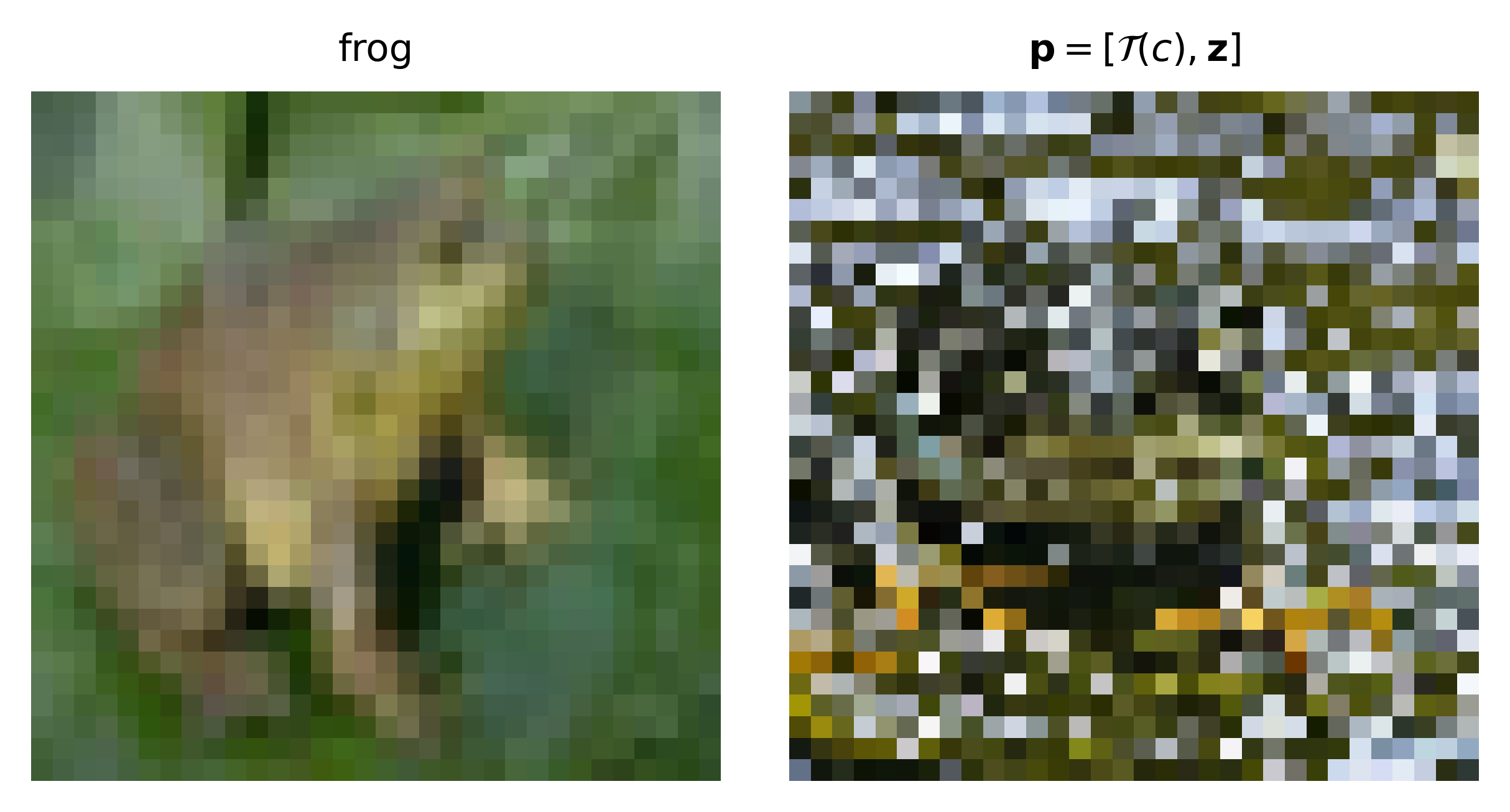} &
\includegraphics[width=0.18\textwidth]{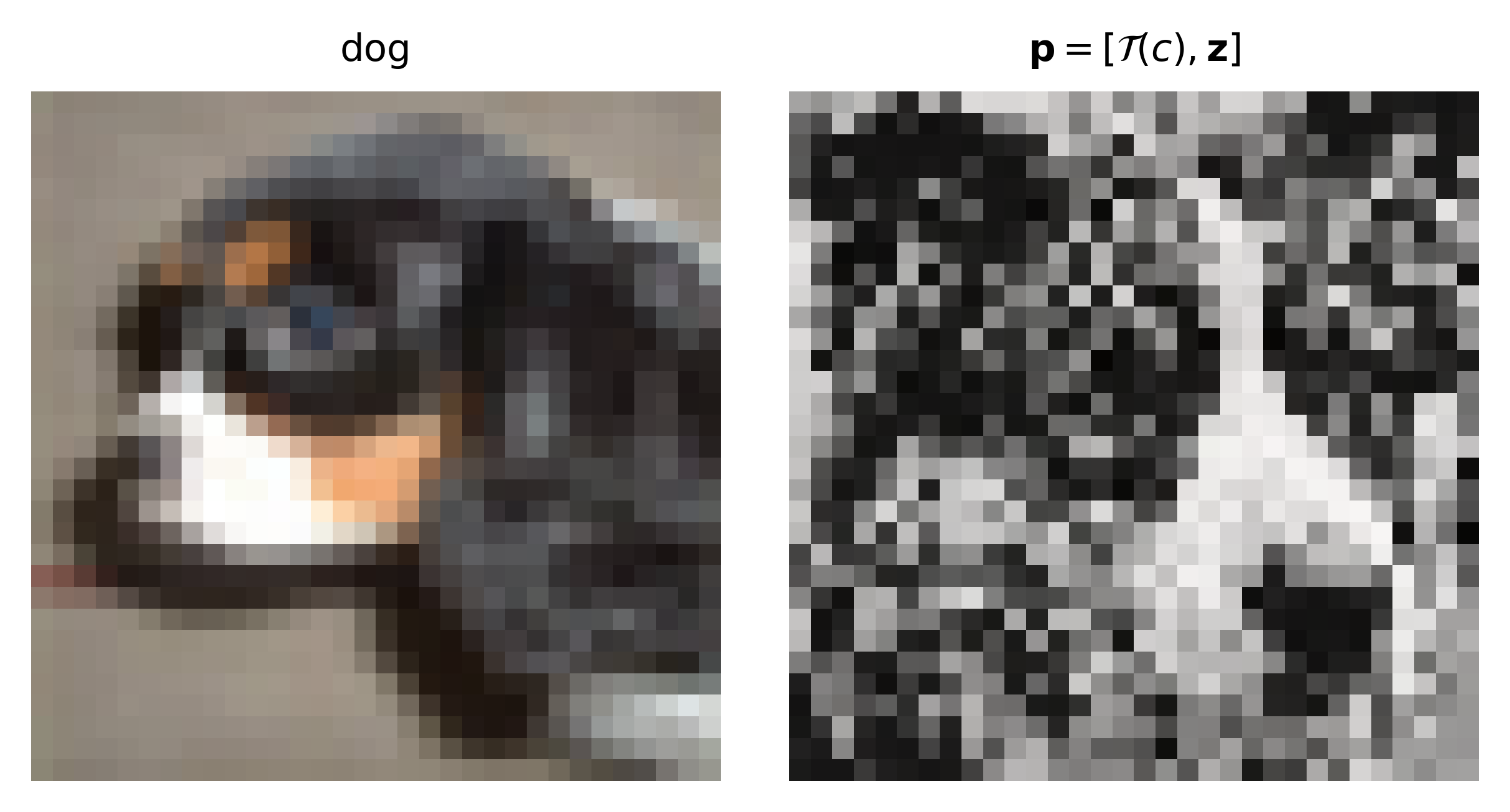} &
\includegraphics[width=0.18\textwidth]{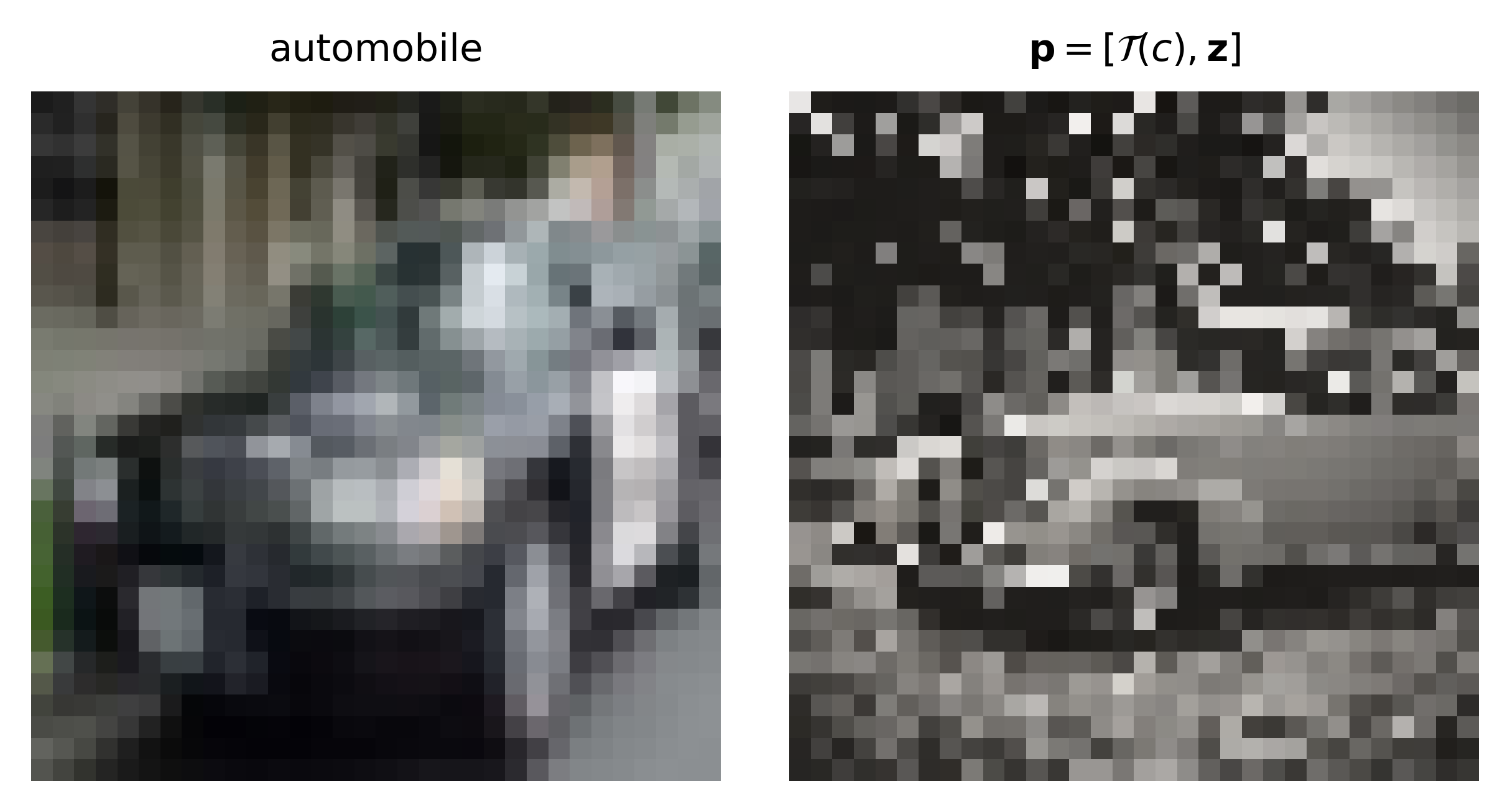} &
\includegraphics[width=0.18\textwidth]{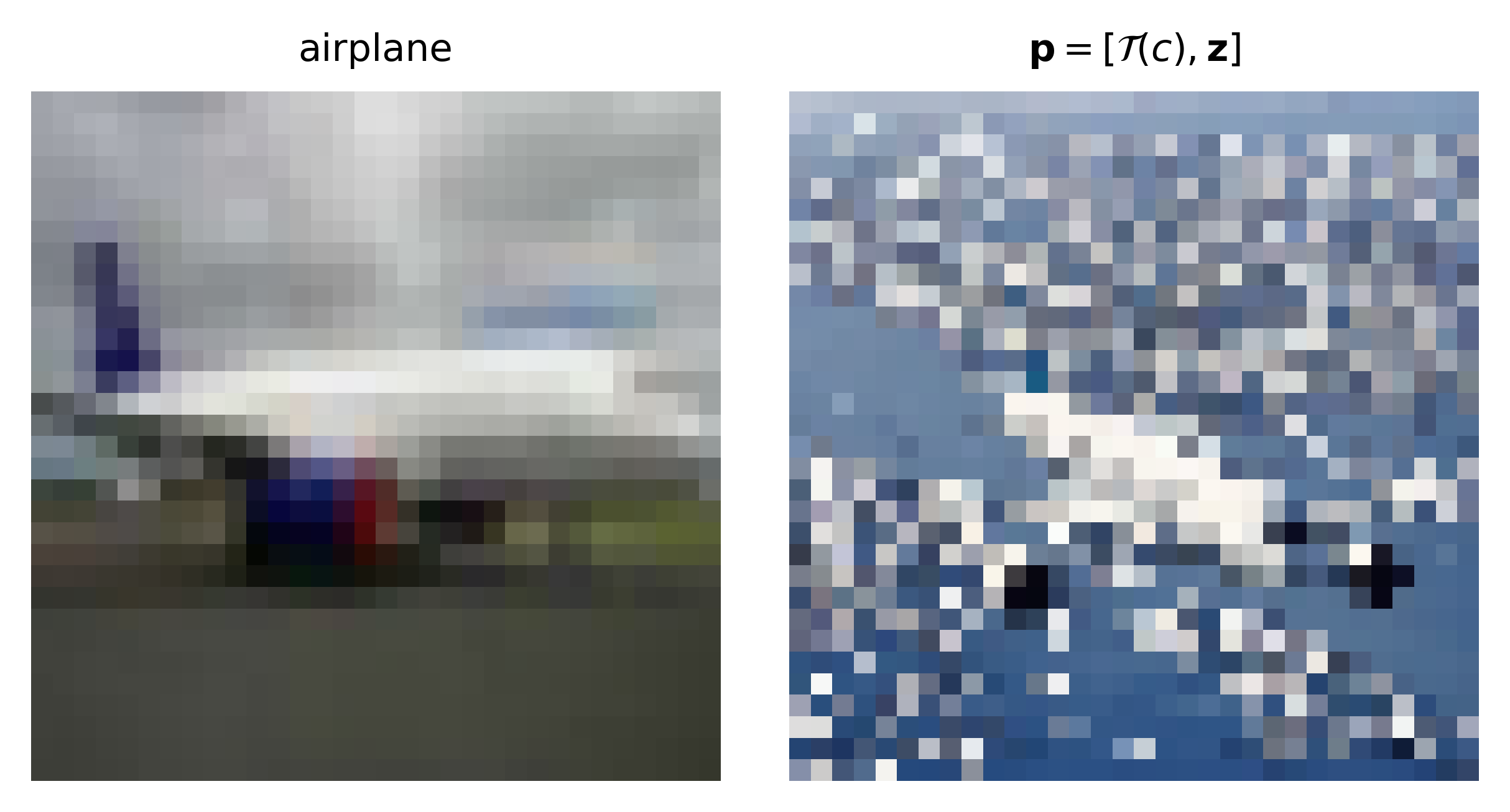} &
\includegraphics[width=0.18\textwidth]{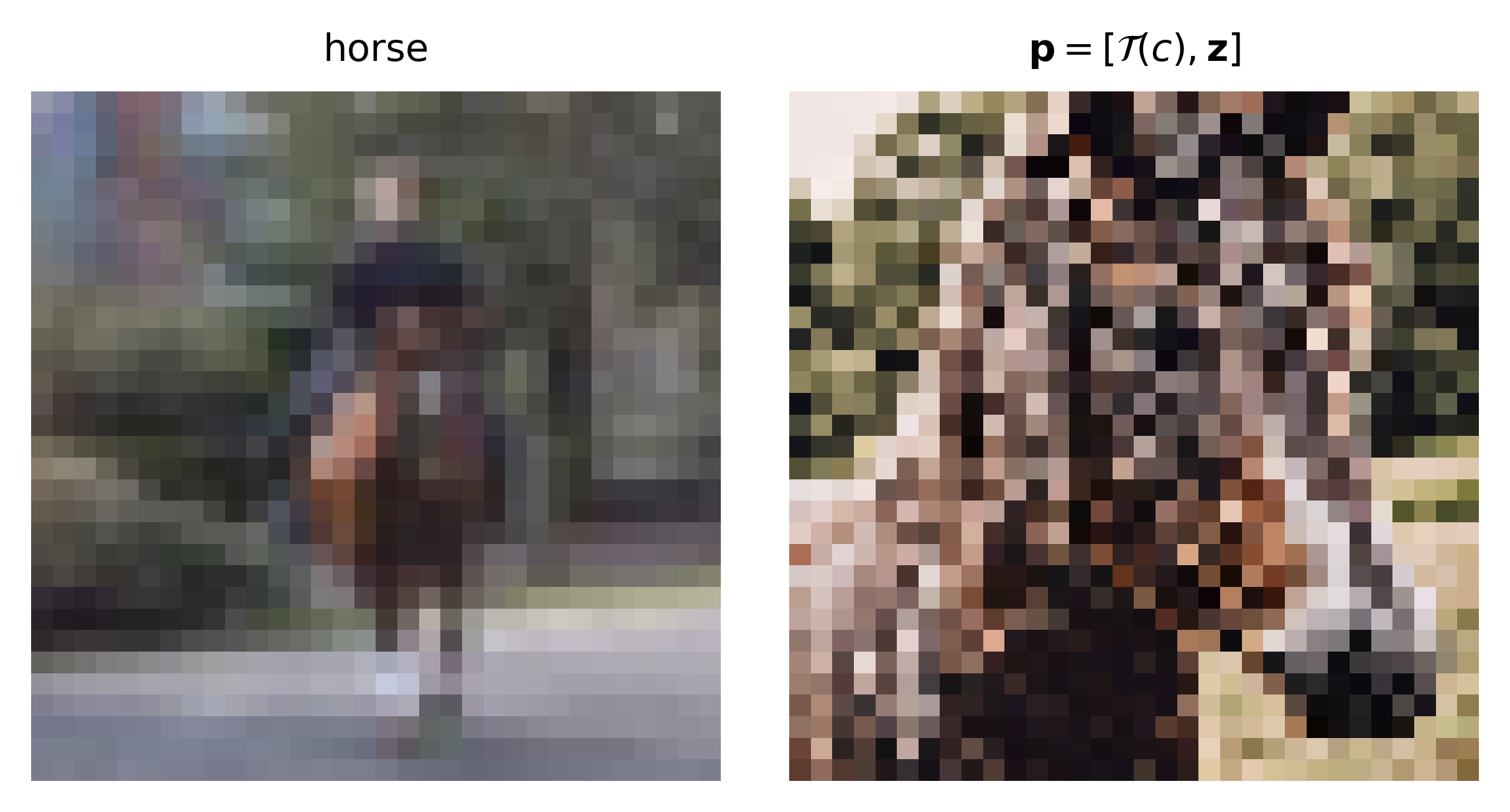}
\end{tabular}

\vspace{0.5em}
\textbf{EuroSAT} \\
\begin{tabular}{ccccc}
\includegraphics[width=0.18\textwidth]{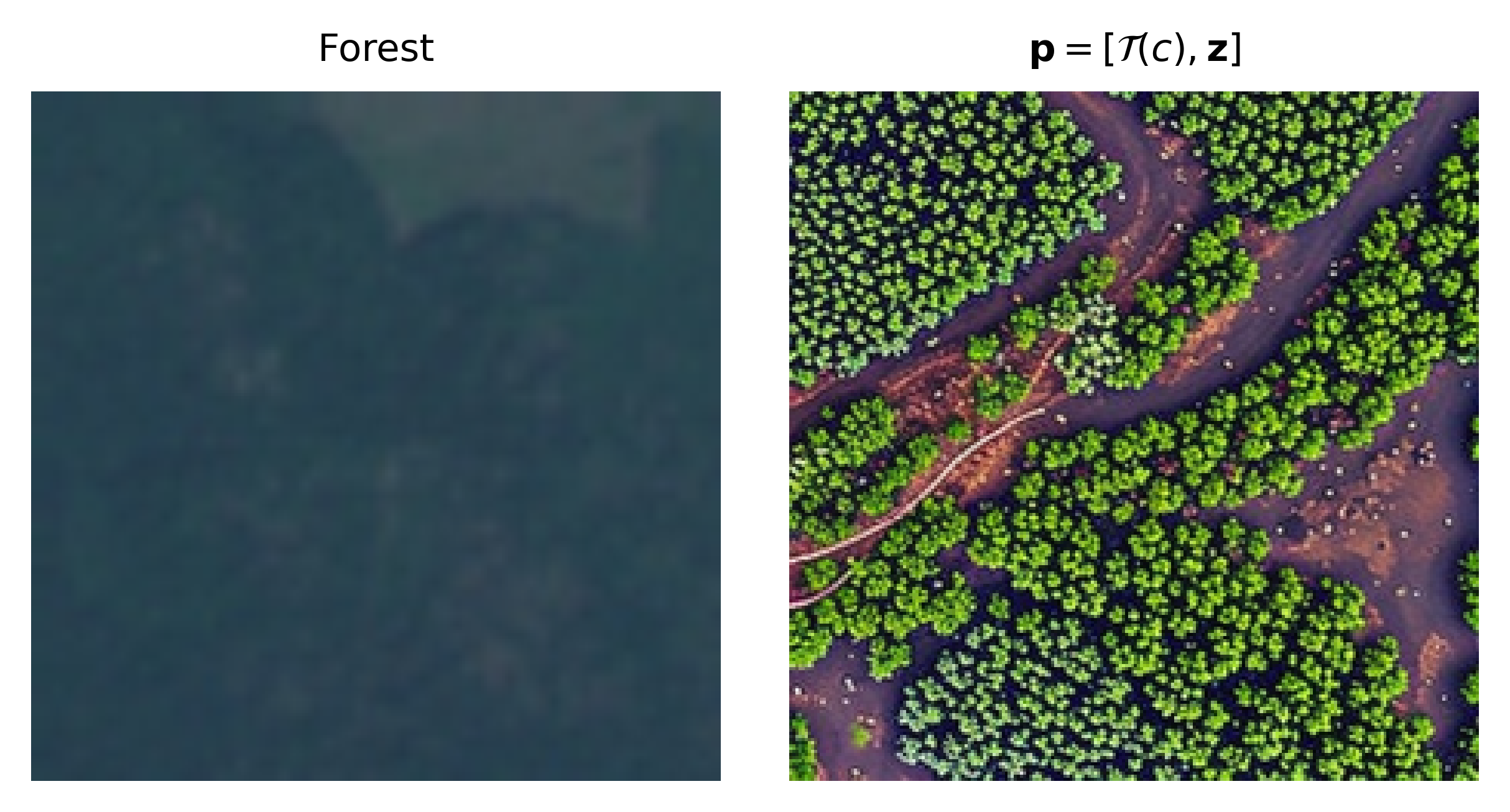} &
\includegraphics[width=0.18\textwidth]{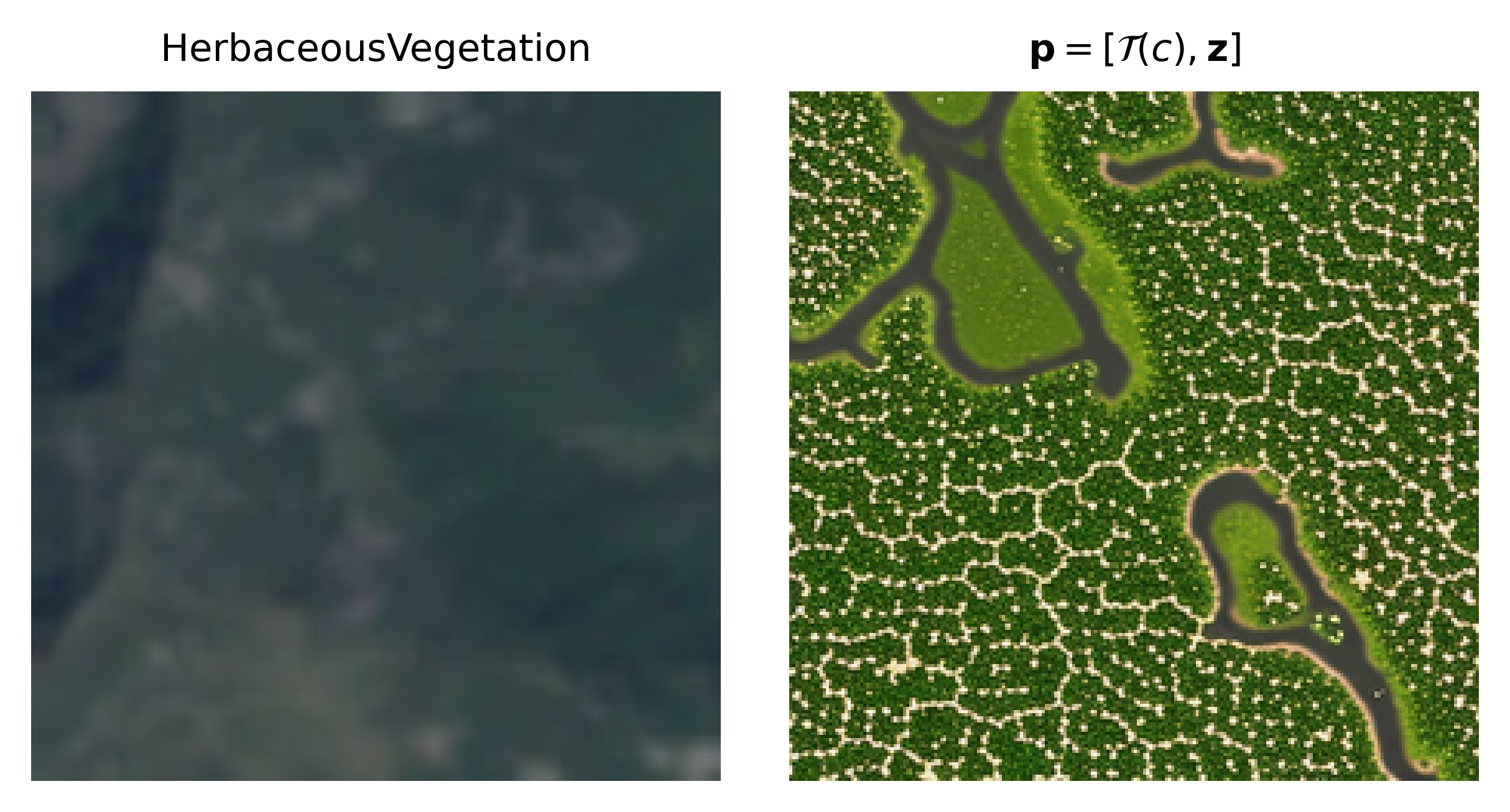} &
\includegraphics[width=0.18\textwidth]{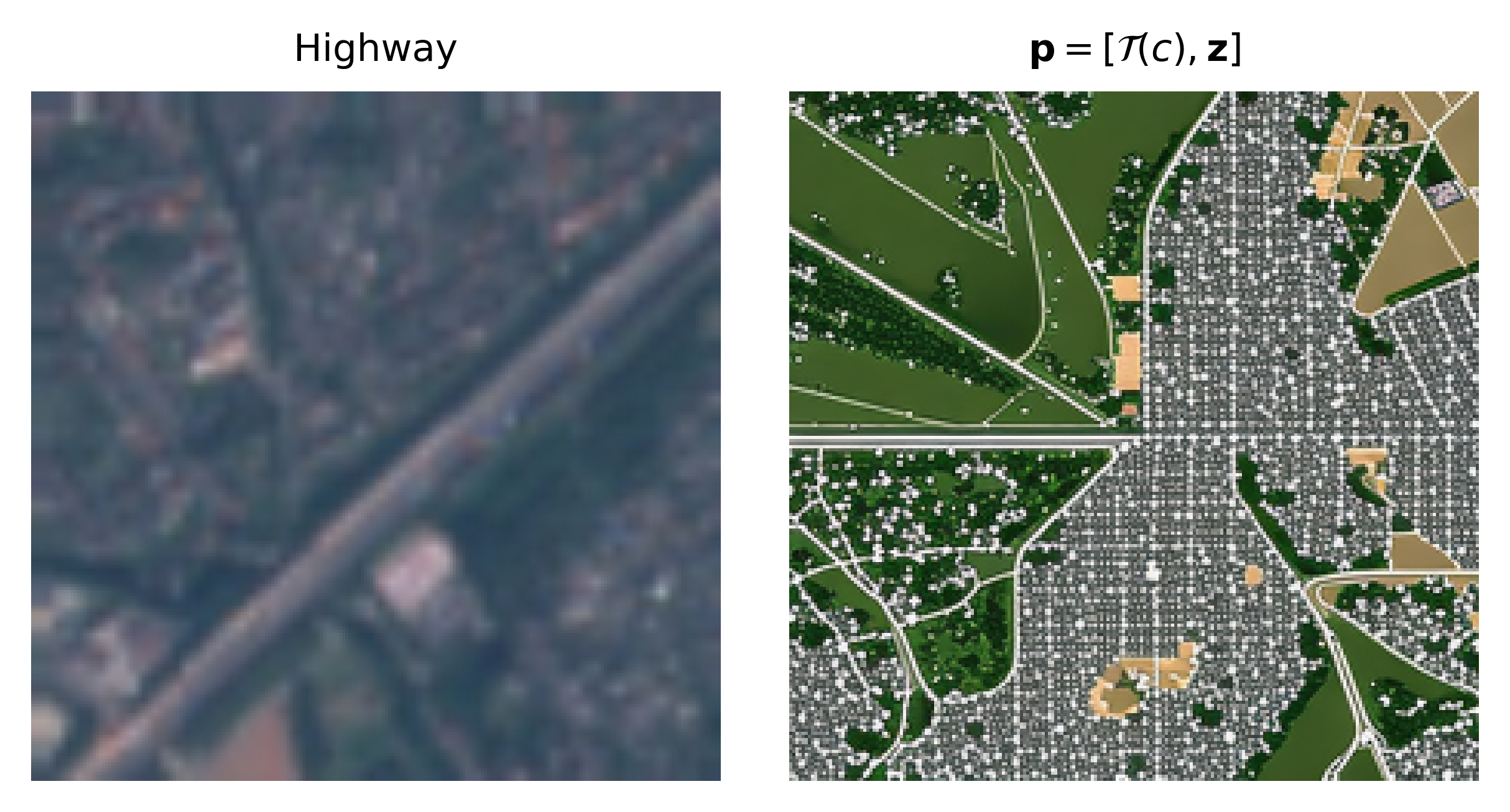} &
\includegraphics[width=0.18\textwidth]{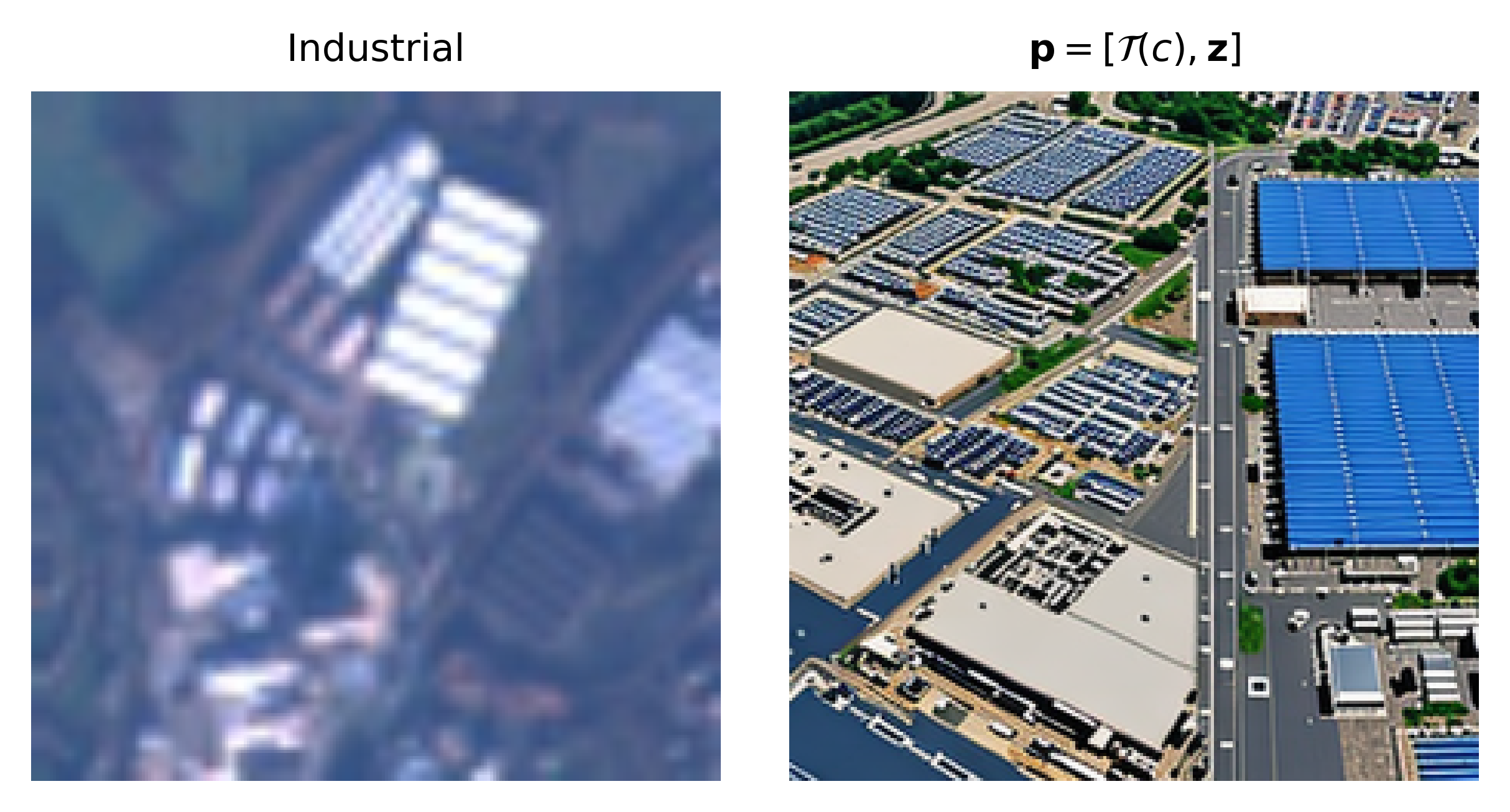} &
\includegraphics[width=0.18\textwidth]{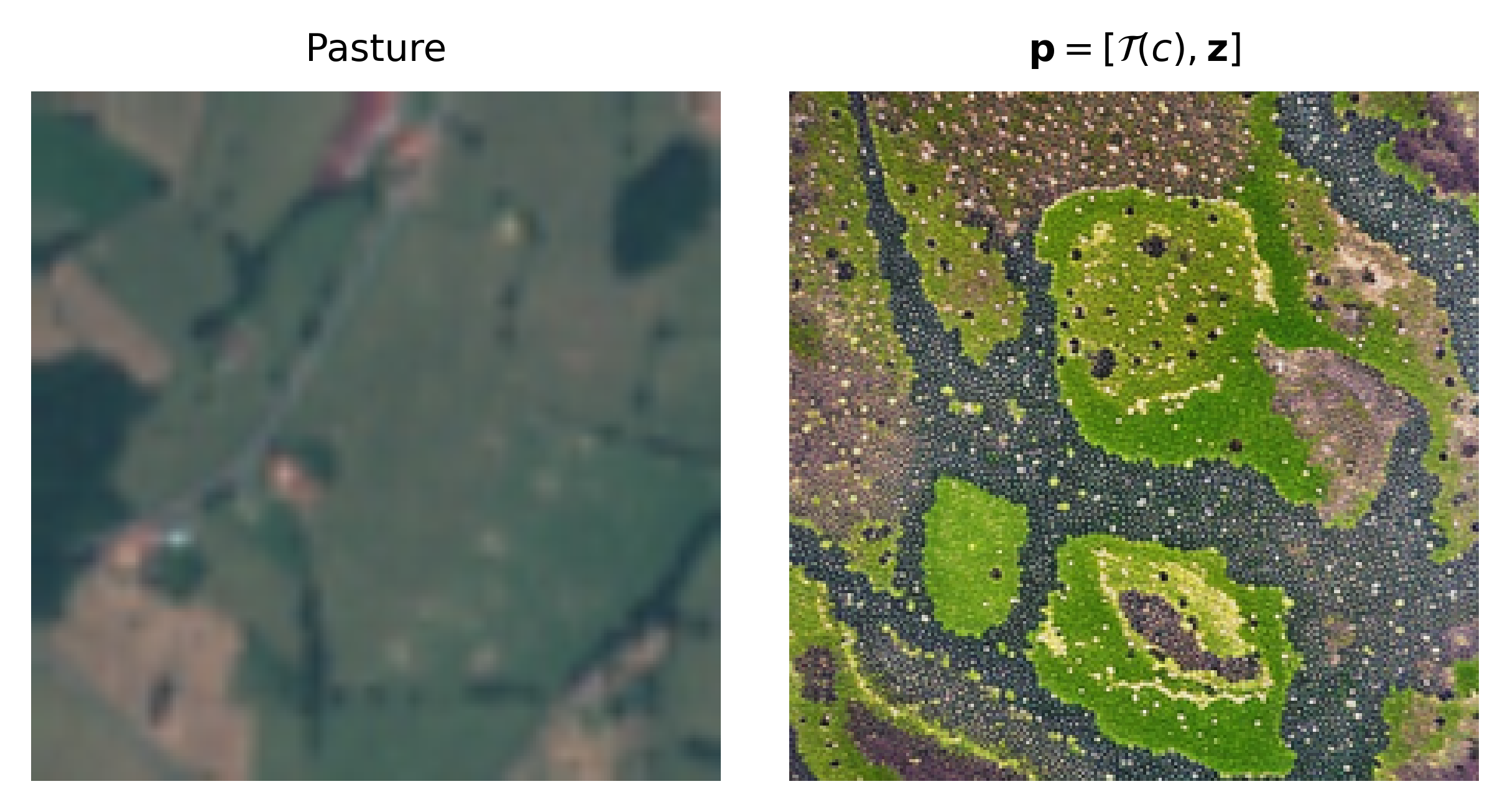} \\
\includegraphics[width=0.18\textwidth]{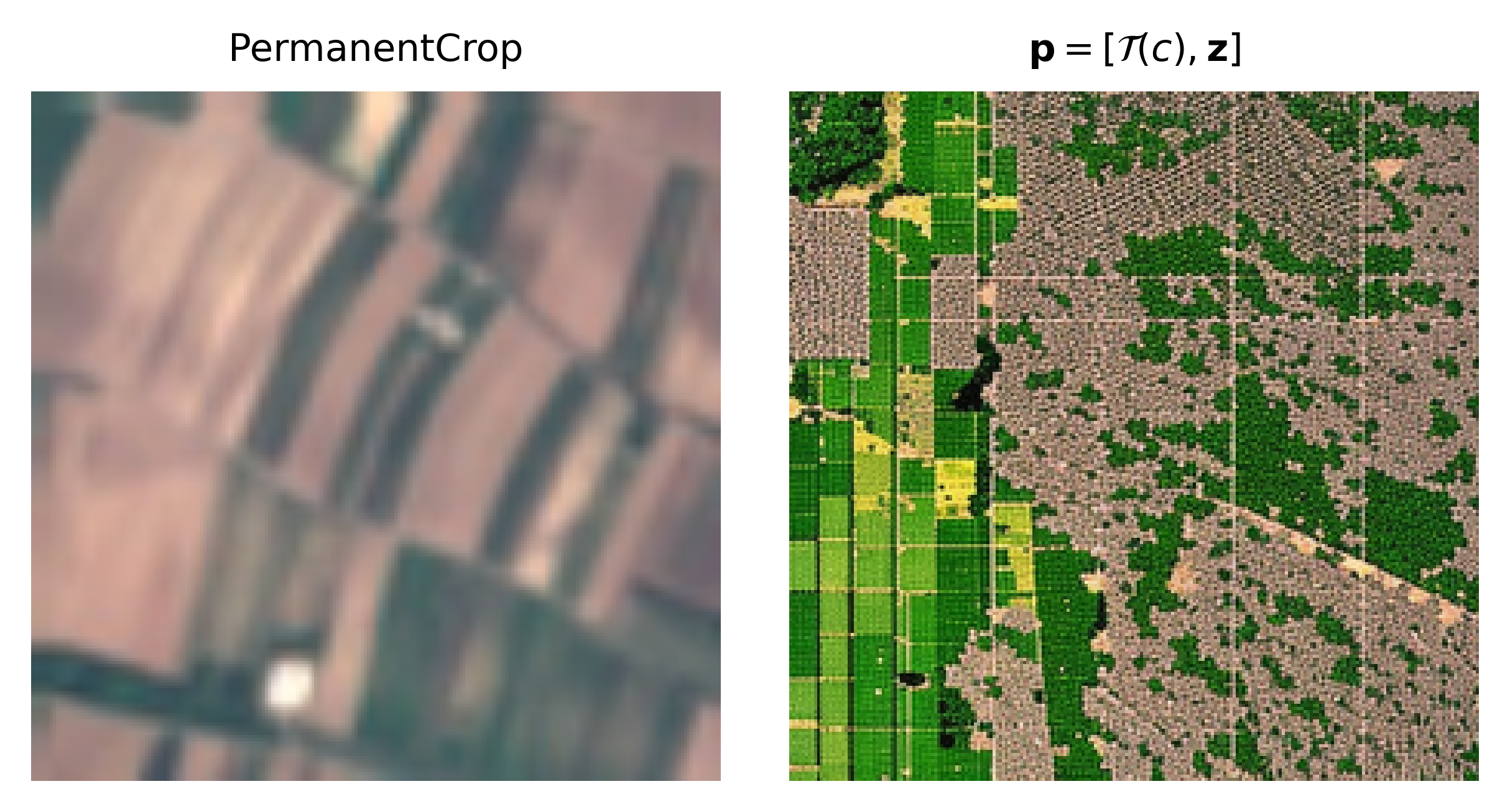} &
\includegraphics[width=0.18\textwidth]{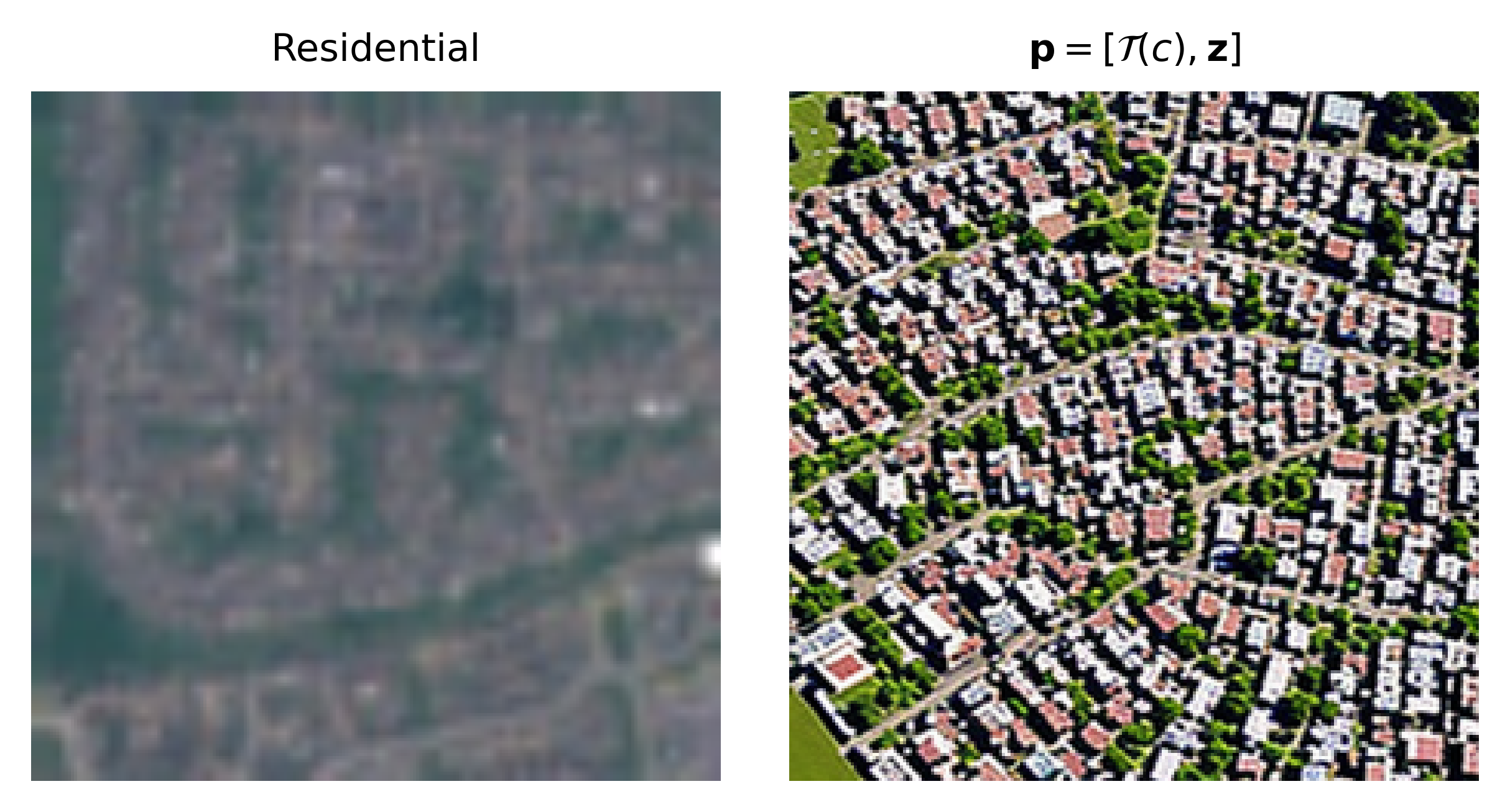} &
\includegraphics[width=0.18\textwidth]{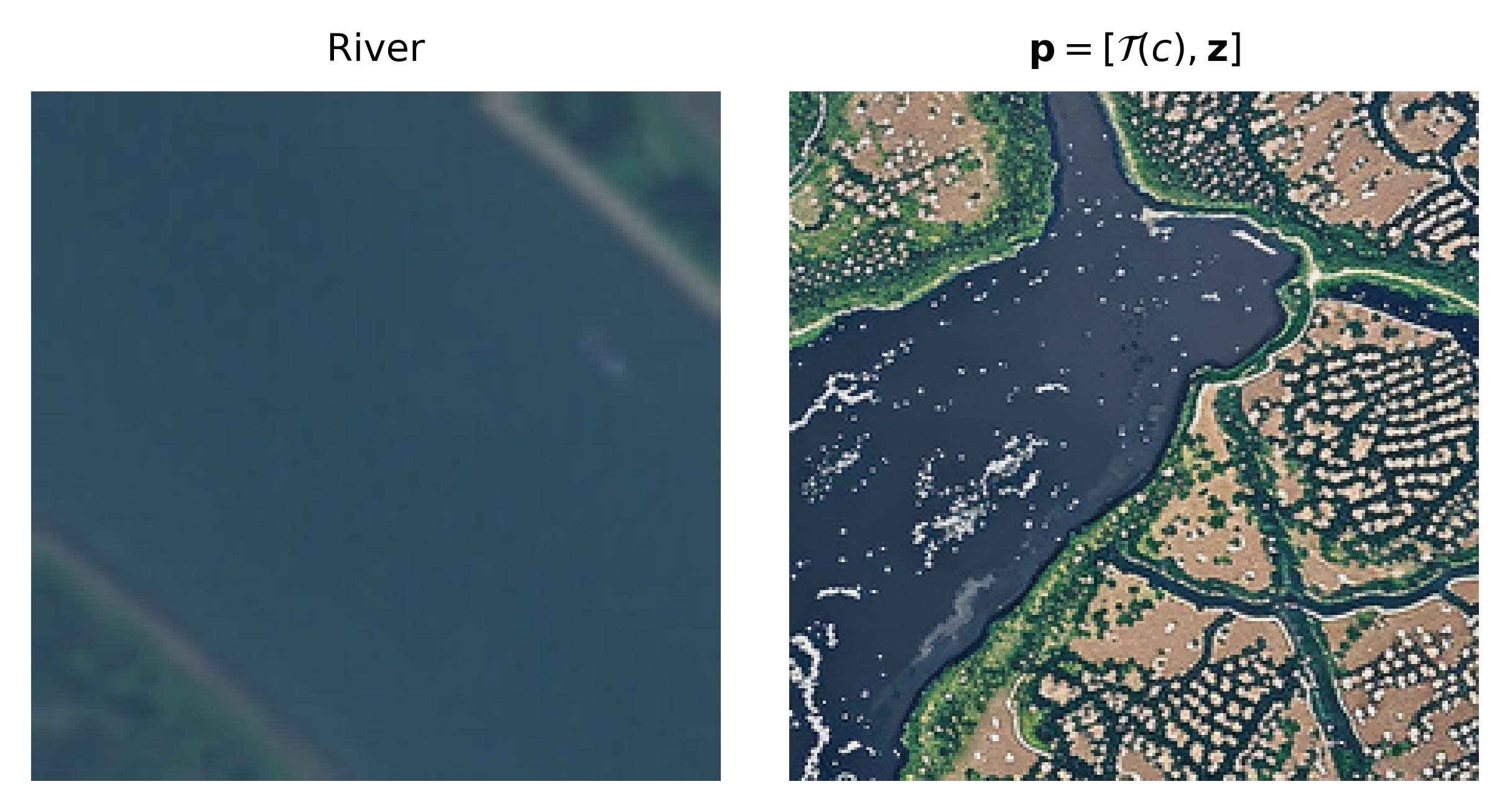} &
\includegraphics[width=0.18\textwidth]{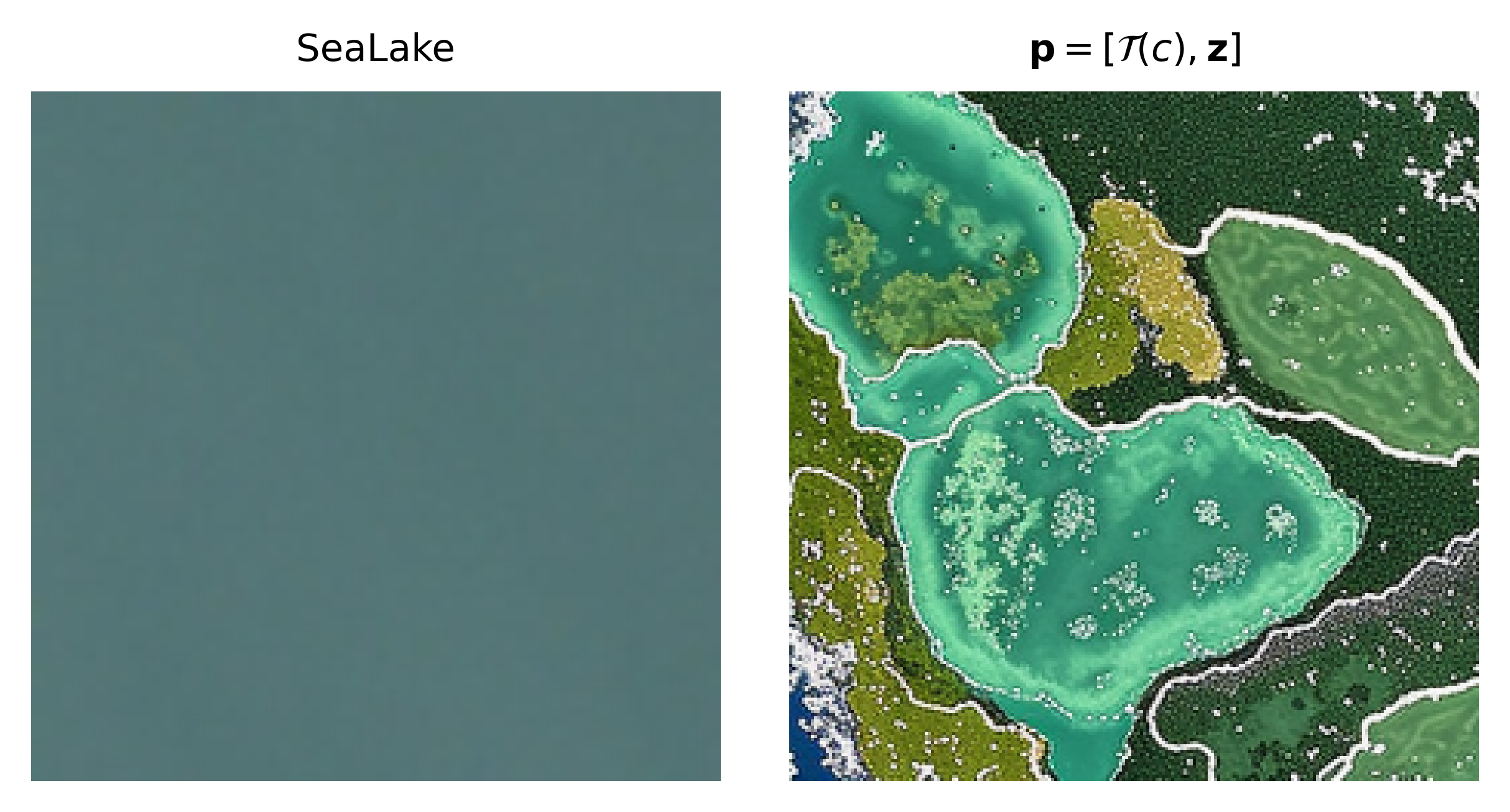} &
\includegraphics[width=0.18\textwidth]{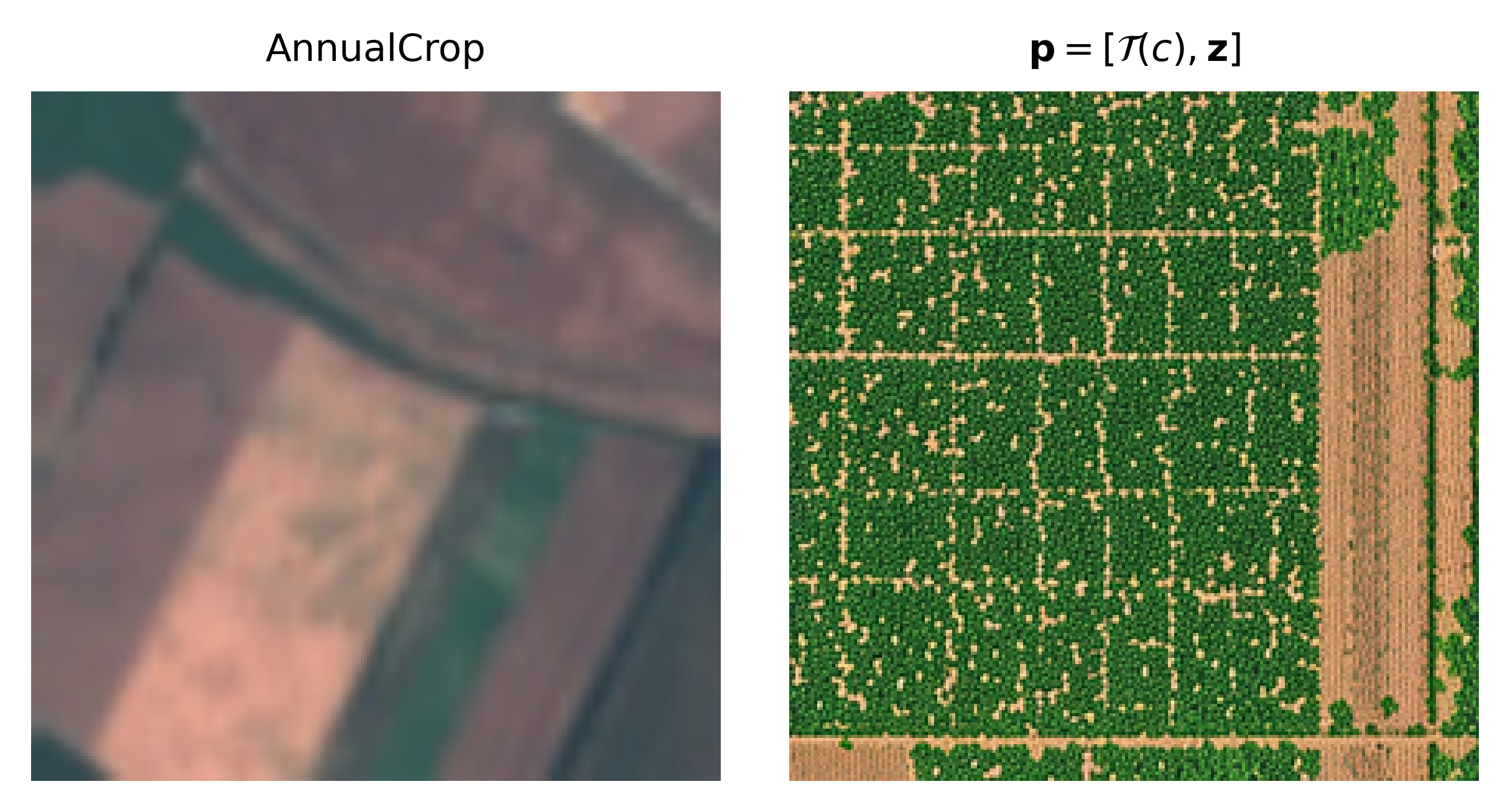}
\end{tabular}

\vspace{0.5em}
\textbf{PlantVillage} \\
\begin{tabular}{ccccc}
\includegraphics[width=0.18\textwidth]{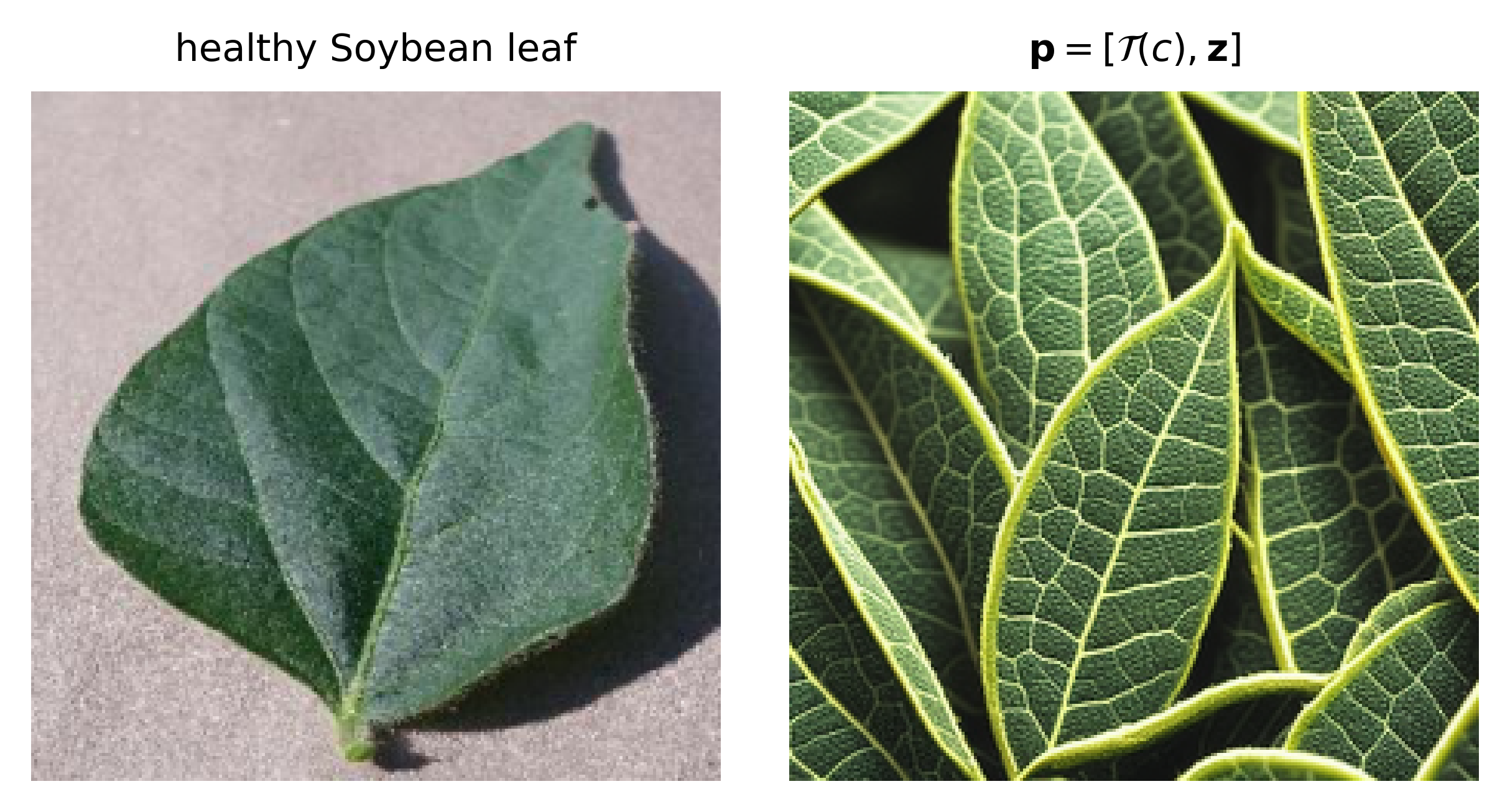} &
\includegraphics[width=0.18\textwidth]{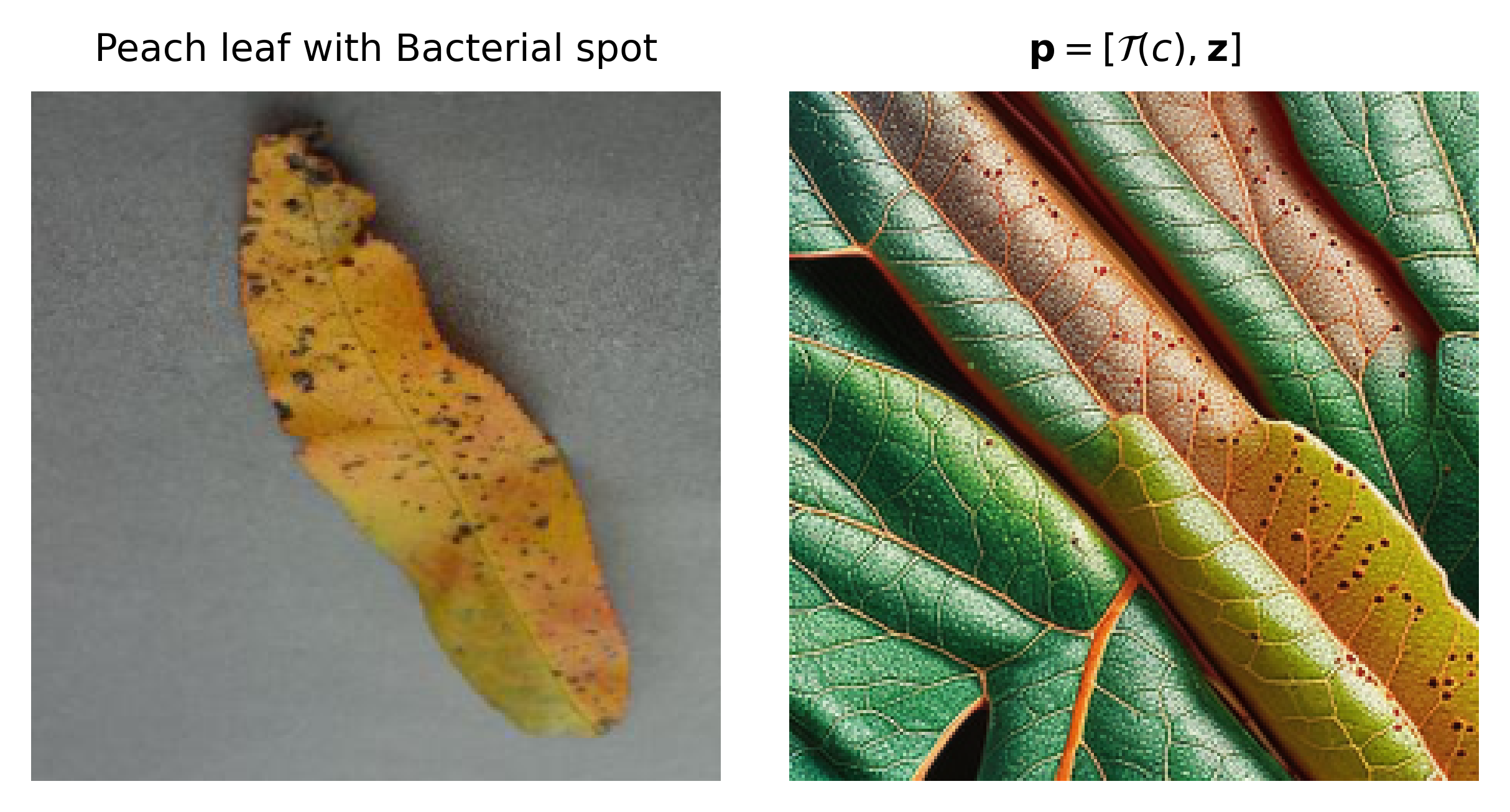} &
\includegraphics[width=0.18\textwidth]{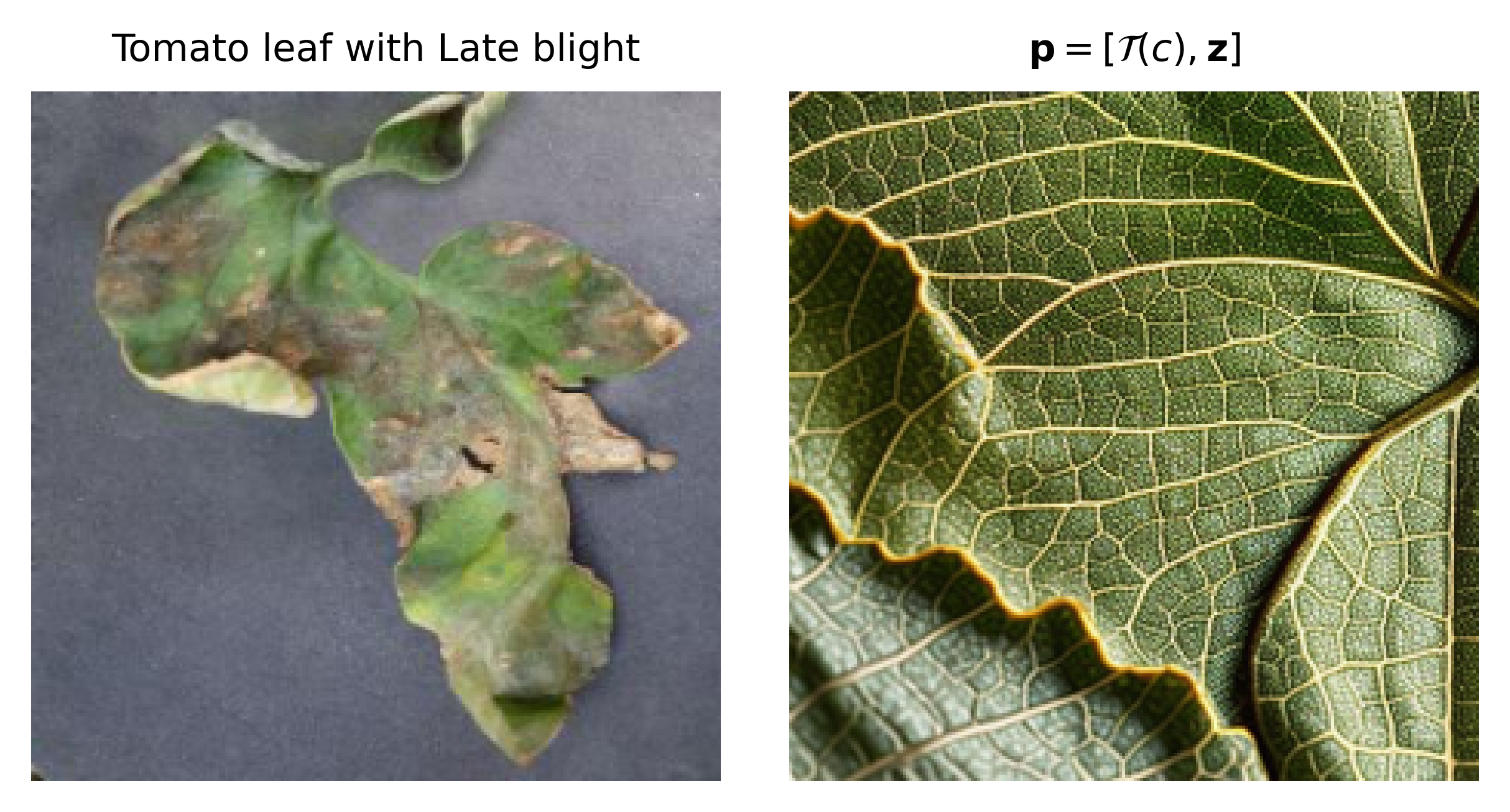} &
\includegraphics[width=0.18\textwidth]{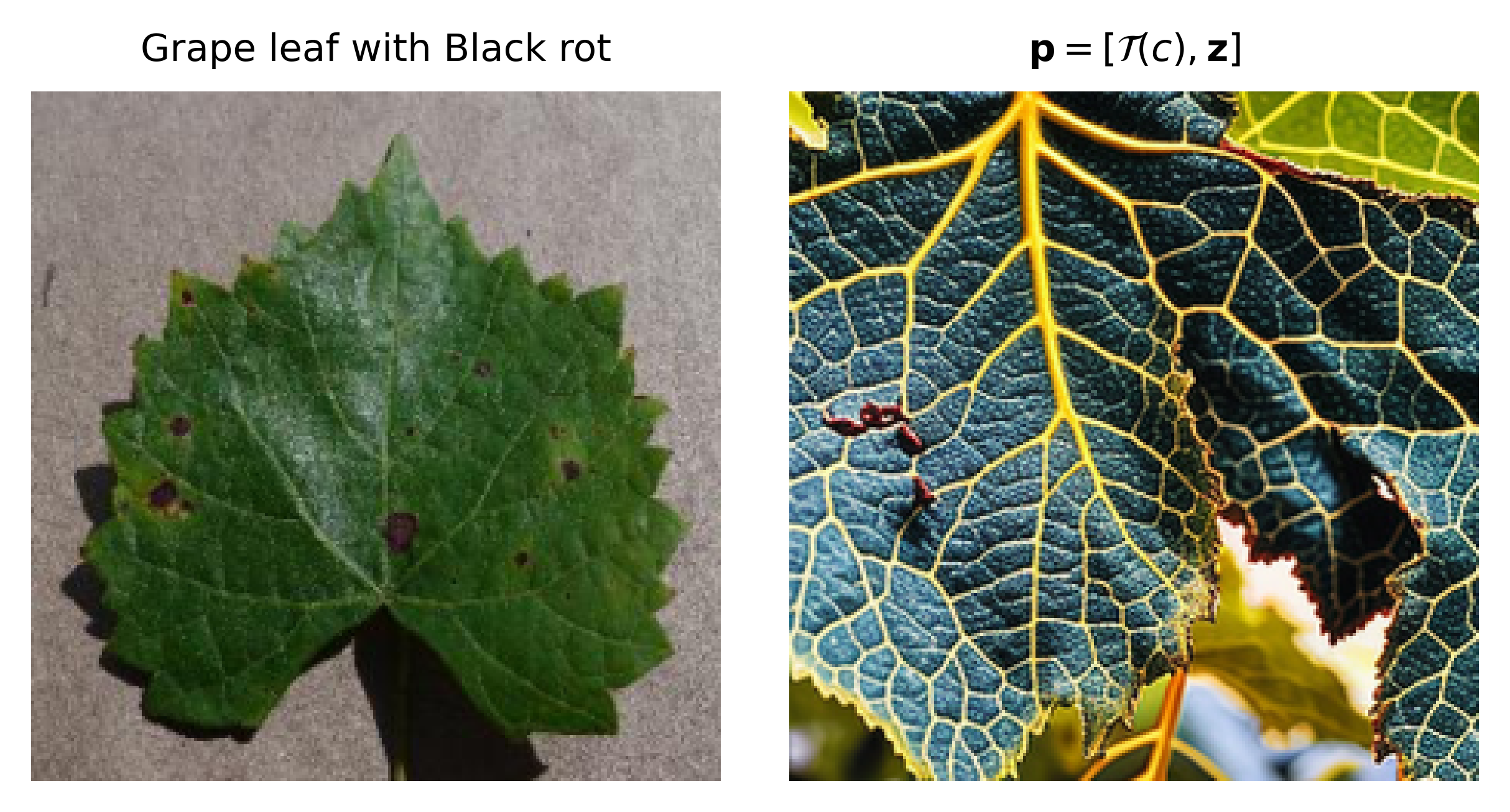} &
\includegraphics[width=0.18\textwidth]{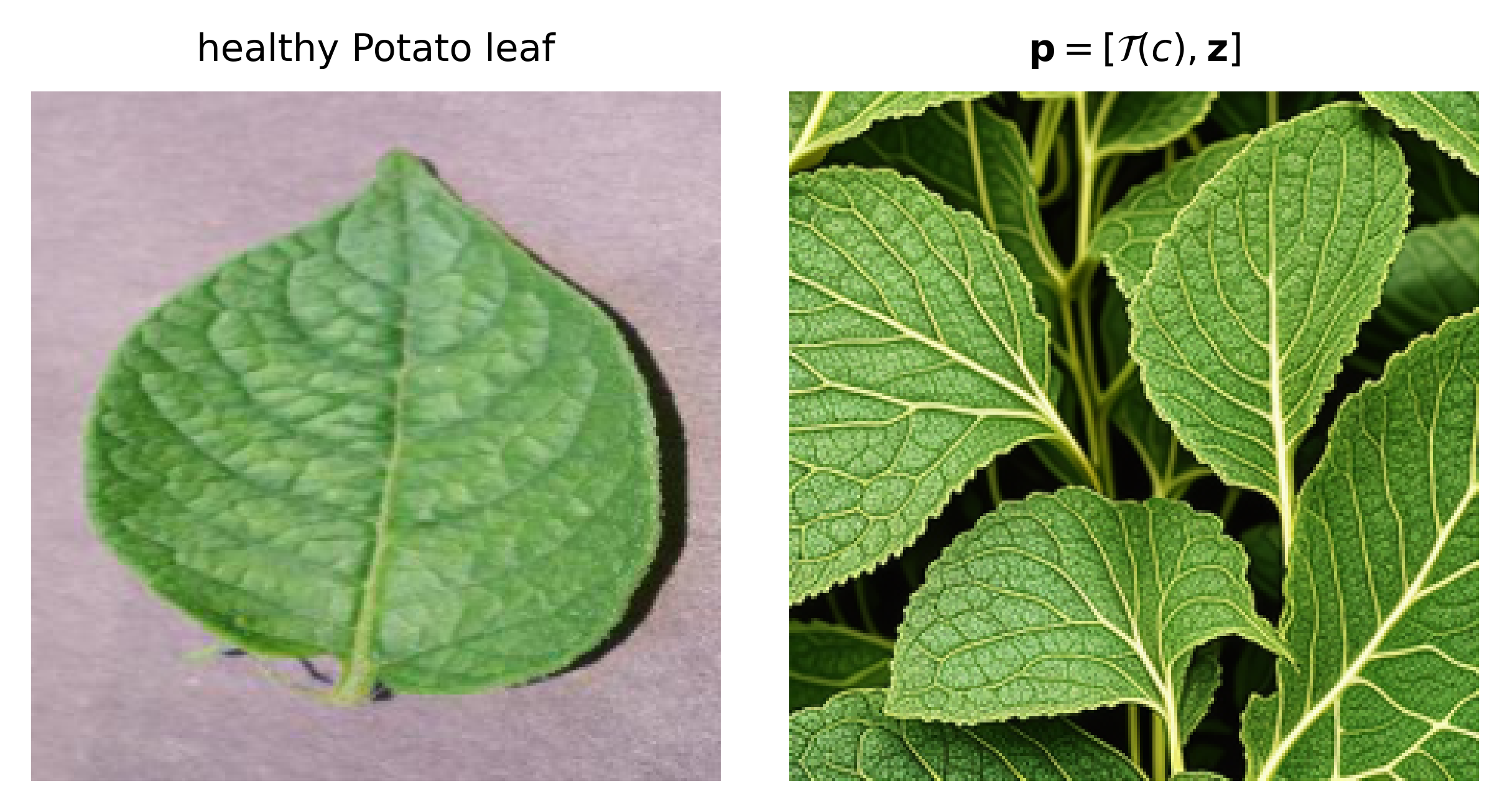} \\
\includegraphics[width=0.18\textwidth]{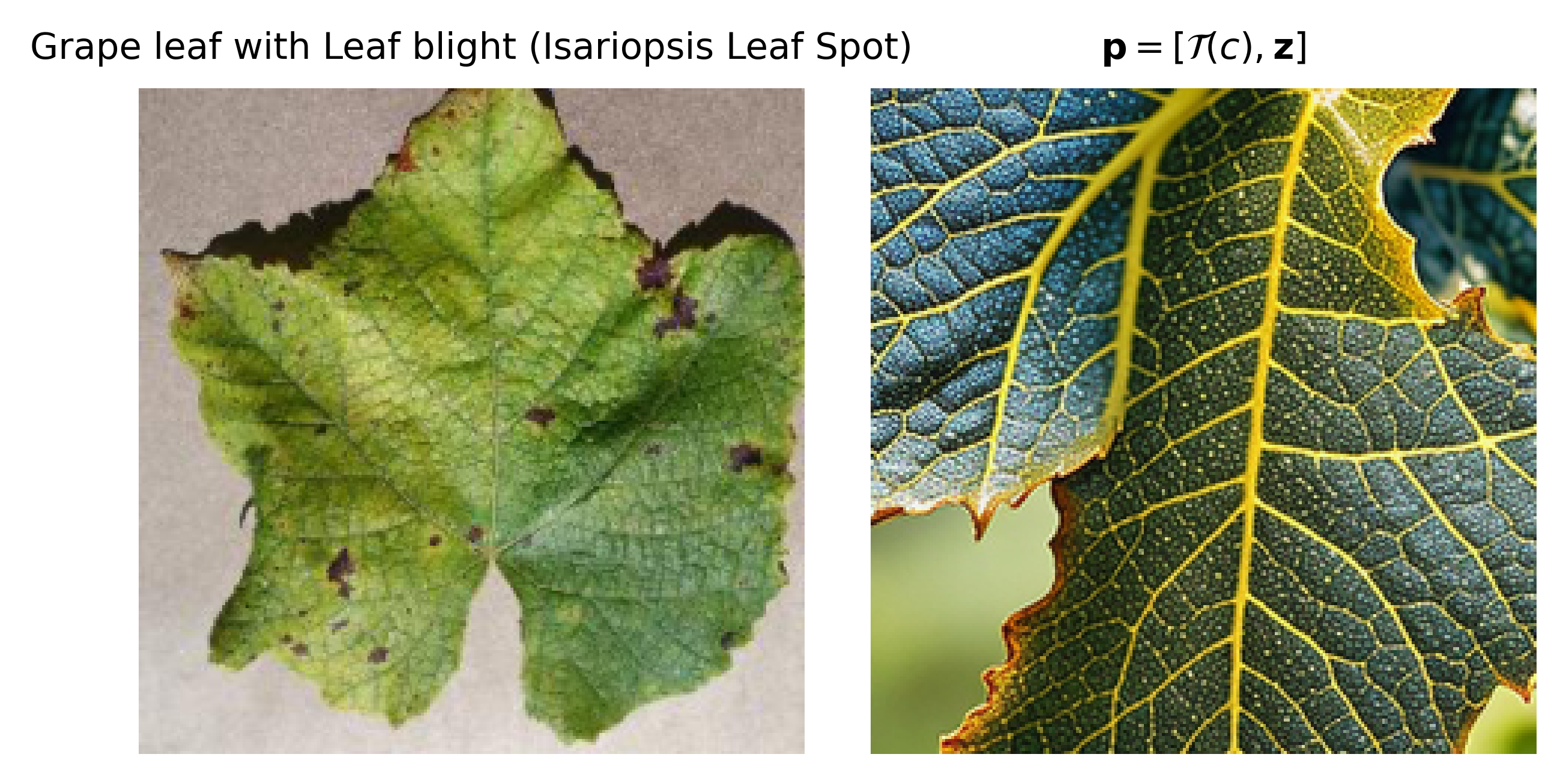} &
\includegraphics[width=0.18\textwidth]{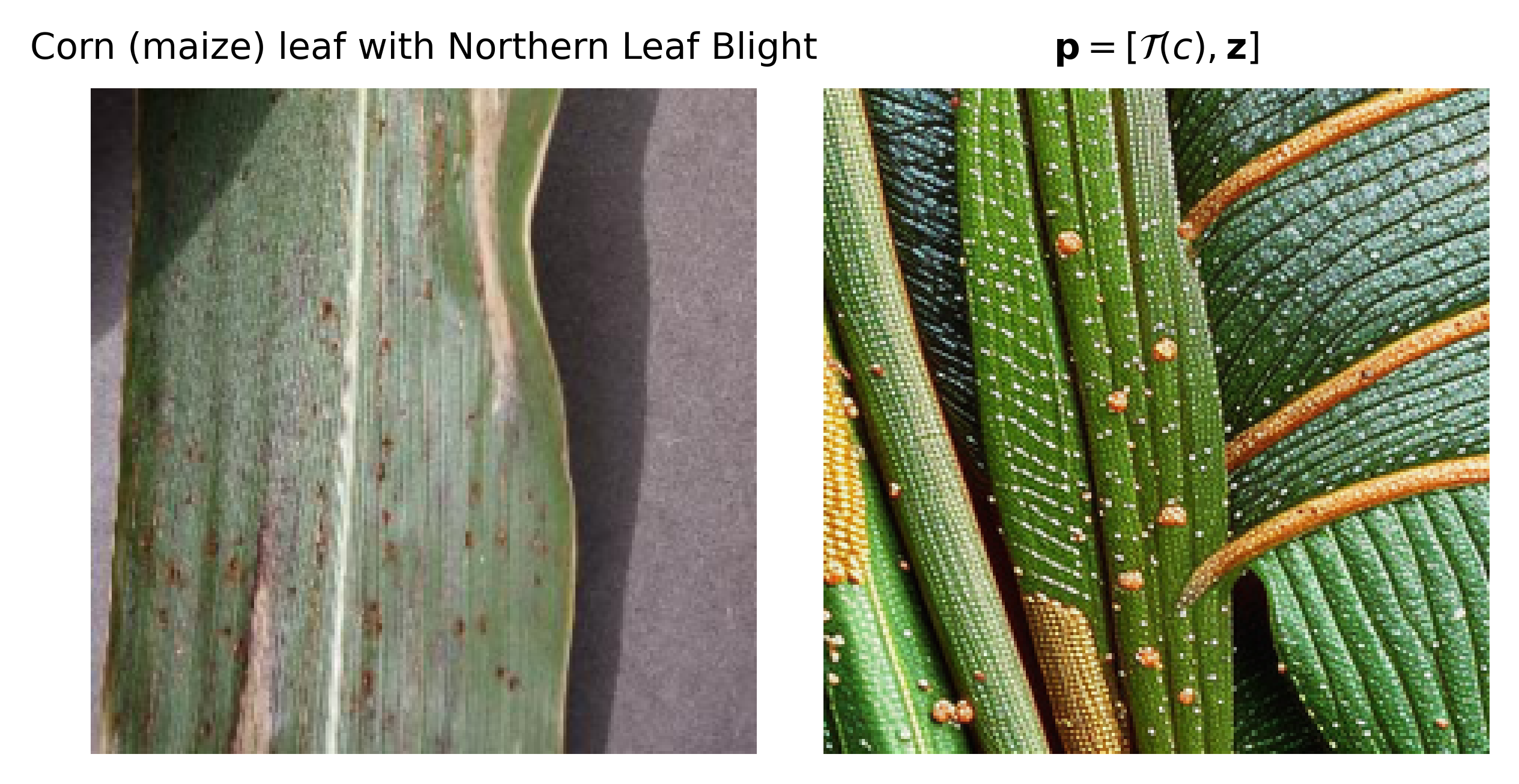} &
\includegraphics[width=0.18\textwidth]{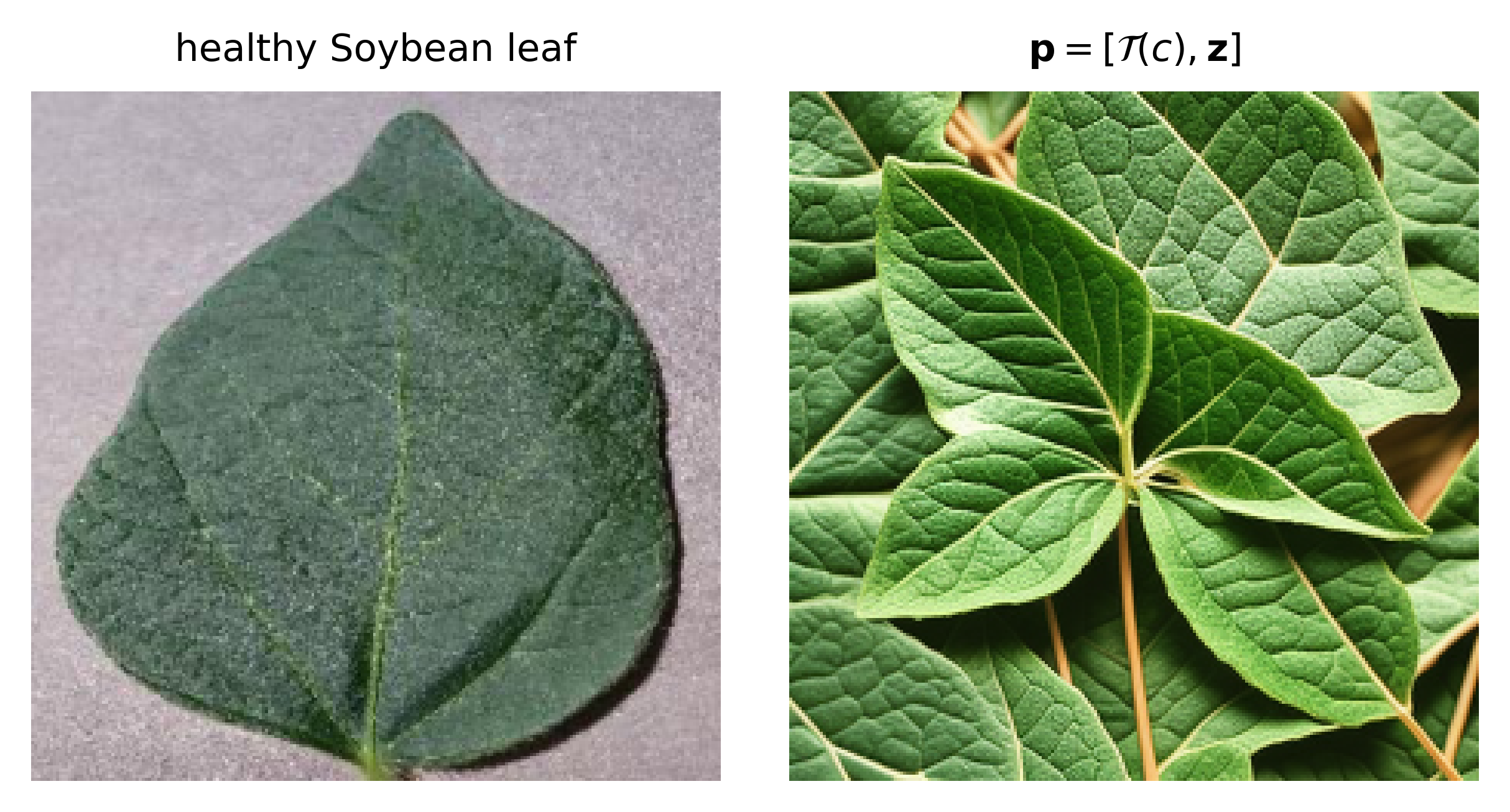} &
\includegraphics[width=0.18\textwidth]{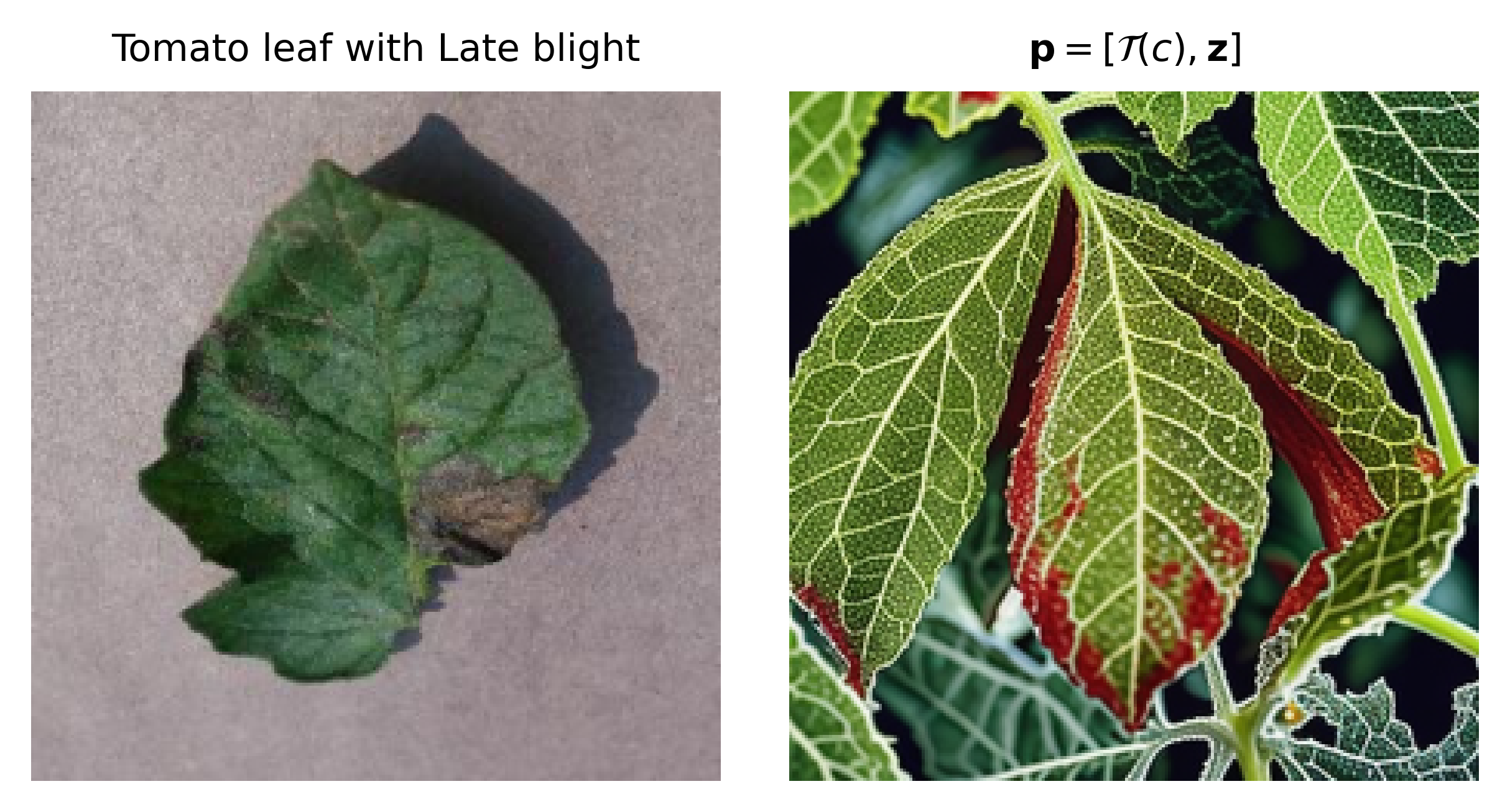} &
\includegraphics[width=0.18\textwidth]{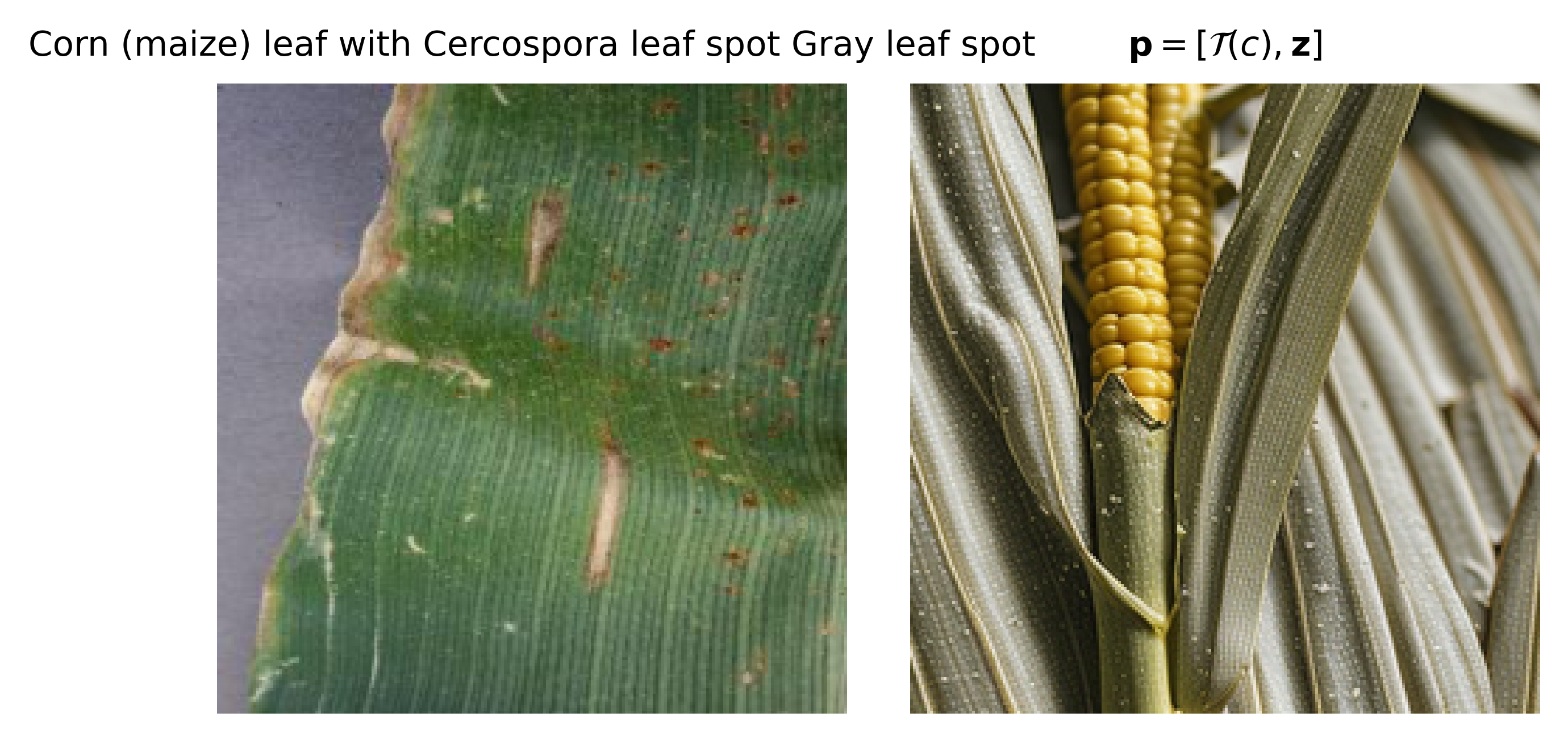}
\end{tabular}

\vspace{0.5em}
\textbf{PathMNIST} \\
\begin{tabular}{ccccc}
\includegraphics[width=0.18\textwidth]{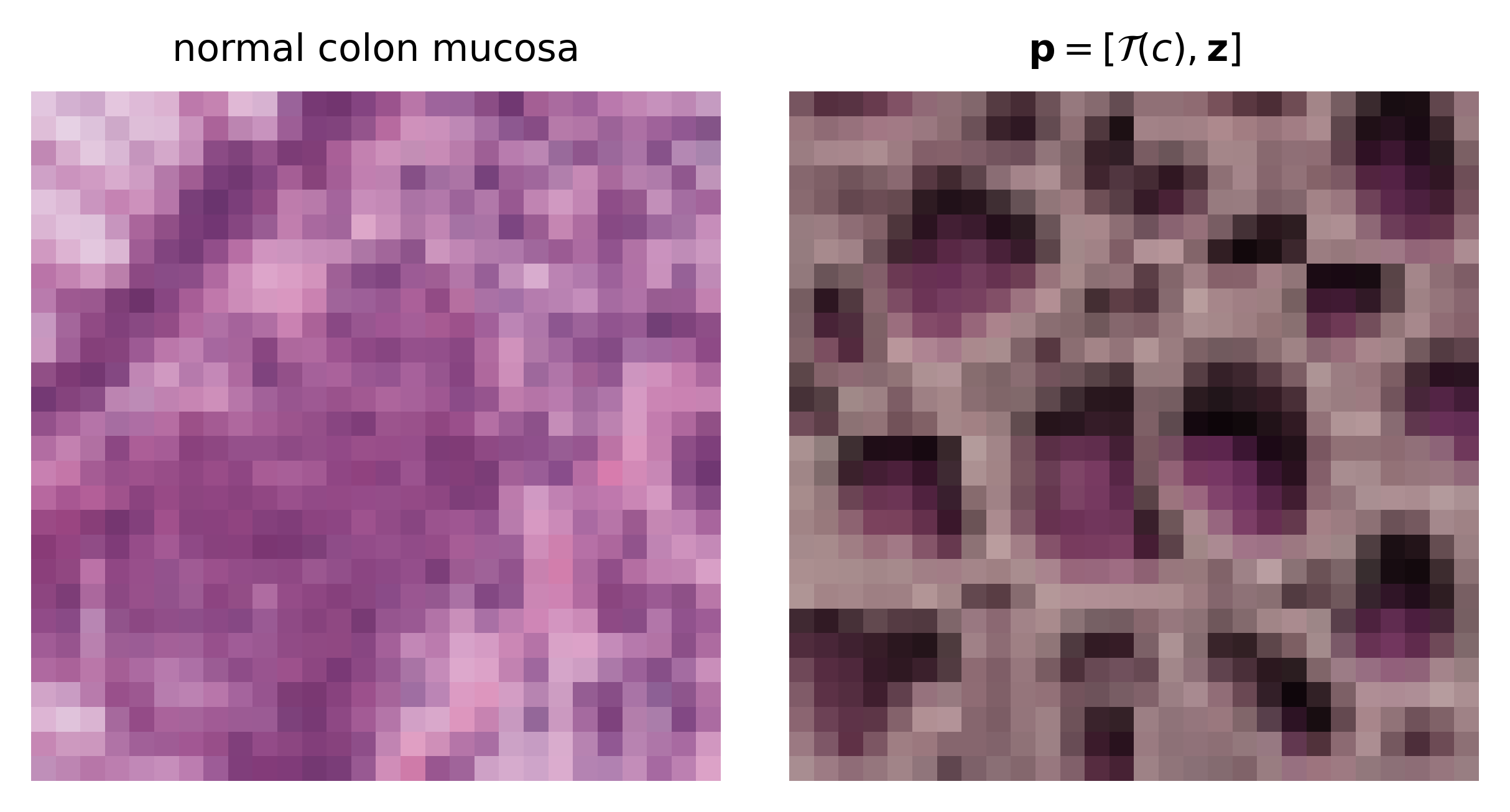} &
\includegraphics[width=0.18\textwidth]{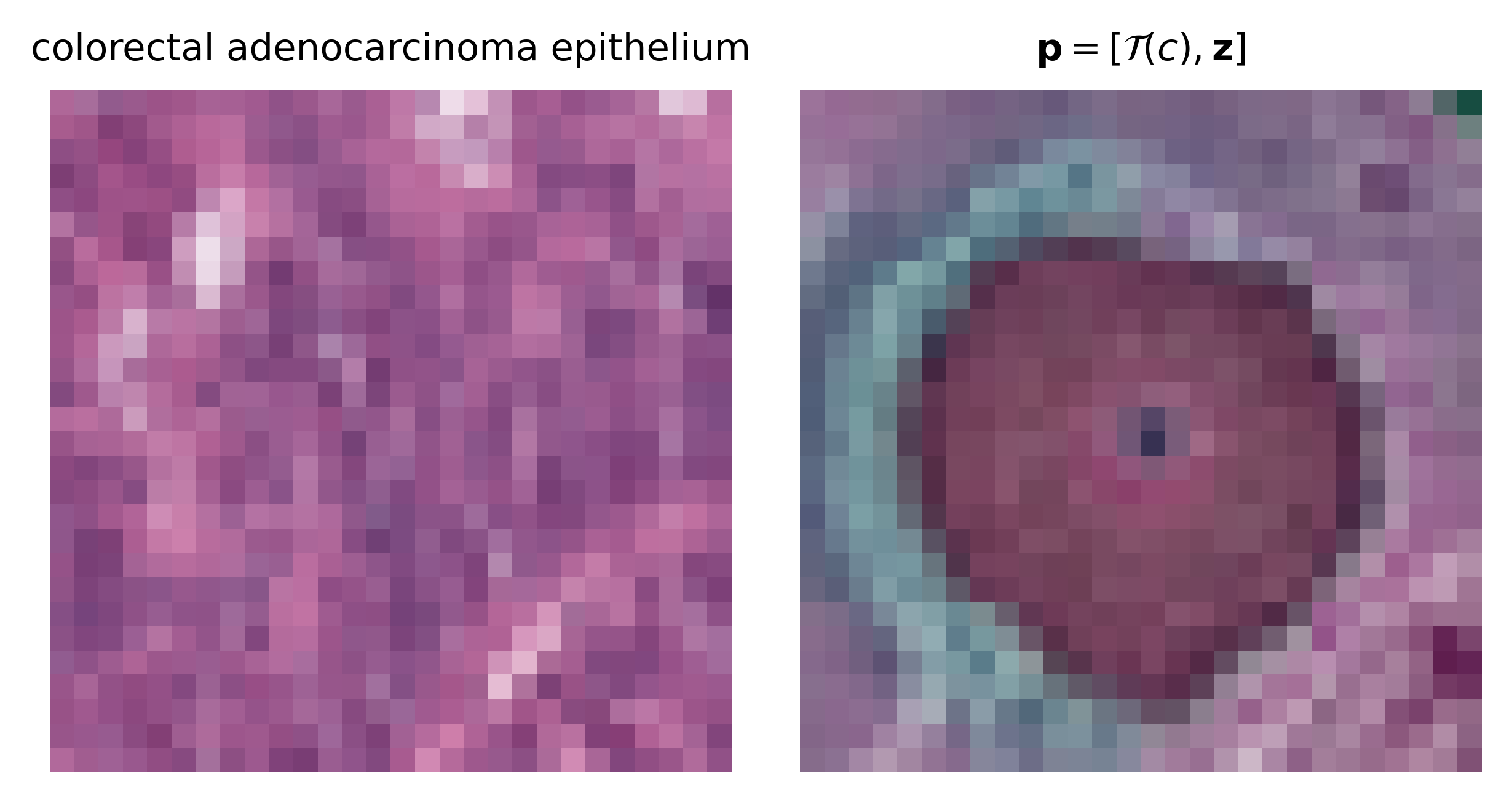} &
\includegraphics[width=0.18\textwidth]{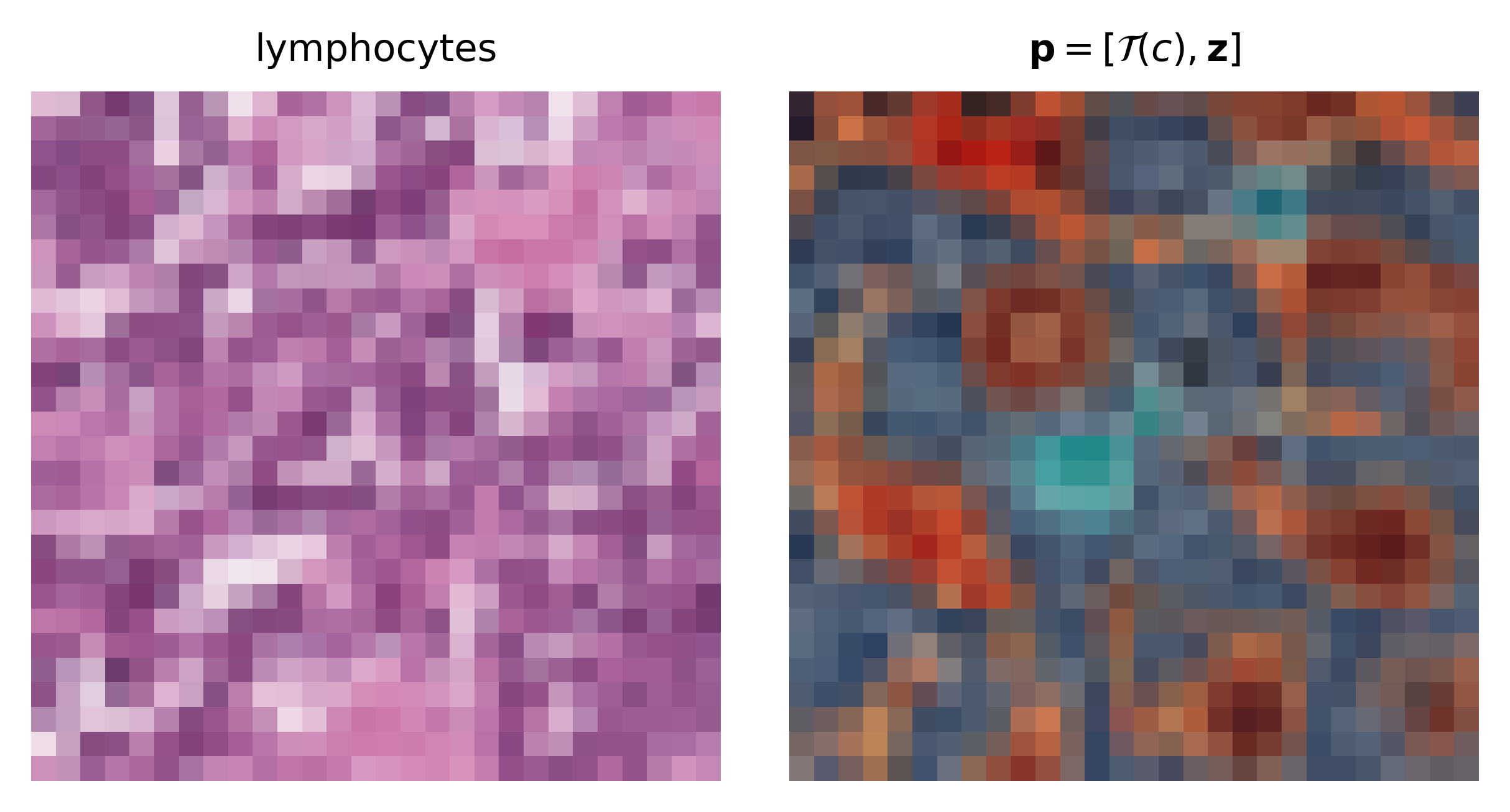} &
\includegraphics[width=0.18\textwidth]{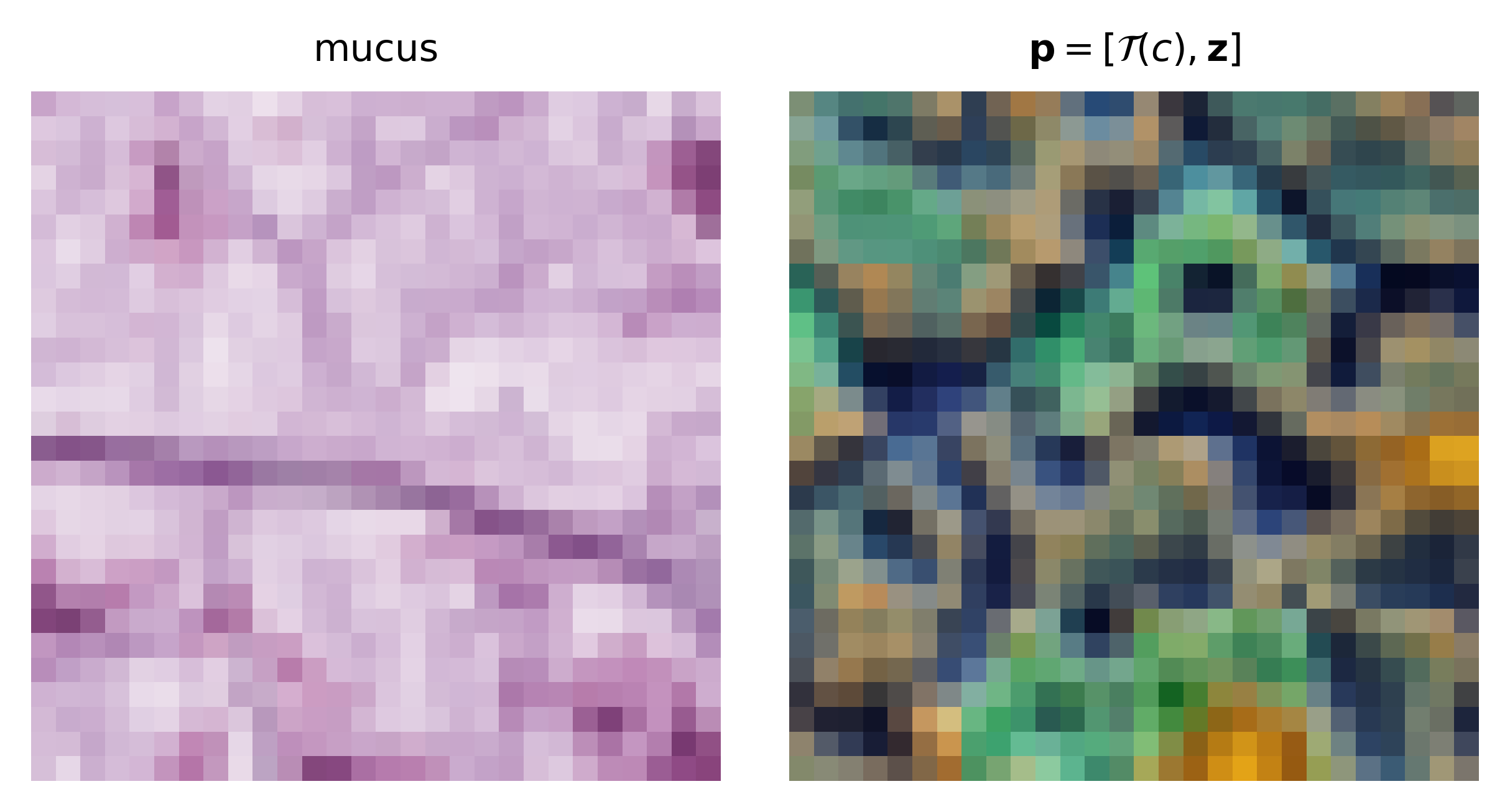} &
\includegraphics[width=0.18\textwidth]{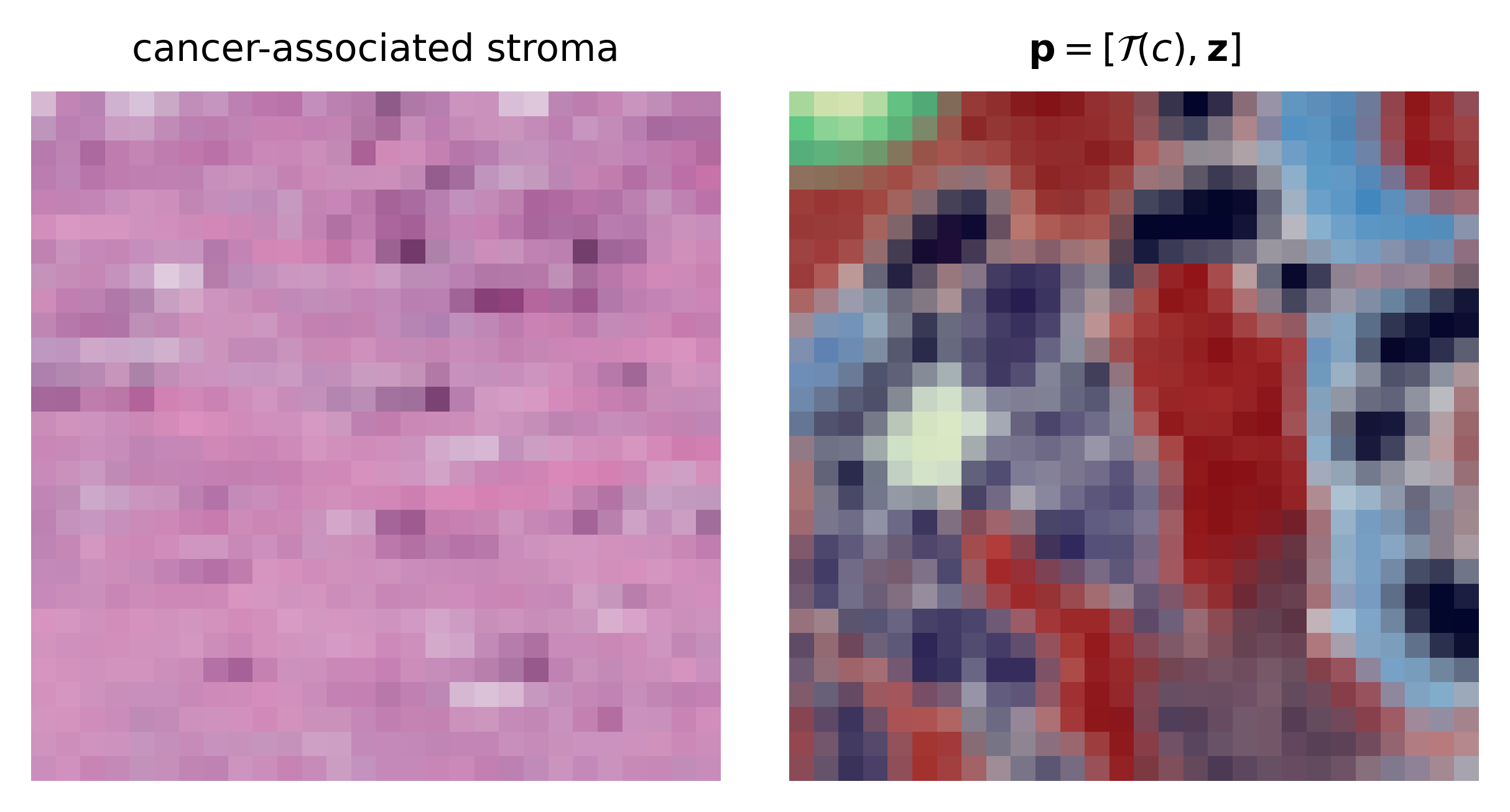} 
\end{tabular}

\caption{
Synthetic samples generated across four datasets using the proposed reconstruction method. 
The images exhibit strong semantic similarity to the corresponding local data distributions, demonstrating the effectiveness of the embedding-driven generation process.}
\label{fig:synthetic_samples}
\end{figure*}

\subsection{Synthetic Data Visualization and Analysis}
This experiment evaluates whether the proposed embedding-to-input reconstruction process generates semantically meaningful and task-relevant samples that can serve as a surrogate dataset for adapting the FM. The reconstruction process is formulated as an inverse problem that maps embedding vectors to input samples~\cite{mahendran2015understanding}. Using the proposed prompt-based mechanism, we generate synthetic samples by optimizing prompts $\{\mathbf{p}_i\}$ corresponding to supervising embeddings $\{\mathbf{h}_i\}$, and obtaining $\hat{x}_i = G(\mathbf{p}_i)$. In practice, we use a batch of $16$ embeddings for synthetic image generation, where embeddings are selected from as many different classes as possible to encourage class diversity in the reconstructed samples.

Fig.~\ref{fig:synthetic_samples} provides representative generated samples across different datasets. For generic object datasets such as CIFAR-10 and Imagenette, the reconstructed samples exhibit clear semantic alignment with their labels and capture meaningful visual structures, indicating that the embedding-guided reconstruction is effective in these settings. For domain-specific datasets such as EuroSAT and PlantVillage, the generated samples also preserve relevant structural patterns (e.g., land-use layouts or leaf textures), suggesting that the proposed approach can generalize beyond standard visual domains. In contrast, the generated samples for PathMNIST are noticeably less realistic and exhibit weaker semantic alignment with the underlying pathology classes. This can be attributed to the limited representational capability of both the pretrained FM and the text-to-image generator in the medical domain. Consequently, the reconstructed samples do not fully capture the distribution of the original dataset, resulting in a lower-quality surrogate dataset for FM adaptation. This observation is consistent with the results in Table~\ref{tab:main_results}, where the performance gains of the proposed framework are smaller on PathMNIST than on the other datasets.

\subsection{Ablation Studies}
To quantify the contribution of each component, we perform ablation studies on CIFAR-10 under both personalization and generalization settings. The results are summarized in Table~\ref{tab:ablation}.

\paragraph{Effect of Reference Space Construction}
$\KM$ relies on edge device embeddings $\{\mathbf{H}_n\}_{n=1}^{N}$ to construct a shared reference space and communicate distributed task knowledge to the server. To evaluate the importance of this component, we remove all edge device embeddings and generate synthetic samples solely from the class anchors produced by the FM text encoder. In this setting, the generated dataset contains only generic semantic information and does not reflect the downstream task knowledge distributed across edge devices. As shown in Table~\ref{tab:ablation}, performance decreases noticeably, demonstrating that embedding-based alignment is essential for transferring task-specific knowledge and constructing an informative surrogate dataset.

\paragraph{Effect of FM Adaptation}
We further evaluate the impact of adapting the FM on the reconstructed dataset. In this variant, the pretrained FM is used directly without LoRA adaptation before distillation. The resulting performance degradation confirms that FM adaptation plays an important role in bridging the gap between the pretrained FM and the target task. By incorporating knowledge reconstructed from edge device embeddings, the adapted FM serves as a more effective teacher for the subsequent distillation stage.

\paragraph{Effect of Global Knowledge Utilization}
Finally, we investigate the contribution of the deployment strategies used after obtaining the distilled global model. For personalization, removing local adaptation and directly deploying the global model reduces accuracy, indicating that local fine-tuning remains important for capturing edge device-specific distributions. For generalization, replacing the logit-adjusted objective with standard training also degrades performance, highlighting the benefit of explicitly compensating for class imbalance during local adaptation.

\begin{table}[t]
\centering
\caption{Ablation study of key components of $\KM$ on CIFAR-10 under personalization (P) and generalization (G).}
\label{tab:ablation}
\begin{tabular}{l|cc}
\hline
\textbf{Variant} & \textbf{P} & \textbf{G} \\
\hline
\text{ECLIPSE} & \textbf{71.46} & \textbf{64.82} \\
w/o $\{\mathbf{H}_n\}_{n=1}^{N}$ (anchor-only reference space) & 58.21 & 52.74 \\
w/o FM adaptation & 66.37 & 59.91 \\
w/o personalization & -- & 61.58 \\
w/o generalization strategy ($L_{\mathrm{LA}}$) & 68.83 & -- \\
\hline
\end{tabular}
\end{table}

\subsection{Limitations}
A key limitation of the $\KM$ framework arises in scenarios where the downstream task is highly domain-specific, and the underlying FM lacks sufficient representational capability for that domain. In such cases, the embedding-driven reconstruction process may produce synthetic samples that exhibit a noticeable domain shift relative to the true data distribution. This mismatch reduces the effectiveness of the FM adaptation stage, resulting in a weaker distilled global model and degraded performance on edge devices. Consequently, the performance of the proposed framework depends on the representational quality of the underlying FM and the degree of alignment between its pretraining corpus and the target task domain. Future work will investigate the use of domain-adapted foundation models and domain-specific generative models to further improve performance in specialized application domains.

\section{Conclusion}
This paper presents a data-free federated knowledge transfer framework, referred to as the $\KM$, that enables resource-constrained edge devices to benefit from a server-hosted foundation model without requiring public datasets or local model exchange. By establishing a shared embedding reference space, $\KM$ aligns and aggregates distributed task-specific knowledge, reconstructs a task-aware surrogate dataset, and adapts the FM through server-side learning. Experimental results indicate consistent improvements over conventional FL, data-free FKT, and server-FM baselines, while providing robustness to statistical and model heterogeneity and reducing communication and edge-side computation costs. These findings highlight embedding-based alignment as an effective approach for FM adaptation in data-free federated edge networks.

\bibliographystyle{IEEEtran}
\bibliography{main}

\end{document}